\documentclass{article} %
\usepackage{iclr2024_conference,times}

\usepackage{amsmath,amsfonts,bm}

\def\eqref#1{equation~\ref{#1}}

\def\1{\bm{1}}

\DeclareMathAlphabet{\mathsfit}{\encodingdefault}{\sfdefault}{m}{sl}
\SetMathAlphabet{\mathsfit}{bold}{\encodingdefault}{\sfdefault}{bx}{n}

\usepackage{tablefootnote}
\usepackage{float}
\usepackage{multirow}
\usepackage{scrextend}
\usepackage{hyperref}
\usepackage{parselines} 
\usepackage{scrextend}
\hypersetup{
    colorlinks=true,
    linkcolor=blue,
    filecolor=magenta,      
    urlcolor=cyan
}
\usepackage[utf8]{inputenc}

\usepackage{tcolorbox}

\usepackage{wrapfig}

\usepackage{pifont}
\usepackage{caption}
\usepackage{subcaption}
\usepackage{nicefrac}
\usepackage{xstring}
\usepackage{newfloat}

\newcommand{\DefineRemark}[2]{%
  \expandafter\newcommand\csname rmk-#1\endcsname{#2}%
}
\newcommand{\Remark}[1]{\csname rmk-#1\endcsname}

\usepackage[nameinlink,capitalize]{cleveref}

\crefname{section}{\S}{\S}
\Crefname{section}{\S}{\S}
\crefname{appendix}{App.}{Apps.}
\Crefname{appendix}{App.}{Apps.}
\crefname{theorem}{Thm.}{Thms.}
\Crefname{theorem}{Thm.}{Thms.}
\crefname{proposition}{Prop.}{Props.}
\Crefname{proposition}{Prop.}{Props.}

\crefname{algorithm}{Alg.}{Algs.}
\Crefname{algorithm}{Alg.}{Algs.}
\crefname{assumption}{Asm.}{Asms.}
\Crefname{assumption}{Asm.}{Asms.}
\crefname{mechanism}{Mech.}{Mechs.}
\Crefname{mechanism}{Mech.}{Mechs.}

\newcommand{\myparatightestn}[1]{ \noindent\textbf{{#1}}}

\newcommand{\myparaemphtightestn}[1]{\noindent\emph{{#1}}}

\newcounter{packednmbr}

\newcommand{\concept}[1]{\emph{{#1}}}

\newcommand{\vicunadataset}{Vicuna-80}
\newcommand{\wizardlm}{WizardLM}
\newcommand{\method}{Skeleton-of-Thought}
\newcommand{\methodshort}{SoT}
\newcommand{\methodrshort}{SoT-R}
\newcommand{\methodbase}{normal}
\newcommand{\Methodbase}{Normal}

\DefineRemark{llama2}{LLaMA2}
\DefineRemark{llamachat7B2}{LLaMA2-Chat-7B}
\DefineRemark{llamachat13B2}{LLaMA2-Chat-13B}
\DefineRemark{vicuna7B1.1}{Vicuna-7B V1.1}
\DefineRemark{vicuna}{Vicuna}
\DefineRemark{llama}{LLaMA}
\DefineRemark{vicuna7B1.3}{Vicuna-7B V1.3}
\DefineRemark{vicuna13B1.3}{Vicuna-13B V1.3}
\DefineRemark{vicuna33B1.3}{Vicuna-33B V1.3}
\DefineRemark{ultralm13B}{UltraLM-13B}
\DefineRemark{ultralm}{UltraLM}
\DefineRemark{openchat13B}{OpenChat-13B}
\DefineRemark{stablevicuna13B}{StableVicuna-13B}

\DefineRemark{chatgpt}{ChatGPT-3.5}
\DefineRemark{gpt4}{GPT-4}

\newcommand{\roberta}{RoBERTa}

\DefineRemark{claude}{Claude}

\newcommand{\fastchat}{FastChat}
\newcommand{\llmzoo}{LLMZoo}

\newcommand{\texample}[6]{\begin{tcolorbox}[colback=yellow!5!white,colframe=yellow!50!black,colbacktitle=yellow!75!black,coltitle=black,title={\footnotesize\textbf{Question:~~~} #2}]
  \textbf{\IfEqCase{#4}{{correct}{\textcolor{green}{\ding{51}}}{incorrect}{\textcolor{red}{\ding{55}}}} ~~\footnotesize\methodshort{}'s answer (#1) \outline{(Orange texts are the skeleton.)} }
  \begin{addmargin}[1em]{2em}%
    {\obeylines
      \scriptsize
      #3
    }
   \end{addmargin}
  \tcblower
  \textbf{\IfEqCase{#6}{
        {correct}{\textcolor{green}{\ding{51}}}
        {incorrect}{\textcolor{red}{\ding{55}}}}~~\footnotesize \Methodbase{} answer (#1)}
    \begin{addmargin}[1em]{2em}%
      {\obeylines
        \scriptsize
        #5
      }
   \end{addmargin}
\end{tcolorbox}\egroup}

\newcommand{\example}{\bgroup\obeylines\obeyspaces\texample}

\newcommand{\texamplesot}[4]{\begin{tcolorbox}[colback=yellow!5!white,colframe=yellow!50!black,colbacktitle=yellow!75!black,coltitle=black,title={\footnotesize \textbf{Question:~~~} #2}]
  \textbf{\IfEqCase{#4}{{correct}{\textcolor{green}{\ding{51}}}{incorrect}{\textcolor{red}{\ding{55}}}} ~~\footnotesize\methodshort{}'s answer (#1) \outline{(Orange texts are the skeleton.)} }
  \begin{addmargin}[1em]{2em}%
    {\obeylines
        \scriptsize
        #3
      }
   \end{addmargin}
\end{tcolorbox}\egroup}

\newcommand{\examplesot}{\bgroup\obeylines\obeyspaces\texamplesot}

\newcommand{\outline}[1]{\obeylines\textcolor{orange}{#1}}
\newcommand{\hide}[1]{\emph{\textcolor{gray}{$<$more texts are omitted...$>$}}}

\newcommand{\promptarg}[1]{\textcolor{red}{\{\textit{#1}\}}}
\newcommand{\prompt}{\bgroup\tprompt}

\newcommand{\tprompt}[3]{\begin{tcolorbox}[colback=cyan!5!white,colframe=cyan!50!black,colbacktitle=cyan!75!black,coltitle=white,title={\footnotesize \textbf{#1}}]
  \begin{addmargin}[1em]{2em}%
    \scriptsize
    \textbf{[User:]} #2
      \end{addmargin}
      \begin{addmargin}[1em]{2em}%
          \scriptsize
          \textbf{[Assistant:]} #3
      \end{addmargin}
    \end{tcolorbox}\egroup}

  \DeclareFloatingEnvironment[fileext=frm,placement={!ht},name=Frame]{myfloat}

  \newcounter{promptenv}
  \newenvironment{promptenv}[4][]{
    \begin{myfloat}[tb]\refstepcounter{promptenv} \prompt{\textbf{Prompt~\thepromptenv. #2}}{#3}{#4}
    }{
    \end{myfloat}
  }
\crefname{promptenv}{Prompt}{Prompts}
\Crefname{promptenv}{Prompt}{Prompts}

\usepackage{booktabs}       %
\usepackage{enumerate}
\usepackage{url}
\usepackage{xspace}

\usepackage{minitoc}

\usepackage{xcolor}         %

\newcommand{\revise}[1]{#1}

\title{\method{}: Prompting LLMs for Efficient Parallel Generation}

\author{Xuefei Ning$^1$\thanks{Equal contribution.}\\
\texttt{foxdoraame@gmail.com} 
\And
Zinan Lin$^{2*}$ \\
\texttt{linzinan1995@gmail.com} 
\And
Zixuan Zhou$^{14*}$ \\
\texttt{zhouzx21@mails.tsinghua.edu.cn} 
\And
Zifu Wang$^{3}$ \\
\texttt{zifu.wang@kuleuven.be} 
\And
Huazhong Yang$^{1}$ \\
\texttt{yanghz@tsinghua.edu.cn} 
\And
Yu Wang$^{1}$ \\
\texttt{yu-wang@tsinghua.edu.cn} 
}

\iclrfinalcopy %
\begin{document}

\doparttoc %
\faketableofcontents %

\maketitle

\begin{center}
  \vspace{-22pt}
  {$^{1}$ Department of Electronic Engineering, Tsinghua University, Beijing, China\\$^{2}$ Microsoft Research, Redmond, Washington, USA\\$^{3}$ ESAT-PSI, KU Leuven, Leuven, Belgium\\$^{4}$ Infinigence-AI}%
  \\~\\
  Website: \url{https://sites.google.com/view/sot-llm}\\
  Code: \url{https://github.com/imagination-research/sot}
\end{center}

\begin{abstract}
  This work aims at decreasing the end-to-end generation latency of large language models (LLMs). One of the major causes of the high generation latency is the sequential decoding approach adopted by almost all state-of-the-art LLMs. In this work, motivated by the 
  thinking and writing process of humans, we propose \emph{\method{} (\methodshort{})}, which first guides  LLMs to generate the \emph{skeleton} of the answer, and then conducts parallel API calls or batched decoding to 
  complete the contents of each skeleton point \emph{in parallel}.
  Not only does \methodshort{} provide considerable speed-ups across 12 LLMs,
  but it can also potentially improve the answer quality on several question categories. 
  \methodshort{} is an initial attempt at data-centric optimization 
  for inference efficiency, and \revise{showcases the potential of eliciting high-quality answers by explicitly planning the answer structure in language}.
\end{abstract}

\section{Introduction}

\begin{figure}[tb]
  \centering

    \begin{subfigure}[b]{.5\textwidth}
       \centering
       \includegraphics[width=\linewidth]{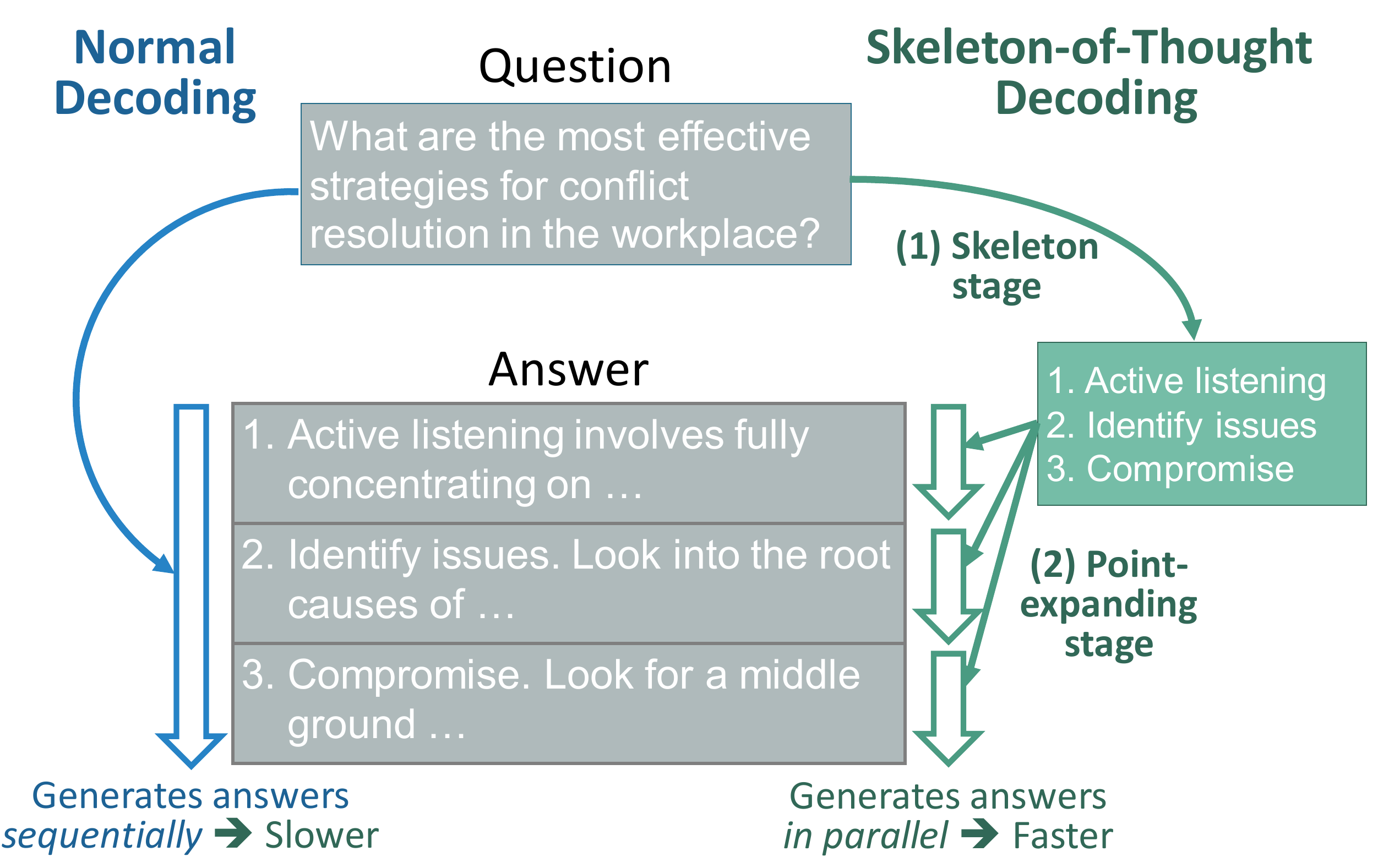}
     \end{subfigure}
     ~~~
     \begin{subfigure}[b]{.45\textwidth}
       	\centering
        \includegraphics[width=\linewidth]{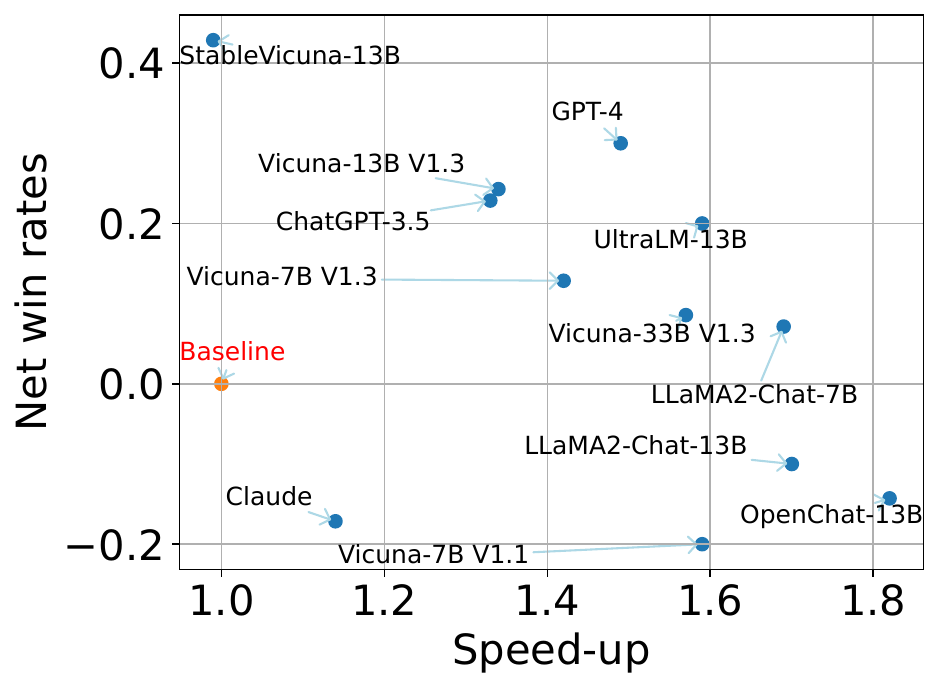}
     \end{subfigure}
  \caption{\textbf{Left:} An illustration of \method{} (\methodshort{}). Instead of producing answers sequentially, \methodshort{} produces different parts of answers \emph{in parallel}.
  In more detail, given the question, \methodshort{} first prompts the LLM to give out the skeleton, then conducts batched decoding or parallel API calls to expand multiple points in parallel, and finally aggregates the outputs to get the final answer. \textbf{Right:} The net win rates and speed-ups of \methodshort{} with router (\methodrshort{}) compared to \methodbase{} generation on \vicunadataset{}. The net win rate is the difference between the fraction of questions that \methodrshort{} has better and worse answers than \methodbase{} generation. The speed-up is the ratio between the latency of \methodbase{} and \methodrshort{} generation. $(1.0, 0.0)$ represents \methodbase{} generation. Higher is better on both axes. For most models, \methodrshort{} not only accelerates the generation but also improves the quality of the answers (evaluated with \fastchat{} metric \citep{zheng2023judging}). See \cref{sec:eval_algo,sec:router} for more details.} %
  \label{fig:sequential-vs-sot}
\end{figure}

Large language models (LLMs)~\citep{brown2020gpt3,touvron2023llama,du2022glm,openai2023gp4,zheng2023judging} have shown exceptional performance in natural language processing and %
chatbot systems.
However, the inference process of the state-of-the-art LLMs is slow, hindering their interactive use.
For example, it takes 22 seconds for Claude \citep{claude} (accessed through Slack API) and 43 seconds for \Remark{vicuna33B1.3} (a 33B LLaMA-based model, running locally on one NVIDIA A100 GPU) to answer the question in \cref{fig:sequential-vs-sot}.

We conclude three major causes of LLMs' slow inference: (1) A \emph{large model size} requires a large amount of memory, memory access, and computation. For example, the FP16 weights of 175B GPT-3 take 350GB memory, which means at least 5$\times$80GB A100 GPUs are needed to keep the model in GPU memory. Even with enough GPUs, the heavy memory access and computation slow down the inference. (2) %
The \emph{attention operation} in the prevailing transformer architecture is I/O bounded and has a quadratic memory and computation complexity in sequence length. (3) The \emph{sequential decoding} approach in inference generates tokens one by one. %
This approach introduces a significant inference latency since the generation of tokens cannot be parallelized. 
There is a bunch of literature addressing the first two axes: \emph{large model size}~\citep{xiao2022smoothquant,frantar2022gptq,lin2023awq,sheng2023flexgen,wang2021spatten} and \emph{attention operation}~\citep{kitaev2020reformer,wang2020linformer,dao2022flashattention,zaheer2020big,chen2023dynamic}. These works either compress/redesign the model~\citep{xiao2022smoothquant,frantar2022gptq,lin2023awq,kitaev2020reformer,wang2020linformer,dao2022flashattention,zaheer2020big} or redesign the serving system~\citep{sheng2023flexgen,chen2023dynamic} and hardware~\citep{wang2021spatten}.

In contrast to prior work, we tackle the third axis and question the common assumption that LLMs have to do fully sequential decoding. We show the feasibility of \textbf{parallel decoding of off-the-shelf LLMs \emph{without} any changes to their model, system, or hardware}. For instance, for the question in \cref{fig:sequential-vs-sot}, we can reduce the latency from 22 seconds to 12 seconds (1.83$\times$ speed-up) with Claude, %
and from 43 seconds to 16 seconds (2.69$\times$ speed-up) with Vicuna-33B V1.3 on an NVIDIA A100.

The idea stems from reflecting on how humans ourselves answer questions. 
Humans do \emph{not} always think about questions and write answers in a sequential fashion. In contrast, for many question types, we first derive the \emph{skeleton} according to some protocols and strategies, and then add evidence and details to explain each point. This is especially the case on %
occasions like offering consultancy, taking tests, writing papers, and so on. 
\revise{This intuition has our back to question the necessity of fully sequential decoding.}
In this paper, we propose \emph{\method{} (\methodshort{})}. Specifically, as shown in \cref{fig:sequential-vs-sot}, we guide the LLM to derive a skeleton first by itself. Based on the skeleton, the LLMs can complete each point \emph{in parallel} so that we get a speed-up. \methodshort{} can be utilized to accelerate both open-source models with batched decoding and API-based models with parallel API calls.

\revise{The current \methodshort{} is suitable for questions that require a long answer whose structure can be planned ahead, while not suitable for questions that require step-by-step reasoning or only need a short answer. Therefore,} to make the overall solution more practical, we design an extension, \methodshort{} with router (\methodrshort{}), which employs a router to only trigger \methodshort{} for suitable questions.

We test \methodshort{} on 12 recently released LLMs. Not only does \methodshort{} provide considerable speed-ups (up to 2.39$\times$), but it can also improve the answer quality in many cases (\cref{fig:sequential-vs-sot}).

Note that in contrast to existing model- and system-level efforts for inference efficiency, \methodshort{} takes a novel ``data-level'' pathway %
by letting the LLM organize its output content. This novel perspective is becoming feasible and is expected to grow in importance, owing to the evolving capabilities of state-of-the-art LLMs. We hope this work can stimulate more research in the realm of data-centric optimization~\citep{zha2023data,dcai2023} for efficiency.

The rest of the paper is organized as follows. We first introduce \methodshort{} in \cref{sec:method} and show its results in \cref{sec:exp}. %
Then, we expand on the \methodrshort{} extension in \cref{sec:router}.
\cref{sec:literature} positions \methodshort{} in the research ecosystem (expanded in \cref{sec:literature_complete}). Finally, we analyze the limitations and share outlooks of \methodshort{} in \cref{sec:limit-and-outlook}.

\section{\method{} (\methodshort{})}
\label{sec:method}

\subsection{Method}

\myparatightestn{Overview.}
Based on the intuition that humans usually think about and answer a question in an organized way, the core idea of this work is to guide the LLM itself to give a skeleton first and then write the overall answer parallelly instead of sequentially.
\cref{fig:sequential-vs-sot} illustrates how \methodshort{} produces the final answer to a user \concept{question} $q$. 

\myparaemphtightestn{(1) Skeleton stage.}
\methodshort{} first assembles a \concept{skeleton request}, $T^s(\mbox{question}=q)$, using the \concept{skeleton prompt template} $T^s$ (\cref{prompt:ts}, and \cref{prompt:ts_full} in \cref{app:implementation_details_prompt}) with the question $q$ as the parameter. 
The skeleton prompt template is written to guide the LLM to output a concise skeleton of the answer. Then, we extract the $B$ points from the \concept{skeleton response} $R^s$ of the LLM.

\myparaemphtightestn{(2) Point-expanding stage.} Based on the skeleton, we let the LLM expand on each point in parallel. Specifically, for the point with index $b$ and skeleton $R^s_b$, \methodshort{} uses $T^{pe}(\mbox{question}=q, \mbox{skeleton}=R^s, \mbox{point index}=b, \mbox{point skeleton}=R_{b}^s)$ as the \concept{point-expanding request} for the LLM, where $T^{pe}$ is the \concept{point-expanding prompt template} (\cref{prompt:tp}). Finally, after completing all points, we concatenate the point-expanding responses $\{R^{pe}_b\}_{b=1,\cdots,B}$ to get the \concept{final answer}. 

\begin{promptenv}{Skeleton Prompt Template $T^s$}{You're an organizer responsible for only giving the skeleton (not the full content) for answering the question. Provide the skeleton in a list of points (numbered 1., 2., 3., etc.) to answer the question. Instead of writing a full sentence, each skeleton point should be very short with only 3$\sim$5 words. Generally, the skeleton should have 3$\sim$10 points. Now, please provide the skeleton for the following question.\\\promptarg{question}\\Skeleton:}{1.}
  \label{prompt:ts}
\end{promptenv}

\begin{promptenv}{Point-Expanding Prompt Template $T^{pe}$}{You're responsible for continuing the writing of one and only one point in the overall answer to the following question.\\\\\promptarg{question}\\\\The skeleton of the answer is\\\\\promptarg{skeleton}\\\\Continue and only continue the writing of point \promptarg{point index}. Write it **very shortly** in 1$\sim$2 sentence and do not continue with other points!}{\promptarg{point index}. \promptarg{point skeleton}}
  \label{prompt:tp}
\end{promptenv}

\myparatightestn{Parallel point expanding.}
We conduct \emph{parallel} point-expanding so that \methodshort{} is able to achieve a speed-up than \methodbase{} decoding. 

\myparaemphtightestn{(1) For proprietary models with only API access}, we can issue multiple parallel API calls to get an end-to-end latency gain at the cost of an increased number of API requests and tokens. %

\myparaemphtightestn{(2) For open-source models that we can run locally}, we let them process the point-expanding requests as a batch (paddings are added to the left of the point-expanding requests). We explain below why this could achieve speed-ups.
A typical LLM generative process consists of two phases: (a) the \emph{prefilling} phase in which the prompt is parsed to generate the key-value cache for further use, and (b) the \emph{decoding} phase in which tokens are generated one by one in a sequential manner.
The decoding phase accounts for the majority of the end-to-end latency, especially when generating a long response. Note that the decoding phase is bottlenecked by weight loading instead of activation loading or computation.\footnote{This is true when the number of concurrent queries is small; see \cref{sec:limit-and-outlook} for discussion on other scenarios.}
Consequently, running LLM inference with increased batch sizes does not increase the per-token latency much.
Therefore, \methodshort{} 
allows us to decode roughly $B\times$ more tokens within the same amount of time if we parallelly decode $B$ points. 
See \cref{app:why_sot_reduce_latency_local} for the expanded discussions and the supporting experiments.
Please refer to \cref{app:implementation_details} for more implementation details. %

\section{\methodshort{} Evaluation}
\label{sec:exp}
\vspace{-0.1cm}

\myparatightestn{Datasets.} We evaluate \methodshort{} on two recent assistant-style datasets: (1) \vicunadataset{}~\citep{vicuna2023}, which contains 80 questions spanning nine categories, such as \textit{coding}, \textit{math}, \textit{writing}, \textit{roleplay}, and so on, and (2) \wizardlm{}~\citep{xu2023wizardlm}, which contains 218 questions spanning more categories and diverse difficulties. Due to space constraints, we only report \vicunadataset{} results in the main paper, and defer \wizardlm{} results to the \cref{app:efficiency,app:quality}. %

\myparatightestn{Models.} We test \methodshort{} on 12 models, including 9 open-source models and 3 API-based models. We obtain the weights of all the open-source models from Hugging Face. See \cref{app:model_details} for more details.

\subsection{Evaluation of Efficiency}
\label{sec:eval_eff}

\myparatightestn{API-based models.} We record the latency of every API call with \verb/start = time.time(); ...; elapsed_time = time.time() - start/, and add the latency of the skeleton API call and the slowest point-expanding API call as the \methodshort{} latency.

\myparatightestn{Open-source models.} All open-source models we currently evaluate are based on the \Remark{llama} 7B, 13B, or 33B architectures. Thus, to enable fast analysis, we first make a latency profiling table for each \Remark{llama} architecture on NVIDIA A100. The table contains the architecture's (1) latency for prefilling sequences of length 1 to 700 with different batch sizes (from 1 to 16), and (2) decoding one token with a context of length 1 to 1024 with different batch sizes (from 1 to 16). With these three latency profiling tables, given the number of points $B$, the token lengths of the requests and responses in the skeleton and point-expanding stages, we can quickly estimate the \methodshort{} latency by simply looking up entries in the tables and adding them up. See \cref{sec:app-profiling-estimate} for a more detailed description of how we conduct the profiling and estimate the latency. %

In addition to the above approach, 
we also compare the actual latency of \methodshort{} and \methodbase{} sequential generation (abbreviated as ``\methodbase{}'' in the following discussion)  in \cref{sec:app-actual-eff-test}.

The rest of this section shows the speed-ups of \methodshort{} on different models (\cref{sec:eval_eff_model}) and question categories (\cref{sec:eval_eff_category}). In addition, we also report the latency breakdown of \methodshort{} stages in \cref{app:latency_breakdown_stage_phase} and the \methodshort{} speed-ups on an RTX 3090 GPU in \cref{sec:app-3090}.

\subsubsection{Speed-up Breakdown: Models}
\label{sec:eval_eff_model}

\begin{figure*}[ht]
  \begin{center}
    \subfloat[Different models.]{
      \includegraphics[width=0.45\linewidth]{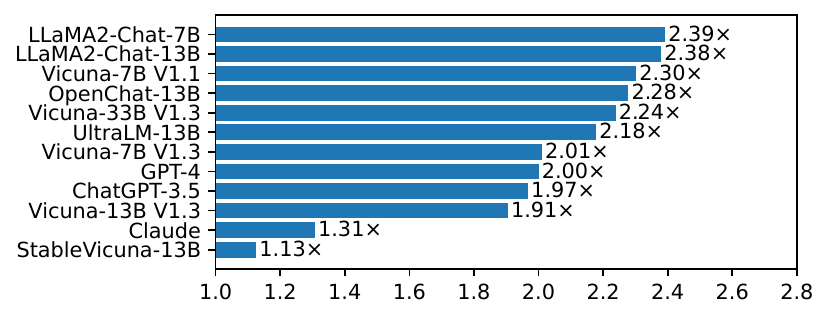}
      \label{fig:speed-up_model_average}
    }
    ~
    \subfloat[Different categories.]{
      \includegraphics[width=0.45\linewidth]{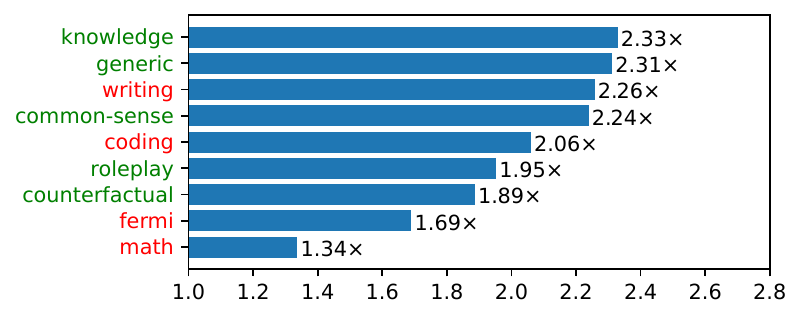}
      \label{fig:speed-up_category_average}
    }
    \caption{Average speed-ups of \methodshort{} on different models and question categories.}
  \end{center}
  \vspace{-0.5cm}
\end{figure*}

We investigate how \methodshort{} reduces the end-to-end latency on different models. \cref{fig:speed-up_model_average} shows the average speed-up for each model across all question categories. We can see that  \methodshort{} obtains a $>$2$\times$ speed-up (up to 2.39$\times$) on 8 out of 12 models. %

We report the detailed statistics about token lengths and numbers of points in \cref{fig:statistics}.
(1) In terms of \emph{the point number $B$} (\cref{fig:outline_num_points_average}), \Remark{llama2}, \Remark{vicuna7B1.1}, \Remark{vicuna7B1.3}, and \Remark{chatgpt} yield relatively fewer points ($<$6), while \Remark{gpt4} and \Remark{stablevicuna13B} generates the largest number of points on average ($\approx$9).
(2) Regarding \emph{the point-expanding response length}, \cref{fig:naive_response_average,fig:outline_response_2_average,fig:maxlen_div_len_average} show that the API-based models, \Remark{chatgpt}, \Remark{claude}, and \Remark{gpt4}, follow the point-expanding request better and generate shorter point-expanding responses than the open-source models.  %
One can also notice that \Remark{stablevicuna13B}'s longest point-expanding responses for many question categories %
can be as lengthy as the overall \methodbase{} answer, 
since it fails to adhere to the ``Write it **very shortly**'' instruction in the point-expanding request. Consequently, \methodshort{} cannot accelerate \Remark{stablevicuna13B} well.
(3) Regarding \emph{the length balance degree between point responses}, \cref{fig:balance_average}
shows that \Remark{llama2} and the API-based models generate more balanced point-expanding responses.
(4) As for \emph{the overall length of the final aggregated answer} (\cref{fig:totlen_div_len_average}), employing \methodshort{} on most models results in answers that are, on average, 1$\sim$2$\times$ longer than the \methodbase{} answer.

\subsubsection{Speed-up Breakdown: Question Categories}
\label{sec:eval_eff_category}

Here we investigate how \methodshort{} reduces the end-to-end latency for different question categories. \cref{fig:speed-up_category_average} shows the average speed-up for each question category across all models. The question categories for which \methodshort{} can provide high-quality answers are marked in green, and other categories are marked in red (see \cref{sec:eval_algo_category} for the answer quality evaluation). We can see that \methodshort{} can obtain speed-ups for all question categories. For the five question categories that \methodshort{} can provide high-quality answers (i.e., \textit{knowledge}, \textit{generic}, \textit{common-sense}, \textit{roleplay}, \textit{counterfactual}), \methodshort{} can speed up the overall answer generation process by 1.89$\times$ to 2.33$\times$ in the meantime.

\subsection{Evaluation of Answer Quality}
\label{sec:eval_algo}

In order to compare the answer quality of the \methodbase{} sequential generation (abbreviated as ``\methodbase{}'' in the following discussion) and \methodshort{} generation, we adopt two LLM-based evaluation frameworks: \fastchat{} \citep{zheng2023judging} and \llmzoo{} \citep{llm-zoo-2023}. The evaluation process is to present a question and a pair of answers (from \methodbase{} or \methodshort{} generation) to an LLM judge (\Remark{gpt4} in the main paper; see \cref{app:quality_gpt3.5} for the results evaluated using \Remark{chatgpt}) and ask for its preference.

Here are more details about the evaluation of the answer quality:

\myparaemphtightestn{(1) Detailed metrics.} \fastchat{} provides one metric for the general answer quality. In addition to a general metric, \llmzoo{} provides five detailed metrics on the answers' coherence, diversity, immersion, integrity, and relevance. 

\myparaemphtightestn{(2) Question categories.} \fastchat{} provides two special evaluation prompts for coding and math questions for more accurate evaluation, whereas \llmzoo{} does not. Following the implementation in \llmzoo{}, we exclude math and coding questions in all \llmzoo{} evaluation results.

\myparaemphtightestn{(3) Extentions to avoid evaluation bias.}
To avoid the potential bias from the order of the two answers presented to the LLM judge, we extend \fastchat{} and \llmzoo{} evaluation frameworks by running the evaluation twice with either ordering of the two answers. In either evaluation, a score of 1, 0, and -1 is assigned when \methodshort{} wins, ties, or loses, respectively. The final evaluation is that \methodshort{} wins/ties/loses when the sum of the two scores is positive/zero/negative. For example, if \methodshort{} wins in one evaluation and loses in the other evaluation, the result is ``tie''. If \methodshort{} wins (loses) in one evaluation and ties in the other, the result is ``win'' (``lose'').

\myparaemphtightestn{(4) Net win rates.} We further define net win rates to give a summarized view of the answer quality. Given the number of questions that \methodshort{} wins (\#win) and loses (\#lose), we define \emph{net win rates} as $\nicefrac{\text{\#win}-\text{\#lose}}{\text{total number of questions}}$.
0\% means that \methodshort{} performs competitively to the \methodbase{} baseline (wins and loses in the same number of questions). Higher values mean that \methodshort{} performs better. 

In the following sections, we first present the overall quality of \methodshort{} answers (\cref{sec:eval_algo_overall}), and then go into the details across different question categories (\cref{sec:eval_algo_category}), models (\cref{sec:eval_algo_model}), and metrics (\cref{sec:eval_algo_metric}).

\subsubsection{Overall Quality}
\label{sec:eval_algo_overall}

\begin{figure}[ht]
    \centering
    \includegraphics[width=0.75\linewidth]{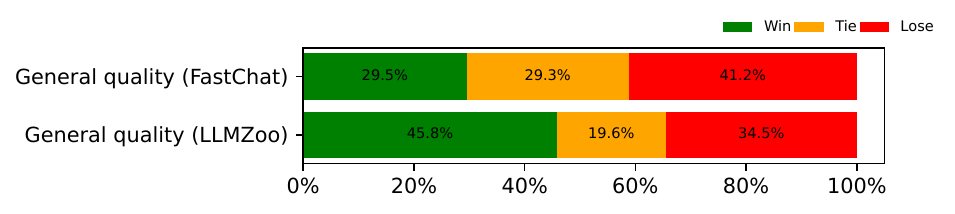}
    \vspace{-0.4cm}
    \caption{Win/tie/lose rates of \methodshort{} v.s. \methodbase{} generation using ``general'' metrics from \fastchat{} and \llmzoo{}. \methodshort{} performs better than or equal to \methodbase{} generation in around 60\% cases.}
    \label{fig:win_tie_lose_bar_GENERAL_gpt4}
    \vspace{-0.4cm}
\end{figure}

In \cref{fig:win_tie_lose_bar_GENERAL_gpt4}, we show the win/tie/lose rates (the percentage of the cases when \methodshort{} wins/ties/loses compared to \methodbase{} generation) across all models and questions using the two metrics from \fastchat{} and \llmzoo{} that capture the general quality of the answers. We notice a discrepancy between the two metrics on when \methodshort{} is strictly better than the baseline (45.8\% v.s. 29.5\%). Despite that, the two metrics agree that \methodshort{} is not worse than the baseline in around 60\% of the cases, and the win rates are close to the lose rates.
\emph{This result suggests that the answers of \methodshort{} maintain good quality of that of the normal generation.}

\subsubsection{Quality Breakdown: Models}
\label{sec:eval_algo_model}
\begin{figure}[ht]
    \centering
    \begin{subfigure}[b]{0.49\textwidth}
         \centering
         \includegraphics[width=\textwidth]{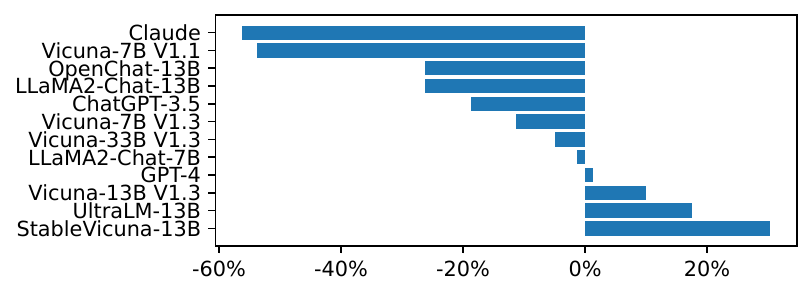}
         \caption{Metric: general quality (\fastchat{}).}
         \label{fig:net_win_rates_model_fastchat_gpt4}
     \end{subfigure}
     \hfill
     \begin{subfigure}[b]{0.49\textwidth}
         \centering
         \includegraphics[width=\textwidth]{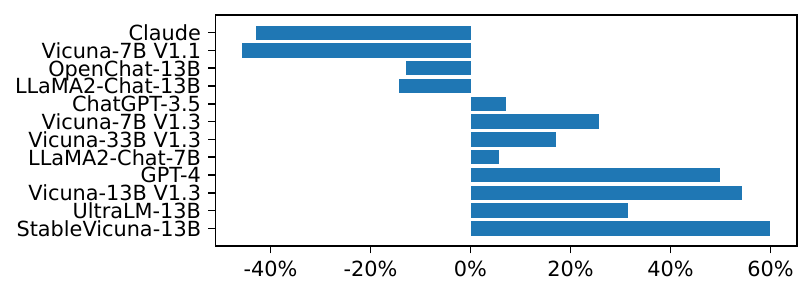}
         \caption{Metric: general quality (\llmzoo{}).}
         \label{fig:net_win_rates__llmzoo_gpt4}
       \end{subfigure}
       \vspace{-0.3cm}
     \caption{Net win rates of \methodshort{} on different models.}
     \label{fig:net_win_rates_model_gpt4}
     \vspace{-0.3cm}
\end{figure}

We compute net win rates on all models in \cref{fig:net_win_rates_model_gpt4}. Again, we see that the two general metrics from \fastchat{} and \llmzoo{} have different absolute values but similar rankings. In particular, both metrics agree that \Remark{openchat13B}, \Remark{vicuna7B1.1}, \Remark{claude}, \Remark{llamachat13B2} have \emph{low} net win rates, whereas \Remark{vicuna13B1.3}, \Remark{stablevicuna13B}, and \Remark{ultralm13B} have \emph{high} net win rates. 

We investigate the answers in \cref{sec:app-eval-quality-model-pattern}, and summarize the key takeaways as follows.
Some models have low \methodshort{} %
net win rates
as they cannot understand the skeleton and point-expanding prompts well. 
Some other models have low \methodshort{} %
net win rates
as their normal answers already have good quality, making it hard for \methodshort{} to beat them (e.g., \Remark{claude}). 
For models that are able to understand the \methodshort{} prompts and the normal answers are not good enough,
\methodshort{} can improve the answer quality. %
We expect that further improving SoT prompts or
fine-tuning the models can make it easier for LLMs to understand the skeleton and point-expanding
prompts and ultimately result in better answer quality.

\subsubsection{Quality Breakdown: Question Categories}
\label{sec:eval_algo_category}
\begin{figure}[th]
    \centering
    \begin{subfigure}[b]{0.49\textwidth}
         \centering
         \includegraphics[width=\textwidth]{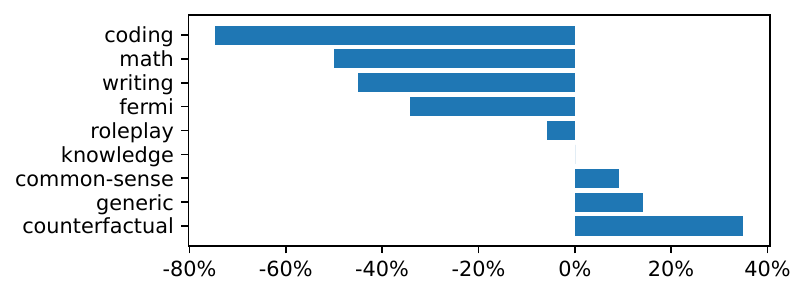}
         \caption{Metric: general quality (\fastchat{}).}
         \label{fig:net_win_rates_category_fastchat_gpt4}
     \end{subfigure}
     \hfill
     \begin{subfigure}[b]{0.49\textwidth}
         \centering
         \includegraphics[width=\textwidth]{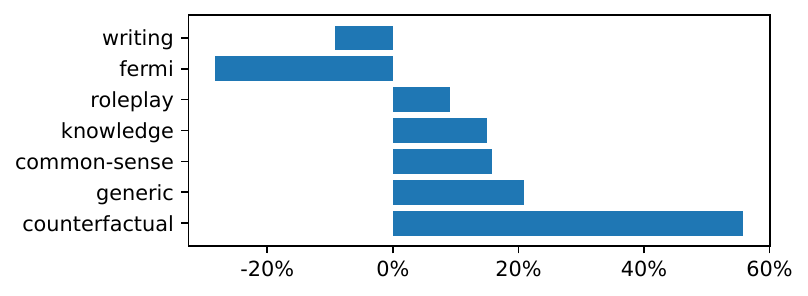}
         \caption{Metric: general quality (\llmzoo{}).}
         \label{fig:net_win_rates_category_llmzoo_gpt4}
       \end{subfigure}
       \vspace{-0.3cm}
     \caption{Net win rates of \methodshort{} on different question categories.}
     \label{fig:net_win_rates_category_gpt4}
\end{figure}

We compute net win rates on all question categories
in \cref{fig:net_win_rates_category_gpt4}. Similar to \cref{fig:win_tie_lose_bar_GENERAL_gpt4}, we see that \llmzoo{} tends to be more optimistic about the quality of \methodshort{} than \fastchat{}. 
Nevertheless, the conclusions are consistent: \methodshort{} performs relatively \emph{well} on \textit{generic}, \textit{common-sense}, \textit{knowledge}, \textit{roleplay}, and \textit{counterfactual}, and relatively \emph{poorly} on \textit{writing}, \textit{fermi}, \textit{math}, and \textit{coding}. %

We investigate the answers in \cref{sec:app-eval-quality-category-pattern}, and summarize the key takeaways as follows.
\methodshort{} performs well when the question can be answered in several points whose details can be expanded independently. This includes a wide range of real-world questions.
On the other hand, it is fundamentally challenging to apply \methodshort{} on questions that require step-by-step thinking, in which the latter steps require the details from the earlier steps, such as math questions. 
To make \methodshort{} general across broader question categories, one promising pathway is to enable \methodshort{} to adaptively fall back to normal generation, which we explore in \cref{sec:router}. 
Interestingly, our results suggest that some LLMs are already able to do that occasionally without special prompting or tuning (see \cref{sec:app-eval-quality-category-pattern}).

\subsubsection{Quality Breakdown: Metrics}
\label{sec:eval_algo_metric}
\begin{figure}[ht]
    \centering
    \vspace{-0.4cm}
    \includegraphics[width=0.55\linewidth]{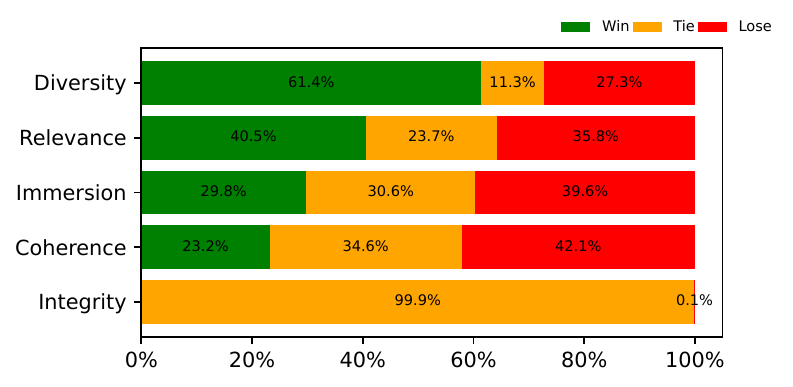}
    \vspace{-0.4cm}
    \caption{Win/tie/lose rates of \methodshort{} v.s. \methodbase{} generations using metrics from \llmzoo{}. \methodshort{} performs well on diversity and relevance, and relatively worse on coherence and immersion.}
    \label{fig:win_tie_lose_bar_DETAILS_gpt4}
    \vspace{-0.1cm}
\end{figure}

In \cref{fig:win_tie_lose_bar_DETAILS_gpt4}, we show more detailed metrics from \llmzoo{} to reveal in which aspects \methodshort{} can improve or hurt the answer quality. On average, we can see that \methodshort{} improves the diversity and relevance while hurting the immersion and coherence.

Through answer investigation (\cref{sec:app-eval-quality-metric-pattern}), we summarize the key takeaways as follows.
The skeleton stage of \methodshort{} explicitly require LLMs to discuss the answers from multiple aspects without filler words. This improves the diversity and relevance of the answers. %
As for coherence and immersion, \methodshort{} is not worse than the \methodbase{} generation around 60\% of the time. One future direction is to improve the \methodshort{} prompts or pipeline so that the answers can be better in more metrics.

\section{\methodshort{} with Router (\methodrshort{}): Adapatively Triggering \methodshort{}}
\label{sec:router}
\vspace{-0.2cm}
In \cref{sec:exp}, we see that \methodshort{} provides considerable speed-ups while maintaining (or even improving) answer quality for many question types. However, the biggest limitation is that \methodshort{} is not suitable for questions that require step-by-step reasoning (\cref{sec:eval_algo_category}). 
Towards pushing the practical adoption of \methodshort{}, we explore the possibility of \emph{adaptively triggering \methodshort{}} only when it is suitable. To achieve that, we propose a \emph{router} module that decides if \methodshort{} should be applied for the user request, and then call either \methodshort{} or \methodbase{} decoding accordingly. This paradigm aligns with the recent trends of composing multiple models to solve complicated tasks \citep{Chase_LangChain_2022,shen2023hugginggpt}. To implement the router, we explore two options: LLM prompting as the router (no model training is needed) (\cref{sec:fallback_prompting}), and trained \roberta{} as the router (\cref{sec:trained_roberta}). The evaluation is provided in \cref{sec:router_exp}.

\vspace{-0.1cm}
\subsection{Prompting Router}
\label{sec:fallback_prompting}

We directly ask an LLM if the question is suitable for \methodshort{}. More specifically, we ask the LLM if the desired answer is in a list of independent points (see \cref{app:prompting_router} for the prompt). If the answer is yes, we will use \methodshort{}; otherwise, we will use \methodbase{} generation (i.e., directly feeding the question to the LLM). We employ \Remark{gpt4} as the LLM router given its strong capability. %

\subsection{Trained Router}
\label{sec:trained_roberta}

While leveraging GPT-4 as the router obviates the need for model training, its performance remains sensitive to prompt design. %
Therefore, we approach the problem as a sequence classification task by fine-tuning a small language model as the router. Specifically, we annotate the LIMA dataset \citep{zhou2023lima} as the training set to train a \roberta{} model \citep{liu2019roberta}, which has only 120M parameters.
Details about the annotation and training can be found in \cref{app:annotation_process_router,app:training_details_roberta}.

\subsection{\methodrshort{} Evaluation}
\label{sec:router_exp}
We compare \methodshort{} and \methodrshort{} under the same evaluation setup in \cref{sec:exp}. Besides the prompting and trained routers, we also consider a ``human router'' where we manually judge whether \methodshort{} should be applied for each question. This serves as a benchmark for comparison.

\vspace{-0.3cm}
\subsubsection{Evaluation of Efficiency}
\vspace{-0.2cm}

\begin{figure*}[ht]
    \vspace{-0.3cm}
    \begin{minipage}[t]{0.48\linewidth}
      \centering
      \includegraphics[width=1.0\linewidth]{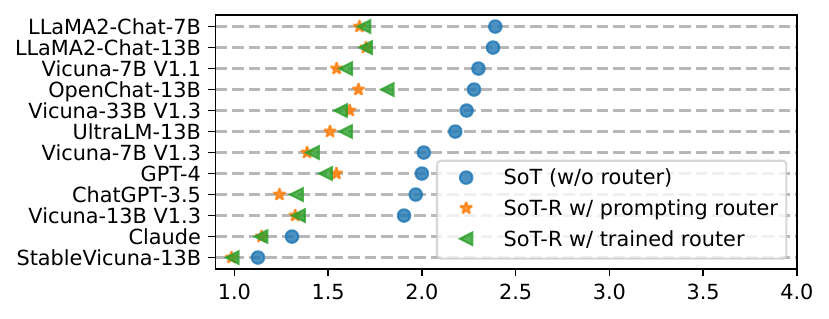}
        \vspace{-0.7cm}
      \caption{Speed-ups of \methodshort{} and \methodrshort{} on different models across all question categories of the \vicunadataset{} dataset.}
      \label{fig:efficiency_router_vicuna}
    \end{minipage}
    \hfill
    \begin{minipage}[t]{0.48\linewidth}
      \centering
      \includegraphics[width=1.0\linewidth]{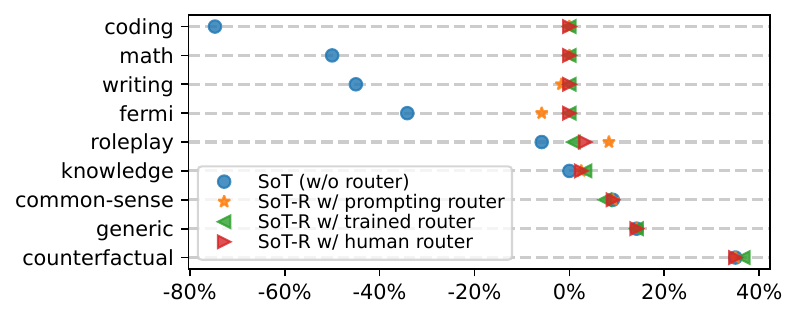}
        \vspace{-0.7cm}
      \caption{Net win rates of \methodshort{} and \methodrshort{} on different question categories of the \vicunadataset{} dataset (evaluated with the \fastchat{} metrics).}
      \label{fig:quality_router_gpt4_fastchat}
    \end{minipage}
\end{figure*}

\cref{fig:efficiency_router_vicuna} shows the speed-ups of \methodshort{} and \methodrshort{} for different models on \vicunadataset{} (see \cref{sec:more-effiency-sotr} for results on the \wizardlm{} dataset). We can see that: (1) As expected, \methodrshort{} obtains lower speed-ups than \methodshort{}, since \methodshort{} is not triggered for some questions and the router induces a small latency overhead. Nevertheless, \methodrshort{} can still benefit most models with $>$1$\times$ speed-ups. (2) %
\methodrshort{} with the trained router obtains slightly higher speed-ups for 7 out of 12 models on \vicunadataset{}, while \methodrshort{} with the prompting router obtains higher speed-ups for all models on WizardLM (\cref{fig:efficiency_router_wizardlm_model}).

\vspace{-0.2cm}
\subsubsection{Evaluation of Answer Quality}
\vspace{-0.1cm}
\cref{fig:quality_router_gpt4_fastchat} shows the net win rates (averaged across all models) of \methodshort{} and \methodrshort{} on \vicunadataset{} with the \fastchat{} metrics (see \cref{app:quality_router} for results of the \wizardlm{} dataset and \llmzoo{} metrics). We can see that: (1) \methodrshort{} significantly improves the answer quality on questions where \methodshort{} is not suitable (e.g., \textit{coding}, \textit{math}, \textit{writing}, \textit{fermi}) by falling back to \methodbase{} decoding. At the same time, \methodrshort{} maintains answer quality improvements on questions where \methodshort{} is good at. (2) The trained router performs similar to (on \vicunadataset{}) or better than (on \wizardlm{}; see \cref{app:quality_router}) the prompting router. This accords with our intuition in \cref{sec:trained_roberta}. (3) The prompting and trained routers could even surpass human router (e.g., on roleplay questions; see more examples on \wizardlm{} in \cref{app:quality_router}).

We discuss the consistency across three routers in \cref{app:router_consistency}. The primary takeaways include: (1) on \vicunadataset{}, there is a notable consistency among all three routers, and (2) on \wizardlm{}, greater discrepancies emerge, with the trained router showing higher alignment with human annotations.

\vspace{-0.2cm}
\section{\methodshort{} In the Context of Literature}
\label{sec:literature}
\vspace{-0.2cm}
This section positions \methodshort{} in related work to reveal how \methodshort{} (1) is connected to, (2) is different from, and (3) can harness the power of other methods. See \cref{sec:literature_complete} for the expanded discussion.

\myparatightestn{Efficient LLM methods at model and system levels.}
At the model level, prior work proposes efficient architectures, including dynamic mixture-of-experts~\citep{lepikhin2021gshard}, low-complexity attention~\citep{kitaev2020reformer}, and multi-query attention~\citep{shazeer2019fast}. However, they usually require a significant re-training cost. In contrast, compression methods require a smaller amount of fine-tuning cost by reducing the complexity of pre-trained LLMs,  %
such as quantization~\citep{frantar2022gptq} and weight or activation sparsification~\citep{semi_first,zaheer2020big}. %

At the system level, prior work (1) optimizes the computational graph~\citep{dao2022flashattention}, (2) optimizes the assignment and scheduling of computational graph on devices~\citep{sheng2023flexgen}, or (3) designs batching or caching mechanisms for serving multiple users~\citep{fang2021turbotransformers}. These techniques address the large memory access and footprint posed by the vast model scale and attention mechanism, and mainly aim at enhancing the throughput rather than the end-to-end latency. As \methodshort{} trades off throughput for end-to-end latency, \textit{\methodshort{} can 
make
these throughput-oriented techniques %
help with
end-to-end latency}. This interesting synergy offers opportunities for achieving better trade-offs between latency and throughput in future serving systems.

\textit{In contrast to model- and system-level techniques, SoT is a data-level technique in a new ``content co-organization for efficiency'' paradigm}.
See \cref{sec:limit-and-outlook} for more discussions. %

\myparatightestn{Efficient LLM methods through parallel generation.}
Some prior work also addresses
the sequential decoding issues. 
Speculative decoding (SD) methods~\citep{stern2018blockwise} employ smaller models %
to generate some consecutive tokens sequentially and apply the target LLMs to verify them parallelly. Non-autoregressive generation (NAG) methods~\citep{gu2018nonautoregressive,xiao2023survey} sample and refine consecutive tokens parallelly, often with the support of a modified and tuned model.

Relying on either assisting models or special models and sampling schemes, SD and NAG methods conduct \textit{parallel verification or sampling and refinement of consecutive tokens}. In contrast, \methodshort{} prompts the LLM \emph{itself} to plan the contents in a way that permits \textit{the parallel generation of tokens in different segments}, by exploiting the emerging instruction-following and planning ability of LLMs. %

\myparatightestn{Prompting methods for LLMs.}
Recent years have witnessed the emergence of the ``pre-train, prompt, and predict'' paradigm, %
which has shown promise in enhancing LLMs' quality in math and commonsense reasoning~\citep{wei2022chain,kojima2022large,wang2022selfconsistency,chen2022program} and planning for multi-modality tasks~\citep{shen2023hugginggpt,zhu2023ghost}. %
Instead of focusing on answer quality, \textit{\methodshort{} is a first attempt at exploiting the power of prompting to improve efficiency}.

\vspace{-0.3cm}
\section{Limitations, Future Work, and Open Questions}
\label{sec:limit-and-outlook}
\vspace{-0.3cm}

\myparatightestn{Answer quality evaluation.} Our answer quality evaluation is far from perfect due to the limited prompt set, the potential bias of GPT-4 judges, and the inherent difficulty of evaluating LLM generations. %
Currently, we did not conduct human evaluation since it is easy for a human to tell whether an answer is generated with \methodshort{} due to its distinctive pattern, which might cause evaluation bias.

\myparatightestn{Eliciting or improving LLMs' ability.}
\cref{sec:eval_algo_metric} demonstrates \methodshort{}'s potential of enhancing answer quality. It is part of a broader trend in recent research, exemplified by work including CoT~\citep{kojima2022large,wei2022chain}, ToT~\citep{yao2023tree}, and ReAct~\citep{yao2022react}, which collectively affirm the notion that \emph{explicitly articulating the thought process in language can elicit high-quality answers from LLMs}.
These findings resemble human thinking: rather than relying solely on the first intuition or purely sequential thinking, 
we often document step-by-step reasoning or thought organization to attain 
high-quality answers. This intriguing parallel prompts us to explore further how we can draw from the human thinking process to facilitate more effective and efficient AI.

For instance, \methodshort{} currently 
ignores the dependencies between points. A conceptually better way is to organize the points as \emph{Graph-of-Thoughts}, where the edges represent the dependencies, and each point is decoded conditioned on the contents of its ancestor points.
In addition, instead of complying with a \emph{static} graph, we expect the need %
of having
\emph{dynamic Graph-of-Thoughts}, where the high-level thought structure is adjusted dynamically by LLMs themselves. This could potentially combine the efficiency and global thinking advantages of \methodshort{} with the logical reasoning and impromptu thinking strengths of methods like CoT~\citep{kojima2022large,wei2022chain}. Notably, a contemporary work~\citep{besta2023graph} has attempted to design Graph-of-Thoughts to elicit reasoning.
Furthermore, %
it is interesting to explore how the \methodshort{} answers can be used to fine-tune LLMs to generate more structured answers in a self-improving way~\citep{zelikman2022star,huang2022large}.

\myparatightestn{Efficiency and overhead of \methodshort{} in different scenarios.}
Serving systems commonly adopt batch processing to handle concurrent queries. %
This raises a concern of whether \methodshort{} may hurt %
serving throughput due to parallel requests.
(1) When there is an unsaturated number of concurrent queries,
\methodshort{} can effectively 
reduce latency and
enhance GPU utilization. %
Example scenarios include (a) Edge-side applications with a single user; (b) Centralized services during periods with unsaturated user requests and underutilized computing capacity.
It is %
interesting %
to study the appropriate \methodshort{} triggering conditions based on system workloads.
(2) When there is a saturated number of concurrent queries, %
\methodshort{} is still useful for improving answer quality.
However, in this case, it is important to consider the computation overhead from \methodshort{}. %
We delve into this concern in \cref{sec:app-token-overhead}.

For API-based models, a notable concern arises regarding the increased number of prefilling tokens (\cref{sec:app-token-overhead}). Given that many APIs charge token usage, \methodshort{} may lead to higher costs. To address this, one can %
use prompt tuning to design shorter \methodshort{} prompts\revise{~\citep{jiang-etal-2023-llmlingua}}. %

\myparatightestn{Data-centric efficiency optimization.}
While data-centric engineering for improving answer \emph{quality}~\citep{zha2023data,dcai2023} is gaining popularity,
its potential for \emph{inference efficiency} is not explored yet. \methodshort{} is the first attempt. %
As LLM capabilities and the amount of LLM-generated data are growing rapidly, data-centric techniques could become more useful in the future.
\revise{To pave the way towards that, there are a lot to explore. For example, the acceleration ratio of \methodshort{} depends on the SoT prompt, the model, and the question, and thus not as predictable and controllable as model- or system-level techniques, which might hinder the practical adoption.}
We look forward to future work
to unlock the full potential of data-centric efficiency optimization.

\section*{Acknowledgements}
We thank Sergey Yekhanin (Microsoft Research), and Tianji Wu (Infinigence AI) for their support and suggestions on the work. We thank Tianyu Fu for many initial discussions on the idea. We thank Ke Hong and Genghan Zhang for their discussions about profiling. We thank Yue Wu for the help on the \Remark{claude} scripts. We thank Da Yu, Chulin Xie, and Saiqian Zhang for their suggestions on revising the first version of the paper. We thank Rui Hu, Cheng Cheng, Jack Jin, Zhoutong Ye, Mingze Sun, Jun Yan, Zhi Zhang, Yuxuan Tong, Nianhui Guo, and Andrea Santilli for their suggestions on revising the second version of the paper.
We thank Chris Stetkiewicz, Amanda Melfi, and Amber Tingle from Microsoft for their suggestions and help on writing.
We thank the anonymous reviewers for their insightful questions and suggestions.

\bibliographystyle{iclr2024_conference}
\bibliography{top,iclr2024_conference}

\clearpage
\appendix
\part{Appendix}
\parttoc
\clearpage

\section{Model Details}
\label{app:model_details}
\cref{tab:model} summarizes the models on which we evaluate \methodshort{}. We use \Remark{gpt4} in the main paper and \Remark{chatgpt} in \cref{app:quality_gpt3.5} as the judge in \fastchat{} and \llmzoo{} evaluation.
\begin{table}[H]
  \centering
  \caption{Models evaluated with \methodshort{}. All the open-source models are fine-tuned from \Remark{llama} models.}
    \label{tab:model}
    \begin{tabular}{c|c|c|c}
        \toprule
        Access & Model Name & Institution & Released Date \\\midrule
        \multirow{8}{*}{Open-Source} &
        \Remark{llamachat7B2} \citep{touvron2023llama2} & Meta \& Microsoft & 2023/07\\
        &
        \Remark{llamachat13B2} \citep{touvron2023llama2} & Meta \& Microsoft & 2023/07\\
        &\Remark{openchat13B}~\citep{openllms23} & Tsinghua & 2023/07\\ %
        & \Remark{vicuna7B1.3}~\citep{vicuna2023} & LMSYS & 2023/06\\
        & \Remark{vicuna13B1.3}~\citep{vicuna2023} & LMSYS & 2023/06\\
        & \Remark{vicuna33B1.3}~\citep{vicuna2023} & LMSYS & 2023/06\\
        & \Remark{stablevicuna13B}~\citep{stablevicuna2023}& CarperAI &2023/05 \\ %
        & \Remark{ultralm13B}~\citep{ding2023enhancing} & OpenBMB \& Tsinghua & 2023/05\\
        & \Remark{vicuna7B1.1}~\citep{vicuna2023} & LMSYS & 2023/03 \\\midrule
        \multirow{2}{*}{API-Based} & \Remark{claude}~\citep{claude} & Anthropic & 2023/05\\
        & \Remark{chatgpt} & OpenAI & 2022/11\\
        & \Remark{gpt4} & OpenAI & 2023/03\\
        \bottomrule
    \end{tabular}
  \end{table}
  
  \cref{tab:model_detail} shows sources of the models we use in the paper.
\begin{table}[H]
  \centering
      \caption{The Hugging Face or API endpoints of the models.}
    \label{tab:model_detail}
    \begin{tabular}{c|c|c}
        \toprule
        Access & Model Name & Hugging Face or API Endpoints \\\midrule
        \multirow{8}{*}{Open-Source} &
        \Remark{llamachat7B2} \citep{touvron2023llama2} & meta-llama/Llama-2-7b-chat-hf \\
        &
        \Remark{llamachat13B2} \citep{touvron2023llama2} & meta-llama/Llama-2-13b-chat-hf \\
        &\Remark{openchat13B}~\citep{openllms23} & openchat/openchat \\
        & \Remark{vicuna7B1.3}~\citep{vicuna2023} & lmsys/vicuna-7b-v1.3 \\
        & \Remark{vicuna13B1.3}~\citep{vicuna2023} & lmsys/vicuna-13b-v1.3\\
        & \Remark{vicuna33B1.3}~\citep{vicuna2023} & lmsys/vicuna-33b-v1.3\\
        & \Remark{stablevicuna13B}~\citep{stablevicuna2023}& CarperAI/stable-vicuna-13b-delta\tablefootnote{For convenience, we use the non-official endpoint TheBloke/stable-vicuna-13B-HF and TheBloke/UltraLM-13B-fp16 to get merged weights.\label{footnote:thebloke}} \\ 
        & \Remark{ultralm13B}~\citep{ding2023enhancing} & openbmb/UltraLM-13b\footref{footnote:thebloke} \\
        & \Remark{vicuna7B1.1}~\citep{vicuna2023} & lmsys/vicuna-7b-delta-v1.1 \\\midrule
        \multirow{2}{*}{API-Based} & \Remark{claude}~\citep{claude} & Claude extension on Slack\tablefootnote{\url{https://www.anthropic.com/claude-in-slack}} \\
        & \Remark{chatgpt} & Azure OpenAI, gpt-35-turbo 0301 version\tablefootnote{\url{https://azure.microsoft.com/en-us/products/ai-services/openai-service}}\\
        & \Remark{gpt4} & OpenAI, gpt-4-0613 version\\
        \bottomrule
    \end{tabular}
\end{table}

\section{Implementation Details of \method{}}
\label{app:implementation_details}

\subsection{Prompt}
\label{app:implementation_details_prompt}

The skeleton prompt is shown in \cref{prompt:ts,prompt:ts_full} and the point-expanding prompt is shown in \cref{prompt:tp}.

\begin{promptenv}{Skeleton Prompt Template $T^s$ (with Two-Shot Demonstrations)}{You're an organizer responsible for only giving the skeleton (not the full content) for answering the question. Provide the skeleton in a list of points (numbered 1., 2., 3., etc.) to answer the question. Instead of writing a full sentence, each skeleton point should be very short with only 3$\sim$5 words. Generally, the skeleton should have 3$\sim$10 points.\\\\Question:\\What are the typical types of Chinese dishes?\\Skeleton:\\1. Dumplings.\\2. Noodles.\\3. Dim Sum.\\4. Hot Pot.\\5. Wonton.\\6. Ma Po Tofu.\\7. Char Siu.\\8. Fried Rice.\\\\Question:\\What are some practical tips for individuals to reduce their carbon emissions?\\Skeleton:\\1. Energy conservation.\\2. Efficient transportation. \\3. Home energy efficiency. \\4. Reduce water consumption. \\5. Sustainable diet. \\6. Sustainable travel.\\\\Now, please provide the skeleton for the following question.\\\promptarg{question}\\Skeleton:}{1.}
  \label{prompt:ts_full}
\end{promptenv}

\myparatightestn{Skeleton prompt template.} 
In order to make the output skeleton short and in a consistent format for the good of efficiency and ease of point extraction, the skeleton prompt template (1) describes the task precisely, %
and (2) provides a partial answer ``1.'' for the LLM to continue writing. 
The skeleton responses are in the desired format in most cases. Therefore, we can use a simple regular expression \verb/(\d+)\.\s?([\s\S]+?)(?=\n|\n*$)/ to extract point indexes and point skeletons from the skeleton response.

We find that \Remark{gpt4} can work well without the two demonstrations in the skeleton prompt. Therefore, we do not include the two demonstrations for \Remark{gpt4} (\cref{prompt:ts}). For all other models, the two demonstrations are included, as shown in \cref{prompt:ts_full}.

\myparatightestn{Point-expanding prompt template.}
It describes the point-expanding task and provides a partial answer. We also provide instructions ``Write it **very shortly** in 1$\sim$2 sentence'' so that the LLMs keep the answers concise. Unlike the skeleton prompt template, we find that demonstrations are not necessary to get reasonable results.

We find that \Remark{claude} and \Remark{gpt4} follows the instruction ``Write it **very shortly** in 1$\sim$2 sentence and do not continue with other points!'' in \cref{prompt:tp} very well, so that the answers are very short. Therefore, we delete ``**very shortly**'' from the prompt template in \Remark{claude} and \Remark{gpt4}. %

\paragraph{Partial answer.} In the \cref{prompt:ts,prompt:tp}, we provide partial answers so that LLMs can follow the desired response format better.

We can put the partial answer at the end of the prompt for the open-source models to continue writing. An implementation detail is that different open-source models have different conversation templates (i.e., different ways to combine user and assistant messages into one string). 
For example, \Remark{vicuna}~\citep{vicuna2023} uses the string ``USER:'' and `` ASSISTANT:'' for the placeholder ``\textbf{[User:]}'' and ``\textbf{[Role]}'' in the \cref{prompt:ts,prompt:tp}, respectively, while \Remark{ultralm}~\citep{ding2023enhancing} uses ``User:'' and ``\textlangle/s\textrangle Assistant:''. We build our open-source model experiments with the help of the FastChat codebase~\citep{zheng2023judging}, in which the conversation templates of many models are already  handled correctly. We implement the conversation templates of \Remark{openchat13B}, \Remark{stablevicuna13B}, and \Remark{ultralm13B} according to their official guides and codes.

For \Remark{chatgpt}, we provide partial answers as a last message in the chat history from the assistant. Note that it is not a documented approach. We find it works well in most cases, in that \Remark{chatgpt} continues the texts from the provided partial answer. However, in some rare cases, \Remark{chatgpt} repeats the provided partial answers.

For \Remark{claude} over Slack, there is no obvious way to give the API a partial answer. We resort to modifying the prompt template slightly by adding \begin{center}\textit{Please start your answer from ``\promptarg{partial answer}'' and do not output other things before that} \end{center} at the end. We find that \Remark{claude} understands and obeys it well. For \Remark{gpt4}, we also take this approach.

\paragraph{System Message.}
We do not include the system message in the prompts for open-source models except \Remark{llama2}.

The partial answer, ``**very shortly**'', and the 2-shot demonstrations discussed above are the only differences between the prompts we used across all models and all evaluations.

\subsection{Supporting Multi-Round Conversation}
To use \methodshort{} in a multi-round conversation, we can just put the question and the final aggregated answer in the history, removing all the \methodshort{} prompts. In this way, using \methodshort{} in one conversation round will not introduce additional prefill cost in future rounds.

\section{Implementation Details of \method{} with Router}
\label{app:implementation_details_router}
\subsection{Prompting Router}
\label{app:prompting_router}
We use \cref{prompt:router} for querying \Remark{gpt4} as the router. If the answer is ``A'' (i.e., the question can be answered in a list of independent points), we will use \methodshort{}. Otherwise, if the answer is ``B'' (i.e., the answer is in a list of points but they depend on each other) or ``C'' (i.e., the answer should \emph{not} be in a list of points), \methodshort{} is not suitable and we will fall back to \methodbase{} decoding. 

\begin{promptenv}{LLM Prompting as the Router}{Question: \promptarg{question}\\\\How would you like to answer the question?\\A. Organize the answer as a list of points or perspectives (in the format of 1., 2., 3., etc.), and the points or perspectives can be answered independently without referring to the contents of the previous points.\\B. Organize the answer as a list of points or perspectives (in the format of 1., 2., 3., etc.), and the contents of later points or perspectives cannot be answered independently without referring to the contents of the previous ones.\\C. Do not organize the answer as a list of points or perspectives.\\\\Just say A, B, or C. Do not explain. Do not provide an answer to the question.}{}
  \label{prompt:router}
\end{promptenv}

\subsection{Trained Router}

We tackle the routing problem as a sequence classification task. We first annotate the LIMA training set~\citep{zhou2023lima}, and then fine-tune a \roberta{} model~\citep{liu2019roberta} using the labeled data. Finally, we apply the tuned \roberta{} as the router on \vicunadataset{} and \wizardlm{}. We detail the steps in the following.

\subsubsection{Annotation Process}
\label{app:annotation_process_router}
In the classification task, a label of 1 (positive) indicates that this question can be answered with \methodshort{}, while a label of 0 (negative) suggests that using the normal generation mode is more suitable. We annotate the LIMA training set, which consists of 1,030 Q\&As sourced from three community webpages: Stack Exchange, wikiHow, and the Pushshift Reddit.
We also annotate the \vicunadataset{} and \wizardlm{} datasets for evaluation.

We use GPT-4 to assist the annotation process. Specifically, we present each question to GPT-4 and analyze its answer to determine whether \methodshort{} can be triggered for this question. We assign a positive label to a question if GPT-4's response meets two criteria: (1) it contains a list of points that can be expanded in parallel, (2) each point provides sufficient details (i.e., the point-expanding response is not too short), which will enable \methodshort{} to achieve a speed-up. Two of the paper's authors conduct the annotation process independently, and discuss the inconsistent annotations to decide the final label.

\subsubsection{Training Details} 
\label{app:training_details_roberta}
We use \texttt{roberta-base} with 120M parameters as the router model. The finetuning is conducted using the AdamW optimizer~\citep{loshchilov2017decoupled} with a weight decay of 0.01. The learning rate undergoes a warm-up phase during the first 1\% of iterations to 5e-5 and then decays linearly. We train the model for 2 epochs using a batch size of 32. Input sequences are either padded or truncated to achieve a consistent length of 512 tokens.

In the application of \methodshort{}, false positives (\methodshort{} is incorrectly triggered when it should not be, resulting in degraded answer quality) are of more significant concern than false negatives (the router misses a potential \methodshort{} trigger, resulting in a reduced speed-up). Thus, to mitigate false positives, we employ the Tversky loss~\citep{wang2023dice} with parameters $\alpha=0.7$ and $\beta=0.3$, which penalizes false positives more heavily than false negatives. We also incorporate label smoothing~\citep{szegedy2016rethinking} with a factor of $\epsilon=0.2$. Overall, the entire fine-tuning process is efficient, completing in 2 minutes on an NVIDIA A100 GPU.

\subsection{Router Consistency}
\label{app:router_consistency}
We present the confusion matrices for the three routers to illustrate their consistency. The results on \vicunadataset{} and \wizardlm{} are shown in \cref{tab:confusion_matrices_vicuna,tab:confusion_matrices_wizardlm}, respectively.

On \vicunadataset{}, we can observe a notable level of agreement among the three routers. Compared with the GPT-4-prompting router, the trained router exhibits a slightly higher number of false negatives w.r.t. the human annotations. Conversely, on \wizardlm{}, given the intricate answer structure and the presence of many ambiguous cases, the routers show significant discrepancies. Specifically, the GPT-4 router produces many false positives, which pose adverse affects on the answer quality (see \cref{app:quality_router}). The \roberta{} router aligns more closely with the human annotations.

\begin{table}[]
  \centering
  \caption{Router confusion matrices on the \vicunadataset{} dataset. \textbf{Left:} Rows are human annotations (H) and columns are the GPT-4 router (G). \textbf{Middle:} Rows are human annotations (H) and columns are the \roberta{} router (R). \textbf{Right:} Rows are the GPT-4 router (G) and columns are the \roberta{} router (R).}
\label{tab:confusion_matrices_vicuna}
\begin{tabular}{c|c|c|}
\multicolumn{1}{c}{} & \multicolumn{1}{c}{G0} & \multicolumn{1}{c}{G1} \\
\cline{2-3}
H0 & 38 & 5 \\
\cline{2-3}
H1 & 0 & 37 \\
\cline{2-3}
\end{tabular}
\quad
\begin{tabular}{c|c|c|}
\multicolumn{1}{c}{} & \multicolumn{1}{c}{R0} & \multicolumn{1}{c}{R1} \\
\cline{2-3}
H0 & 37 & 6 \\
\cline{2-3}
H1 & 5 & 32 \\
\cline{2-3}
\end{tabular}
\quad
\begin{tabular}{c|c|c|}
\multicolumn{1}{c}{} & \multicolumn{1}{c}{R0} & \multicolumn{1}{c}{R1} \\
\cline{2-3}
G0 & 34 & 4 \\
\cline{2-3}
G1 & 8 & 34 \\
\cline{2-3}
\end{tabular}

\caption{Router confusion matrices on the \wizardlm{} dataset. \textbf{Left:} Rows are human annotations (H) and columns are the GPT-4 router (G). \textbf{Middle:} Rows are human annotations (H) and columns are the \roberta{} router (R). \textbf{Right:} Rows are the GPT-4 router (G) and columns are the \roberta{} router (R).}
\label{tab:confusion_matrices_wizardlm}
\begin{tabular}{c|c|c|}
\multicolumn{1}{c}{} & \multicolumn{1}{c}{G0} & \multicolumn{1}{c}{G1} \\
\cline{2-3}
H0 & 94 & 66 \\
\cline{2-3}
H1 & 3 & 55 \\
\cline{2-3}
\end{tabular}
\quad
\begin{tabular}{c|c|c|}
\multicolumn{1}{c}{} & \multicolumn{1}{c}{R0} & \multicolumn{1}{c}{R1} \\
\cline{2-3}
H0 & 135 & 25 \\
\cline{2-3}
H1 & 31 & 27 \\
\cline{2-3}
\end{tabular}
\quad
\begin{tabular}{c|c|c|}
\multicolumn{1}{c}{} & \multicolumn{1}{c}{R0} & \multicolumn{1}{c}{R1} \\
\cline{2-3}
G0 & 93 & 4 \\
\cline{2-3}
G1 & 73 & 48 \\
\cline{2-3}
\end{tabular}
\end{table}

\subsection{Concurrent execution for \methodrshort{}}
\begin{figure*}[]
  \begin{center}
    \includegraphics[width=\textwidth]{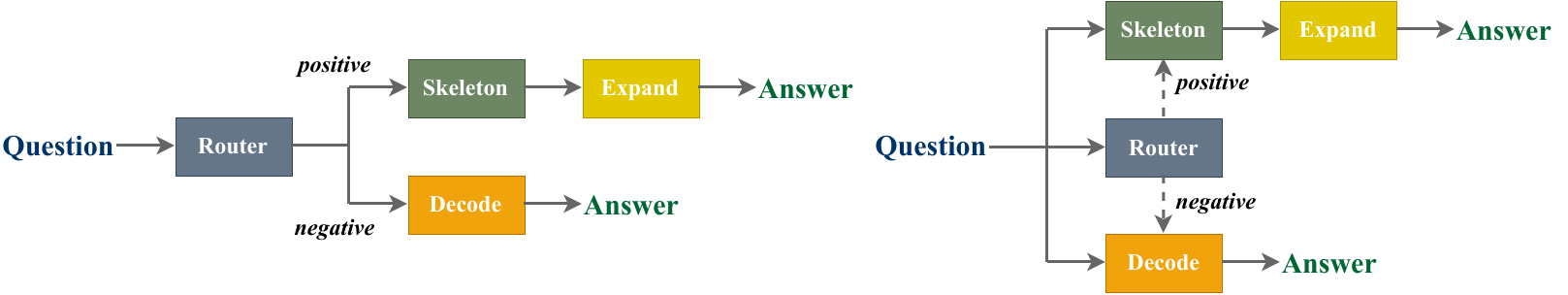}
    \caption{\textbf{Left:} The \methodrshort{} pipeline. \textbf{Right:} A possible approach to further reduce latency at the cost of token overhead.}
    \label{fig:router_pipeline}
  \end{center}
\end{figure*}

In \methodrshort{}, the router serves as an additional stage that extends the two-stage \methodshort{} pipeline, as illustrated in \cref{fig:router_pipeline}. To push the limit of latency optimization, we can run the router, normal generation, and \methodshort{} generation concurrently. Once the router makes a decision, one of the normal and \methodshort{} generation processes can be aborted. However, this approach will increase the token overhead. Therefore, we did not employ this approach in this work and leave it to future work. %

\section{\methodshort{} In the Context of Literature (Expanded)}
\label{sec:literature_complete}

\subsection{Efficient LLMs}

Extensive research has been dedicated to enhancing the throughput and latency of LLM inference. We first discuss model-level architecture design or compression techniques. 
These techniques change the model and can benefit both the latency and throughput but require finetuning to retain the model quality.
Then, we discuss system-level efforts that optimize the computational graph or the assignment and scheduling of the computational graph on computation and storage devices.
Most system-level efforts accelerate the prefilling phase or focus on improving the throughput.
Finally, we discuss some research efforts that share a similar motivation to ours, namely, addressing the efficiency issue of sequential decoding.

\paragraph{Model-level optimization.} 
Considerable architectural design efforts have emerged to (1) improve the scalability w.r.t. model size by introducing mixture-of-expert inference~\citep{lepikhin2021gshard,fedus2022switch}, (2) address the quadratic complexity w.r.t. input size of attention by designing new attention mechanisms~\citep{kitaev2020reformer,wang2020linformer}, (3) reduce the memory access and footprint of attention by using multi-query attention~\citep{shazeer2019fast}, and so on. However, these methods usually require a substantial re-training cost. The model compression techniques require a smaller amount of fine-tuning by reducing the model complexity of a pre-trained LLM from certain aspects~\citep{ganesh2021compressing}. Representative techniques include quantization~\citep{xiao2022smoothquant,frantar2022gptq,lin2023awq}, the static or dynamic pruning of weights, activation, and attention~\citep{semi_first,zaheer2020big,wang2021spatten,chen2023dynamic}, and so on.

Zooming out from LLM compression to the whole field of model compression, we can see that model co-design or compression for efficiency has received tremendous attention in the past few years and has grown into large research fields, such as pruning~\citep{han2015deep,group-lasso}, quantization~\citep{krishnamoorthi2018quantizing}, factorization~\citep{denton2014exploiting}, and neural architecture search~\citep{zoph2016neural,elsken2019neural,cai2019once}. %
\textit{Different from the model co-design paradigm, SoT is in a ``\textbf{content co-organization for efficiency}'' paradigm for improving the LLM efficiency}.
Along with the growth in the LLM capabilities and amount of LLM-generated data, data-level techniques could become important tools in the efficient LLM toolbox.

\paragraph{System-level optimization.} %
In the realm of lossless acceleration, considerable efforts have been devoted to addressing the I/O-bound nature of LLMs on modern hardware platforms~\citep{dao2022flashattention}. Numerous studies~\citep{dao2022flashattention,zhai2022bytetransformer,ivanov2021data,fastertransformer} have focused on adjusting the computational graph by fusing and implementing operations in an I/O-friendly way. As a representative method, FlashAttention~\citep{dao2022flashattention} fuses all operations of one attention into one GPU kernel with spatially tiled computation to reduce the off-chip I/O of the attention map. While FlashAttention can effectively accelerate training and the prefilling phase of inference, it cannot accelerate the decoding phase much (when the batch size is small), as it is the I/O of weights rather than activation or attention map that bottlenecks the decoding phase. For example, when the context length is 64, decoding one token using \Remark{llama}-7B needs to load each of the 7B parameters from the off-chip HBM onto the GPU chip at least once, but only transferring about 20M (0.02B) activation values between the off-chip HBM and GPU chip.

In order to satisfy Service Level Objectives, serving systems focus on improving the serving throughput under latency constraints. To this end, serving systems~\citep{fang2021turbotransformers,triton,tfserving}
pack multiple queries together into a batch to improve the hardware utilization.
The batching technique has proven highly effective in enhancing throughput, leading to the development of various variants. For example, some work designs methods to decide which queries to batch together~\citep{fang2021turbotransformers,zhou2022pets}, while others selectively batch parts of the model to enable fine-grained iteration-level batching~\citep{yu2022orca} or multi-task batching~\citep{zhou2022pets}.
Various model parallelism~\citep{lu2017flexflow,huang2019gpipe,narayanan2019pipedream,rajbhandari2020zero,narayanan2021memory,li2021terapipe,zheng2022alpa} and offloading~\citep{ren2021zerooffload,sheng2023flexgen} techniques have been proposed to maximize the throughput of LLM training or inference. In a nutshell, given the computational graph and device configurations, these techniques optimize the split, assignment, and scheduling of computations, storage, and communications on devices.
In addition to the model parallelism and batching techniques, an efficient memory management mechanism for LLM workloads is also an essential feature in the serving systems~\citep{kwon2023efficient,sensetime2023lightllm,sensetime2023openppl}.

To sum up, these system-level techniques mainly help with the throughput in training and batched inference. %
They can be used by \methodshort{} to improve the throughput of the batched decoding of multiple segments. %
This means that \textit{\methodshort{} can harness the power of these throughput-oriented techniques and make them help with the end-to-end latency}, offering a new dimension for better trading off latency and throughput in future serving systems.

Another parallelism perspective to position \methodshort{} is that \textit{\methodshort{} guides the LLM to adjust the sequential workload to become ``inter-content'' parallelizable}, which differs from the parallelism levels in existing serving systems, including inter-instance~\citep{krizhevsky2014dp,rajbhandari2020zero}, inter-operation~\citep{huang2019gpipe,narayanan2019pipedream,narayanan2021memory}, intra-operation~\citep{xu2021gspmd}, and inter-token~\citep{li2021terapipe}.
It may be worthwhile to explore \textit{the integration of \methodshort{} into serving systems to maximize the hardware utilization}. %

\paragraph{Decoding optimization.} %
One bottleneck for the end-to-end latency lies in the autoregressive decoding phase, where tokens must be generated one by one. Due to the dependency between tokens, the computation of different tokens cannot be parallelized, causing severe under-utilization of GPU. In order to improve the end-to-end decoding latency of a given LLM, speculative decoding methods~\citep{stern2018blockwise,leviathan2022fast,chen2023accelerating,hugging2023assisted,sun2023spectr,miao2023specinfer} propose to use cheaper approaches to generate short candidate token sequences, for example, by sequentially decoding with an assisting model much smaller than the given LLM. Then, they use the LLM to parallelly verify the candidates and keep the prefix sequence that matches the LLM's verification results.

Another line of work that shares the motivation of addressing the autoregressive efficiency issue is non-autoregressive generation (NAG) methods~\citep{gu2018nonautoregressive,xiao2023survey}. 
NAG methods sample consecutive tokens parallelly, often with the aid of a modified and tuned model. To maintain the 
answer quality, instead of sampling for one iteration, many NAG methods refine the output parallelly for multiple iterations~\citep{xiao2023survey,santilli2023accelerating}.

To summarize, the speculative decoding methods use assisting models for \textit{letting the LLM conduct parallel verification of consecutive tokens}, and the NAG methods rely on specially designed models, training schemes, or sampling schemes for \textit{the parallel sampling and refinement of consecutive tokens}. In contrast, \methodshort{} prompts the LLM itself to plan the contents in a way that permits \textit{the parallel generation of multiple tokens in different segments}. \methodshort{} exploits the emerging instruction-following and planning ability of SoTA LLMs rather than relying on specially designed modeling, sampling, and training schemes. This is different from all existing work that targets the autoregressive efficiency issue.

\subsection{Prompting Methods for LLMs}

In recent years, the ``pre-train, prompt, and predict'' paradigm has emerged~\citep{liu2023pre}, which designs prompts comprising task descriptions and (optionally) a few demonstrations to guide pre-trained LLMs in generating answers for a wide range of downstream tasks. %
Researchers found that instruction-tuned LLMs~\citep{brown2020gpt3,wei2021flan,ouyang2022instructgpt,chung2022scaling,alpaca2023} possess a strong ability to (1) generalize to new tasks thanks to the diverse natural language descriptions encountered during instruction tuning, and (2) learn in-context using a few demonstrations without weight tuning.

In virtue of these abilities, the field has been manually engineering~\citep{brown2020gpt3,kojima2022large,shen2023hugginggpt,li2023camel}, automatic searching~\citep{shin2020autoprompt}, or continuously tuning~\citep{li2021prefix,lester2021power} the prompts for uncovering the capabilities of LLMs on downstream tasks. There are a bunch of prompting methods that improves the reasoning performance of LLMs by designing thinking flows mimicking human reasoning: (1) mimicking the step-by-step or compositional thinking structure~\citep{wei2022chain,kojima2022large,press2022selfask,yao2023tree,besta2023graph,zhang2023cumulative}, (2) designing multiple reasoning paths and their aggregation~\citep{wang2022selfconsistency,yao2023tree,li2023diverse}, and (3) using tools for calculation and information retrieval~\citep{chen2022program,yao2022react,schick2023toolformer}. As a representative example, the Chain-of-Thought prompts largely improve the performance on tasks that require logical reasoning by simply providing a ``Let's think step by step''~\citep{kojima2022large} instruction or a few demonstrations~\citep{wei2022chain}. Another topic that arises quite a surge of interests is to prompt LLMs to help finish complex multi-modality task~\citep{shen2023hugginggpt,zhu2023ghost}. For example, HuggingGPT~\citep{shen2023hugginggpt} design prompts to guide the LLM to generate structural JSON for the orchestration of multi-model execution to finish complex tasks.

To summarize, the large literature on prompting methods has been aiming at uncovering different capabilities of LLM and improving the answer quality %
on different downstream tasks. In contrast, \textit{\methodshort{} is a first attempt at exploiting the power of prompting to improve efficiency}.

\subsection{Hierarchical Text Generation}

\revise{\methodshort{} can be regarded as being ``hierarchical'' since it has high-level answer structure planning. Prior studies in hierarchical text generation~\citep{li2015hierarchical,shao2019long,puduppully2019data,fan2018hierarchical} all focus on enhancing the answer quality, including improving the long-range coherence, relevance to the topic, or reducing redundancy. These methods craft hierarchical neural architectures that contain different modules to model high-level (sentence-level or document-level) and low-level (word-level) dependencies~\citep{li2015hierarchical,shao2019long,fan2018hierarchical}. They still employ sequential word-by-word generation without parallelization between sentences.}

\revise{Note that the sentence-level representations in previous work~\citep{li2015hierarchical,shao2019long} are ``implicit'' latent variables instead of ``explicit'' language descriptions. Some previous studies~\citep{shao2019long,puduppully2019data} train a dedicated planning module to execute explicit content planning in advance. Nevertheless, these methods all conduct ``closed-form'' planning that only reorders and groups the input keywords, rather than producing ``free-form'' plans on ``what to say'' and ``how to say''. All the hierarchical architectures and planning modules require training or even special data processing~\citep{puduppully2019data}.}

\revise{To summarize, in terms of the objective, the primary focus of \methodshort{} -- efficient generation -- is different from previous hierarchical text generation literature. In terms of the methodology, instead of designing new hierarchical architectures or planning modules, \methodshort{} exploits the emerging planning and instruction-following abilities of LLMs to do explicit (which means the plan is described by interpretable language) and free-form planning. This allows \methodshort{} to be applied to off-the-shelf LLMs for producing structured answers.}

\revise{As the hierarchical text generation literature focuses on enhancing answer quality, they could provide inspiration for future expansions of \methodshort{} to generate high-quality answers for broader types of questions.}

\section{Efficiency Analysis}
\label{app:why_sot_reduce_latency_local}

This section gives a detailed explanation on why \methodshort{} can reduce the overall decoding latency with the same computational resource for local models.

The vanilla approach processes only one question and decodes the answers sequentially, whereas \methodshort{} processes multiple point-expanding requests and the answers in a batch.
We focus on the following question: ``Compared to processing only one sequence, how much peak memory overhead and latency increase will be brought by processing a batch of sequences?''

\begin{table}[tb]
  \centering
    \caption{The latency and average GPU performance of the prefilling and decoding phases when inferencing LLMs. The prefilling token length is 128, the decoding token length is 64, and the batch size is 1. The test is run on one NVIDIA A100 GPU.}
  \label{tab:inf-latency}
  \begin{tabular}{ccc}
    \toprule
     Model   & Prefill/Decode Latency (ms) & Prefill/Decode GPU Perf. (TFLOPS) \\ \midrule
    \Remark{llama}-7B  &  40  / 2735  & 43 / 0.31 \\
    \Remark{llama}-13B &  54  / 3725  & 62 / 0.44 \\
    \Remark{llama}-33B &  100 / 5506  & 85 / 0.75 \\
    \bottomrule
  \end{tabular}
\end{table}

A typical LLM generative process consists of two phases: (1) the prefilling phase in which the prompt is parsed to generate the key-value cache for further use, and (2) the decoding phase in which tokens are generated one by one in a sequential manner. The decoding phase accounts for the majority of the end-to-end latency, especially when generating a long response. As shown in \cref{tab:inf-latency}, when running Vicuna-7B on NVIDIA A100-80G, the actual computing performance is only 0.31 TFLOPS (0.1\% utilization) in the decoding phase, compared to 43 TFLOPS (13.8\% utilization) during prefilling. The utilization is calculated with respect to the FP16\footnote{All of our experiments are run with FP16 inference.} tensor core peak performance -- 312 TFLOPS for NVIDIA-A100. As a result, the latency of decoding only one token is comparable to that of prefilling 128 tokens (40ms).
This huge gap in actual computing performance and thereby the latency arises from the fact that all LLM weights need to be loaded onto the GPU chip at least once only for decoding one token, so the decoding is heavily bottlenecked by the I/O of weights and the GPU computation units cannot be well utilized.

\begin{figure*}[tb]
  \begin{center}
    \subfloat[Latency (ms)]{
      \includegraphics[width=0.315\linewidth]{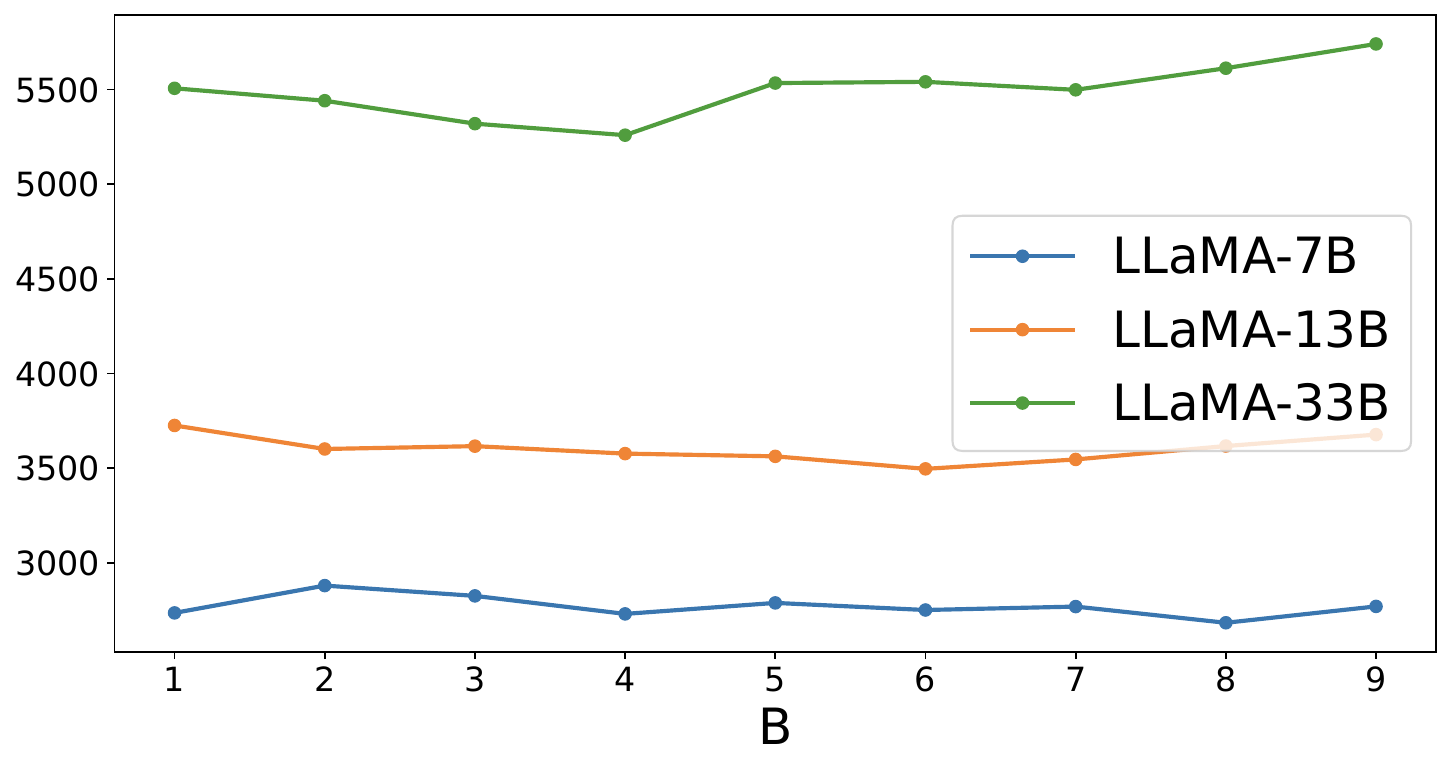}
      \label{fig:efficiency-batch-vs-single-latency}
    }
    \subfloat[Actual GPU Perf. (TFLOPS)]{
      \includegraphics[width=0.306\linewidth]{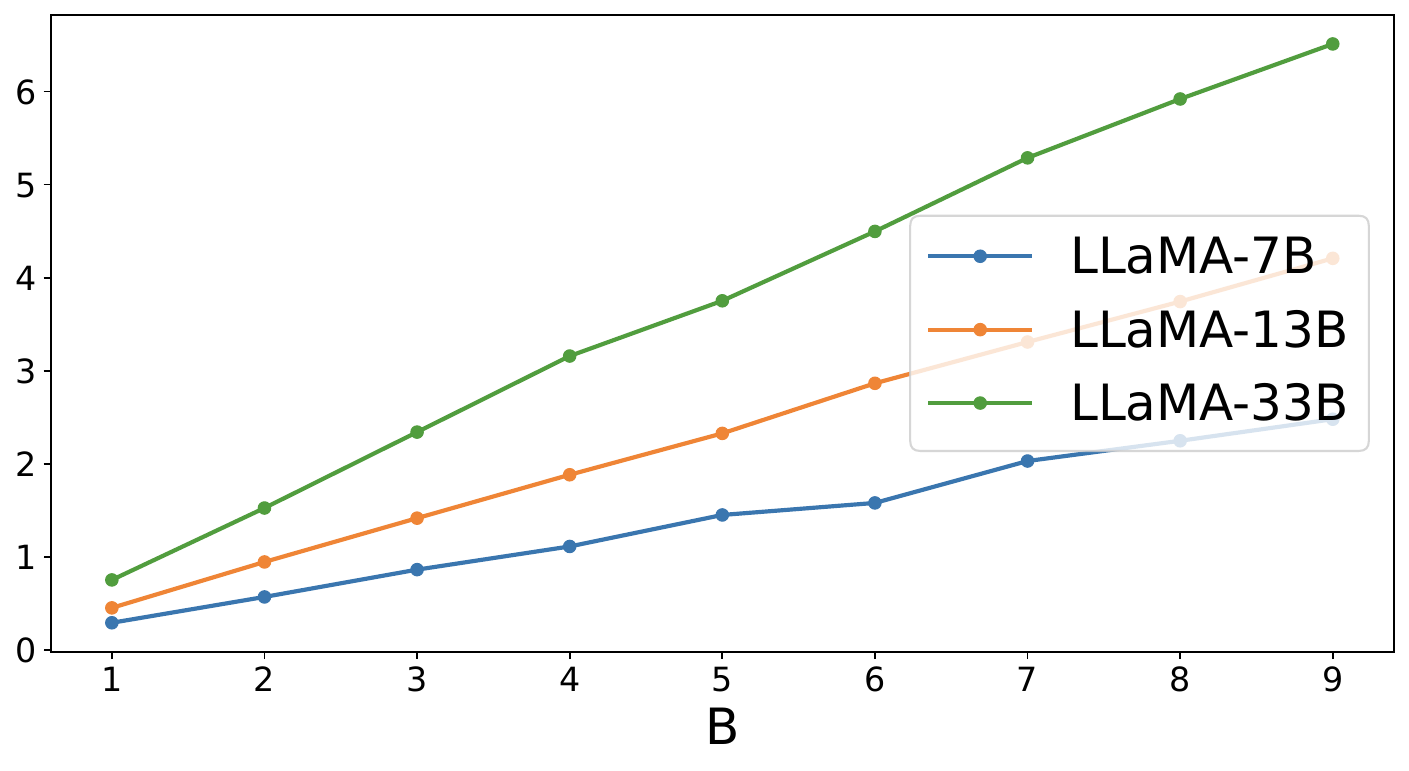}
      \label{fig:efficiency-batch-vs-single-perf}
    }
    \subfloat[Peak Memory (GB)]{
      \includegraphics[width=0.31\linewidth]{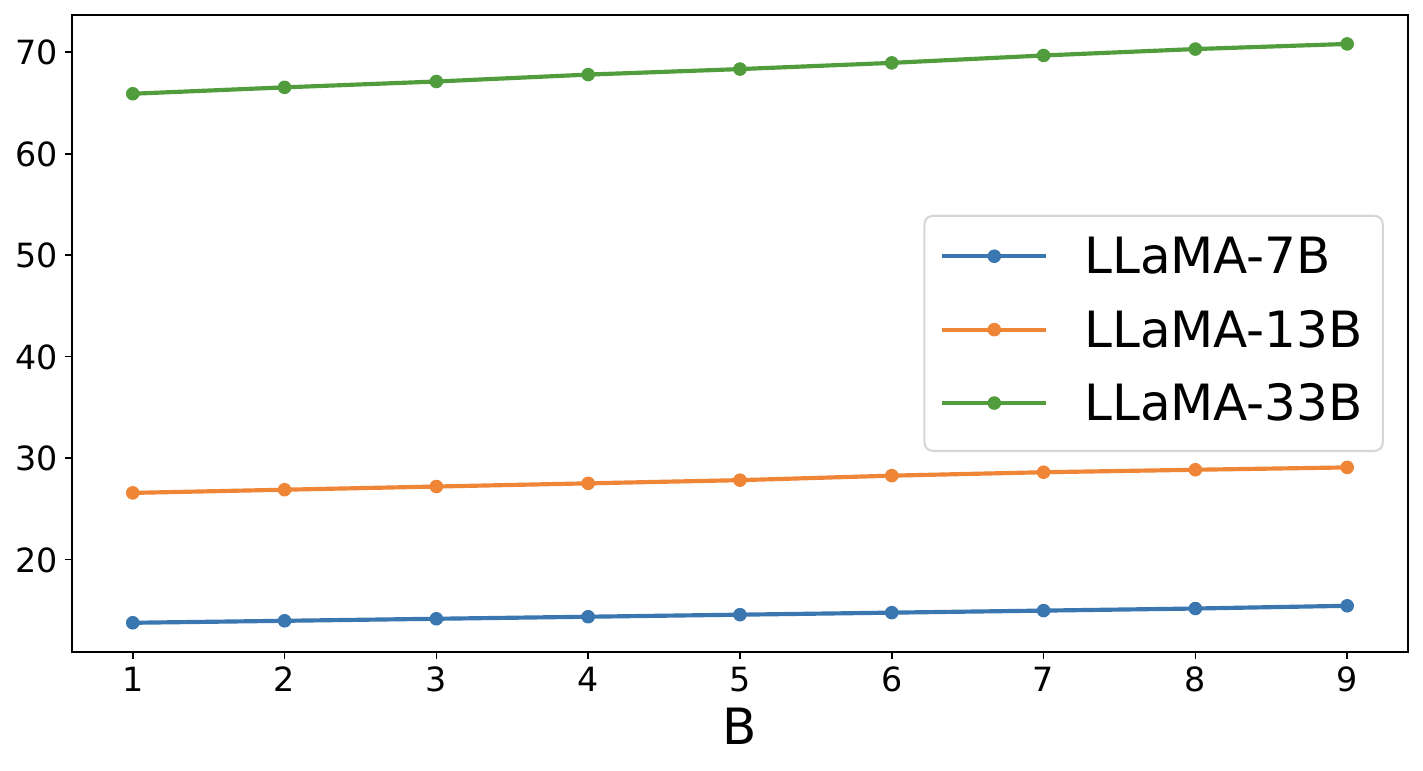}
      \label{fig:efficiency-batch-vs-single-mem}
    }
    \caption{The trends of latency, average GPU performance of decoding one token, and peak memory with respect to the batch size $B$ of sequences. The prefilling token length is 128, and the decoding token length is 64. The test is run on one NVIDIA A100 GPU.}
    \label{fig:efficiency-batch-vs-single}
  \end{center}
\end{figure*}

When conducting batched decoding, as the sequence batch size $B$ increases, the latency of decoding one token for each sequence stays roughly the same (\cref{fig:efficiency-batch-vs-single-latency}), as the amount of LLM weights that needs to be loaded onto the chip does not change.
As a result, the GPU computation utilization ($\frac{\text{Actual GPU Performance}}{\text{Peak GPU Performance}}$) increases almost linearly as $B$ increases (\cref{fig:efficiency-batch-vs-single-perf}). In other words, for generating a final answer of length $N$, if we cut the answer into $B$ segments of length $N/B$ and decode them as a batch, we can get a $B \times$ decoding speed-up compared to sequential decoding. Nevertheless, in practice, as prefilling longer requests brings some overhead, and the lengths of the $B$ segments could be imbalanced, the actual speed-up of the batched point-expanding stage compared with the original prefilling and sequential decoding process is smaller than $B$. %

As for the peak memory overhead, the amount of LLM weights can be one to two orders of magnitude larger than that of all the intermediate activations as long as the prefilling token length is not too large, not to mention that most activations do not need to be saved for back-propagation during inference. Therefore, the LLM weights account for the majority of the memory footprint in our test cases. Consequently, as shown in \cref{fig:efficiency-batch-vs-single-mem}, the peak memory overhead due to the increasing size of the KV cache and activation grows at a slow pace as the batch size $B$ increases. Thanks to the small peak memory overhead, in all of our experiments, we managed to use one GPU to run \methodshort{} without seeking help from other peak memory optimization techniques (e.g., quantization~\citep{frantar2022gptq,lin2023awq}, offloading~\citep{sheng2023flexgen}).

\section{Efficiency Profiling}
\label{sec:app-profiling-estimate}

We run the profiling on the target GPU (NVIDIA A100-80G and NVIDIA RTX 3090) with CUDA 11.7, using the Hugging Face transformer library 4.28.1 and PyTorch 2.0.1. The host of A100-80G has an Intel Xeon Platinum 8358P CPU and 1T memory. The host of RTX 3090 has an Intel Xeon Gold 6246R CPU and 512G memory.

\paragraph{Latency profiling and estimation.} For the decoding phase, we denote $t_{B}^{D}(k)$ as the latency of batched decoding the $k+1$-th token with batch size $B$, where the superscript $D$ stands for ``decode''. For each batch size $B=1,\cdots,16$ and each context length $k=1,\cdots,1024$, we use \verb/torch.cuda.Event/ to record the latency of decoding one token. We run each decoding three times continuously and take their geometric mean as $\{t_{B}^{D}(k)\}_{k=1,\cdots,1024; B=1,\cdots,16}$.
For the prefilling phase, we profile the latency of batched prefilling the inputs with token length $k$ in $\mbox{range}(1,700,10)$ and batch size $B=1,\cdots,16$, and denote it as $t_{B}^{P}(k)$, where the superscript $P$ stands for ``prefill''. We run each test seven times continuously, regard the first two times as the warmup tests, and take the geometric mean of the last five times as $\{t_{B}^{P}(k)\}_{k=1,11,\cdots,691; B=1,\cdots,16}$. Once we get the latency profiling table, given a request with $l_i$ tokens and the decoding batch size $B$, the latency of generating $l_o$ tokens can be estimated as:
\begin{equation}
    T(l_i, l_o, B) = \tilde{t}_{B}^{P}(l_i) + \sum_{k=l_i}^{l_i+l_o-1} t_{B}^{D}(k),
  \end{equation}
  where the subscripts $i$ and $o$ stand for ``input'' and ``output''. Note that we only test the prefilling latency every ten token lengths (i.e., $1, 11, 21, \cdots$) for fast profiling and estimate $\tilde{t}_{B}^{P}(l_i)$ by $t_{B}^{P}(\lfloor\frac{l_i}{10}\rfloor \times 10 + 1)$.

The \methodshort{} decoding process consists of two stages: the skeleton stage and the point-expanding stage. Denoting the token length of the skeleton request and skeleton response as $l_i^{s}$ and $l_o^{s}$, the token length of the longest point-expanding request and the longest point-expanding response as $l_i^{pe}$ and $l_o^{pe}$, the number of the points as $B$, we can compute the latency of the skeleton and point-expanding stages as:
\begin{align}
    L^{s}(l_i^{s}, l_o^{s}) &= T(l_i^{s}, l_o^{s}, 1), \\
    L^{pe}(l_i^{pe}, l_o^{pe}, B) &= T(l_i^{pe}, l_o^{pe}, B).
\end{align}

Using the latency profiling table, we can further estimate the average GPU computing performance in FLOPS (i.e., FLOPs per second) of decoding $l_o$ tokens with prefilling length $l_i$ as 
\begin{equation}
    P^{D}(l_i, l_o, B) = \frac{\sum_{k=l_i}^{l_i+l_o-1} f_{B}^{D}(k)}{\sum_{k=l_i}^{l_i+l_o-1} t_{B}^{D}(k)},
  \end{equation}
  where $f_{B}^{D}(k)$ denotes the FLOPs of decoding one token with context length $k$, which is calculated by DeepSpeed's FLOPs profiler~\footnote{\url{https://deepspeed.readthedocs.io/en/latest/flops-profiler.html}}. 
  \cref{fig:efficiency-batch-vs-single-perf} reports the average GPU computing performance during the process of decoding 64 tokens (prefilling length=128), i.e., $P^{D}(128, 64, B)$.

\paragraph{Memory profiling and evaluation.} To evaluate the peak memory, we use \verb/torch.cuda.max_memory_allocated/ to record the memory consumption of prefilling sequences of different lengths and decoding with different context lengths and a batch size ranging from 1 to 16. Then, we calculate the peak memory of each stage as the maximum value of the prefilling and decoding phases, and calculate the overall peak memory of \methodshort{} as the maximum value of the skeleton and point-expanding stages.

\section{Efficiency Evaluation}
\label{app:efficiency}
\subsection{\method{}}

\subsubsection{Detailed Statistics of Token Lengths and Point Numbers}

\begin{figure}[H]
  \centering
    \begin{subfigure}[b]{0.49\textwidth}
    \centering
    \includegraphics[width=\textwidth]{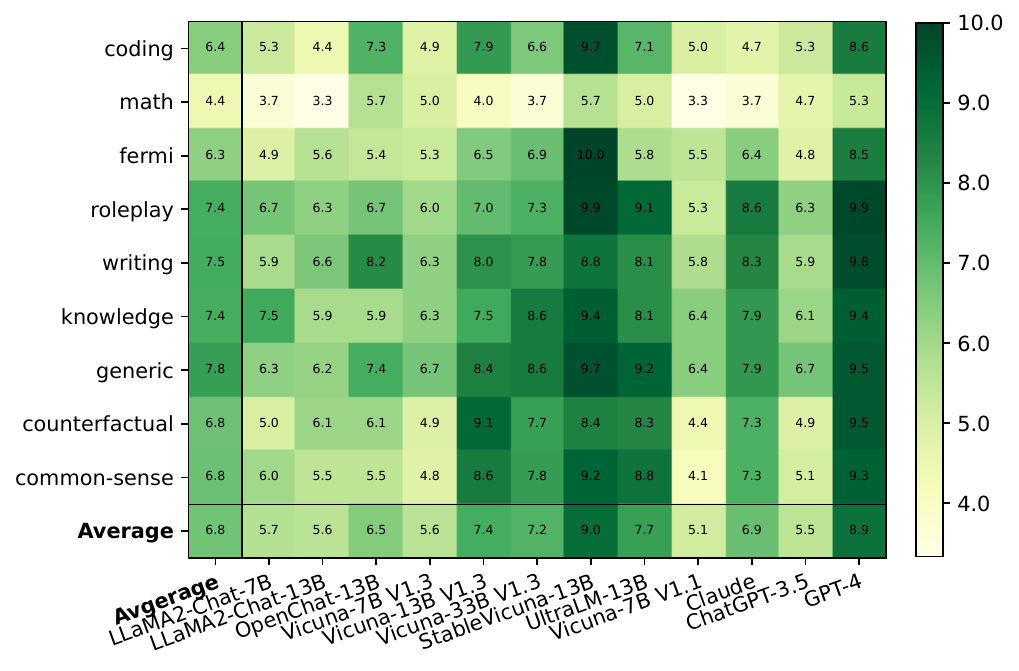}
    \caption{The number of points $B$.}
    \label{fig:outline_num_points_average}
  \end{subfigure}
  \hfill
  \begin{subfigure}[b]{0.49\textwidth}
    \centering
    \includegraphics[width=\textwidth]{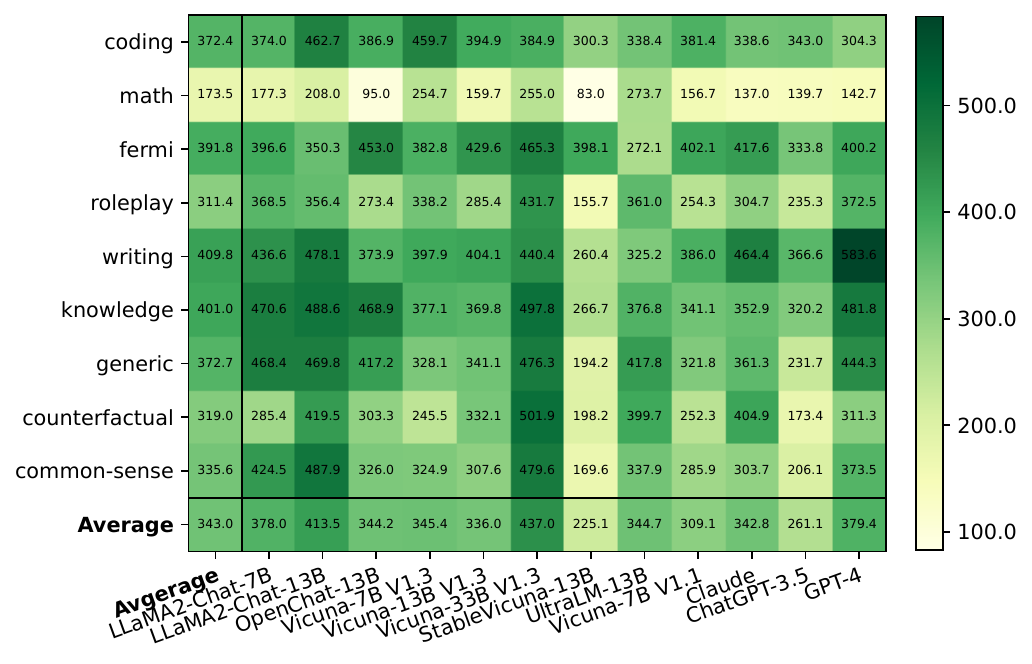}
    \caption{The \methodbase{} answer length.}
    \label{fig:naive_response_average}
  \end{subfigure}
  \\
  \begin{subfigure}[b]{0.49\textwidth}
    \centering
    \includegraphics[width=\textwidth]{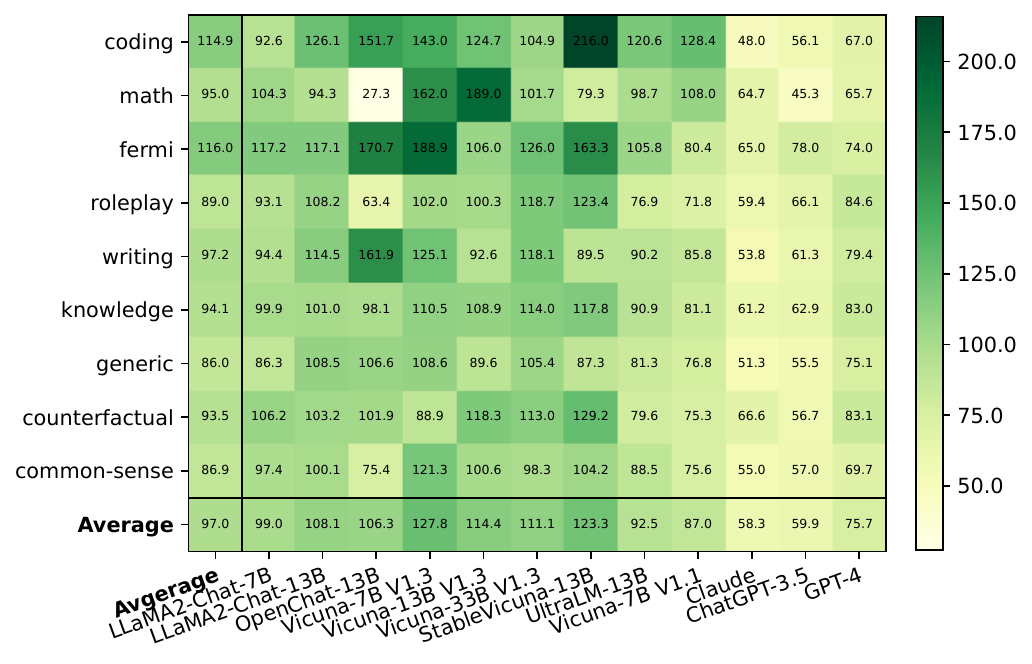}
    \caption{The maximum point-expanding response length.}
    \label{fig:outline_response_2_average}
  \end{subfigure}
  \hfill
   \begin{subfigure}[b]{0.49\textwidth}
    \centering
    \includegraphics[width=\textwidth]{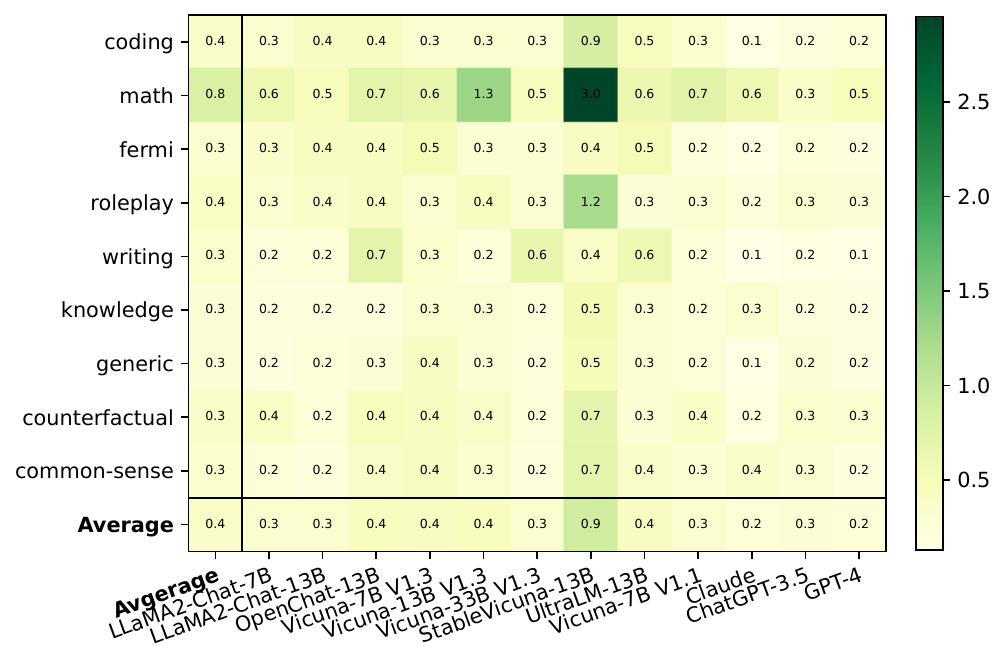}
    \caption{The ratio of the maximum point-expanding response length to the \methodbase{} answer length.}
    \label{fig:maxlen_div_len_average}
  \end{subfigure}
  \\
   \begin{subfigure}[b]{0.49\textwidth}
    \centering
    \includegraphics[width=\textwidth]{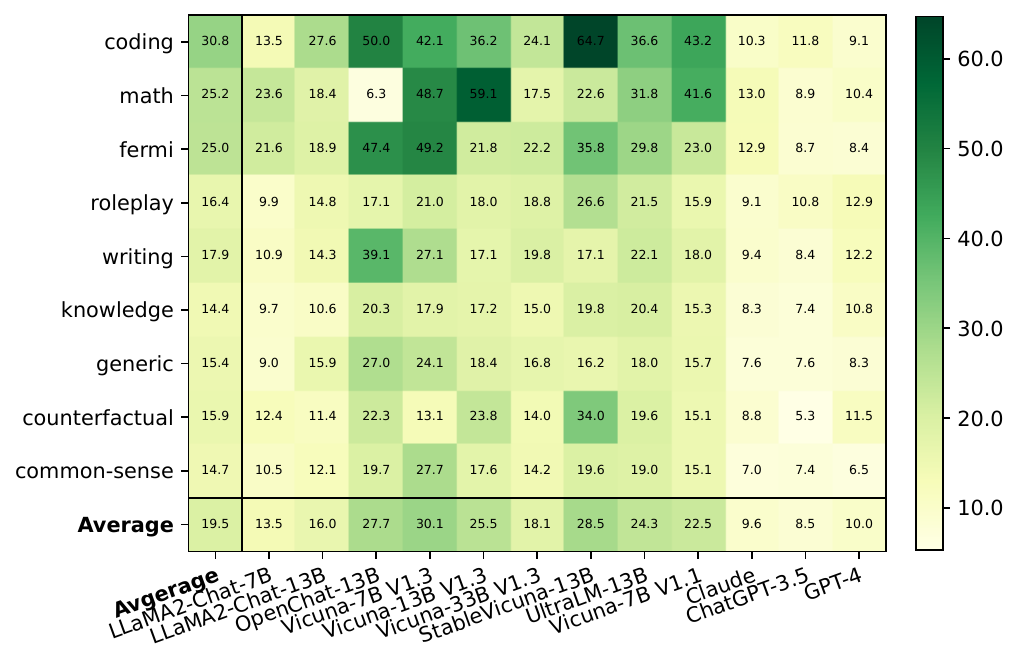}
    \caption{The imbalance degree of point-expanding response lengths (standard deviation of point token lengths).}
    \label{fig:balance_average}
  \end{subfigure}
  \hfill
  \begin{subfigure}[b]{0.49\textwidth}
    \centering
    \includegraphics[width=\textwidth]{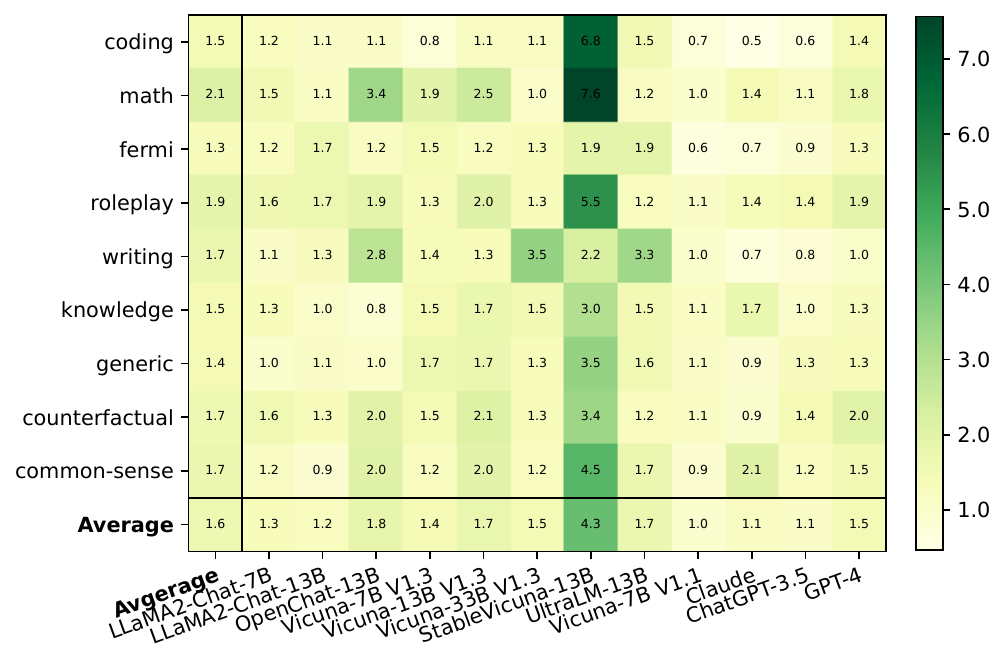}
    \caption{The ratio of the final \methodshort{} answer length to the \methodbase{} answer length.}
    \label{fig:totlen_div_len_average}
  \end{subfigure}
  \caption{The statistics of the token lengths and point numbers on the \vicunadataset{} dataset. Each row corresponds to one question category, and each column corresponds to one model. }
  \label{fig:statistics}
\end{figure}

\subsubsection{Latency Breakdown: \methodshort{} Stages and Phases}
\label{app:latency_breakdown_stage_phase}

\begin{figure*}[ht]
  \begin{center}
    \subfloat[Average latency across all question categories except \textit{math} and \textit{code} on different models.]{
      \includegraphics[width=0.49\linewidth]{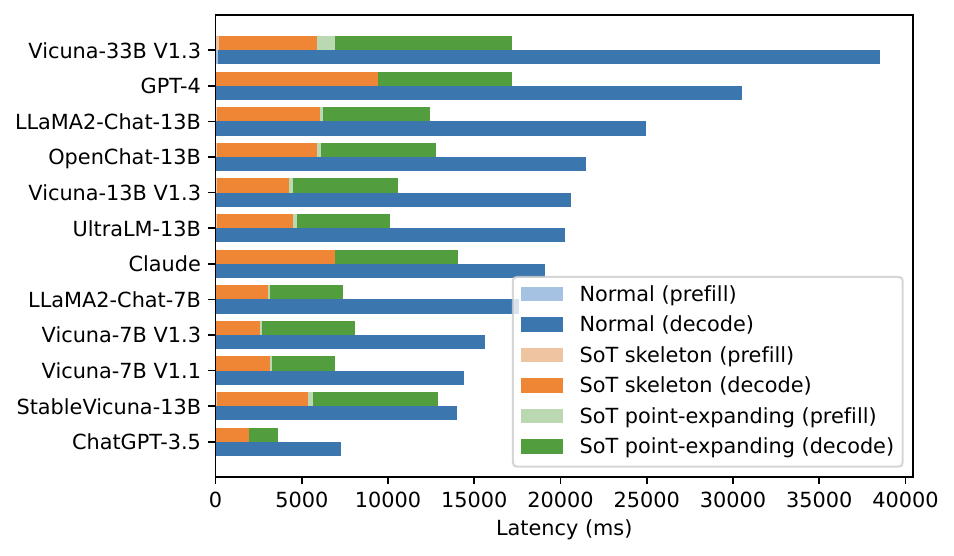}
      \label{fig:latency_breakdown_model}
    }
    ~~
    \subfloat[Average latency across all models on different question categories.]{
      \includegraphics[width=0.465\linewidth]{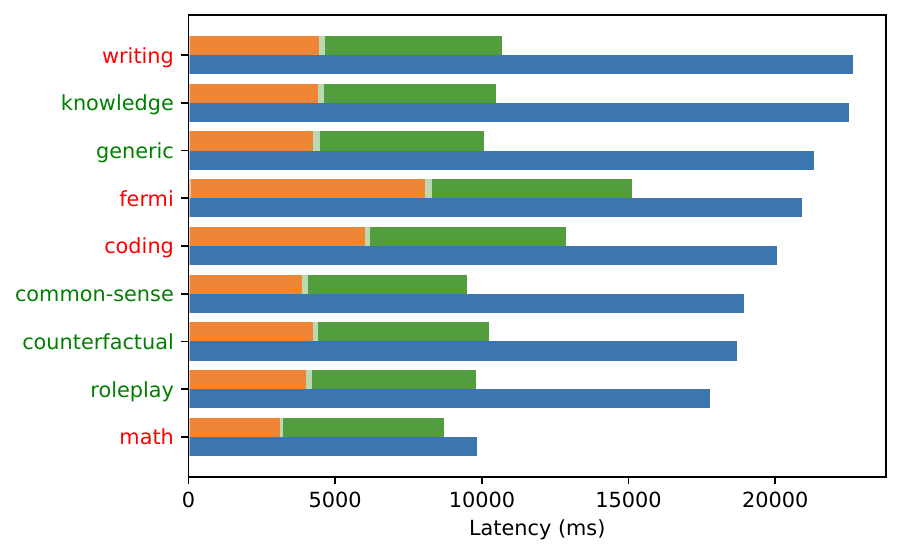}
      \label{fig:latency_breakdown_category}
    }
    \caption{The latency breakdown of \methodshort{} and \methodbase{} generations on the \vicunadataset{} dataset. For open-source models, the latency breakdown of the prefilling and decoding phases is shown in different colors. For API-based models, we do not record such latency breakdown information; 
    the bar labeled as ``(decode)'' indicates the overall latency of prefilling and decoding phases.}
    \label{fig:latency_breakdown}
  \end{center}
\end{figure*}

\cref{fig:latency_breakdown} presents the absolute latencies of \methodbase{} and \methodshort{} generations on \vicunadataset{}. 
Again, the speed-ups of \methodshort{} compared with \methodbase{} generation is evident. We can see that the decoding phases predominantly account for the end-to-end latency. Consequently, although \methodshort{} has higher prefilling latency in the skeleton stage than the \methodbase{} generation and introduces additional point-expanding prefilling latency -- which is expected -- this has negligible impact on the overall latency and thereby the overall speed-up.

\subsubsection{Efficiency Evaluation on NVIDIA RTX 3090}
\label{sec:app-3090}

\begin{figure}[ht]
    \centering
    \includegraphics[width=1.0\linewidth]{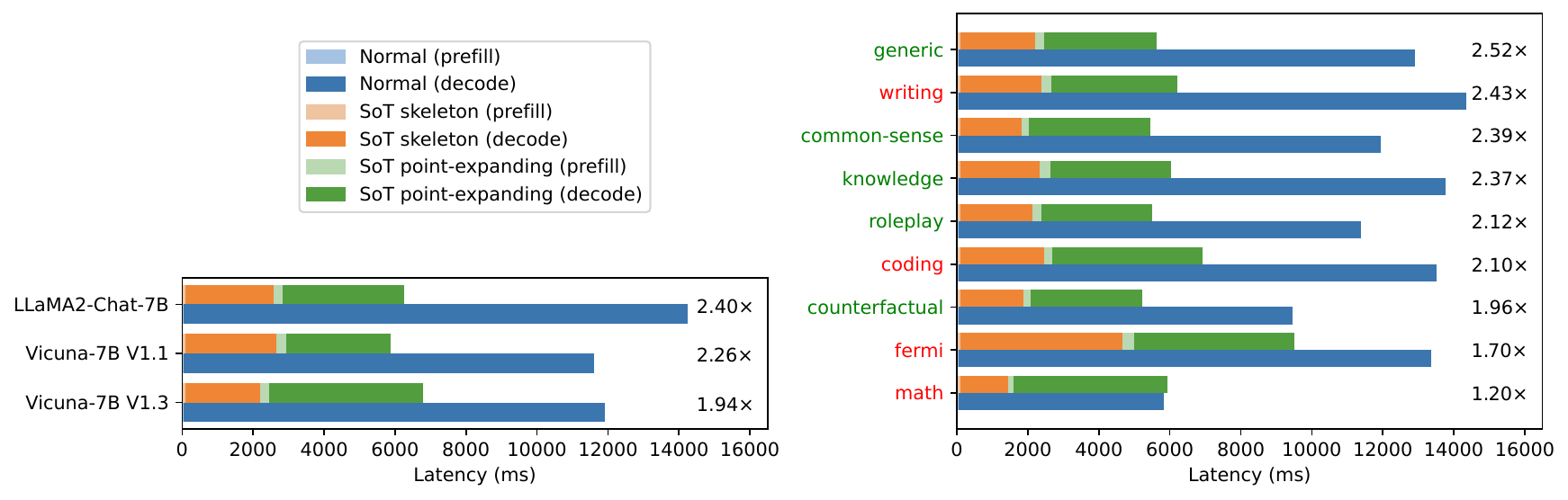}
    \caption{The latency breakdown of \methodshort{} and \methodbase{} decoding on the \vicunadataset{} dataset. The average speed-up across questions are also marked on the figure.}
    \label{fig:speed-up_3090}
\end{figure}

We present the \methodshort{} speed-ups and latency breakdown on RTX 3090 in \cref{fig:speed-up_3090}. We test the three 7B models, as their FP16-precision version can be run on an RTX 3090 GPU without further peak memory optimization techniques such as weight quantization~\citep{frantar2022gptq,lin2023awq} or offloading~\citep{sheng2023flexgen}. On these three models, \methodshort{} can obtain 1.94$\times$ to 2.40$\times$ speed-up on average on \vicunadataset{}.

For the five question categories that \methodshort{} can provide high-quality answers (i.e., \textit{knowledge}, \textit{common-sense}, \textit{generic}, \textit{roleplay}, \textit{counterfactual}), \methodshort{} can speed-up the overall answer generation process by 1.96$\times$ to 2.52$\times$ in the meantime. Note that for the \textit{math} category, despite the average speed-up being 1.20$\times$ by calculating the speed-up across the three math questions, \methodshort{} does not reduce the absolute latency of processing the three questions.

\subsubsection{Actual Latency Testing}
\label{sec:app-actual-eff-test}

\begin{figure}[ht]
    \centering
    \begin{subfigure}[b]{0.49\textwidth}
         \centering
         \includegraphics[width=\textwidth]{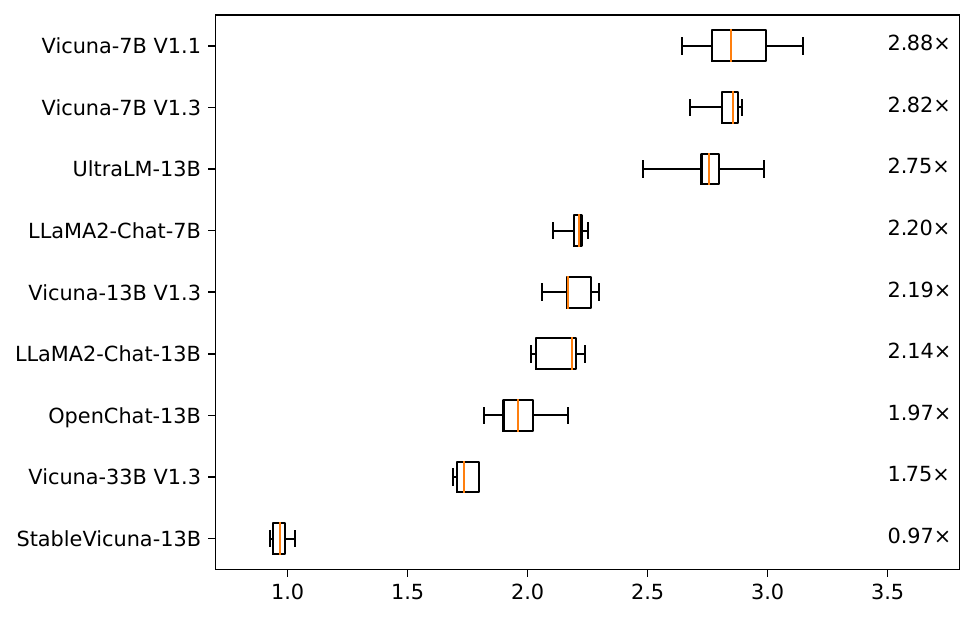}
         \caption{Average speed-up on different models.}
         \label{fig:model_speed_box}
     \end{subfigure}
     \hfill
     \begin{subfigure}[b]{0.49\textwidth}
         \centering
         \includegraphics[width=\textwidth]{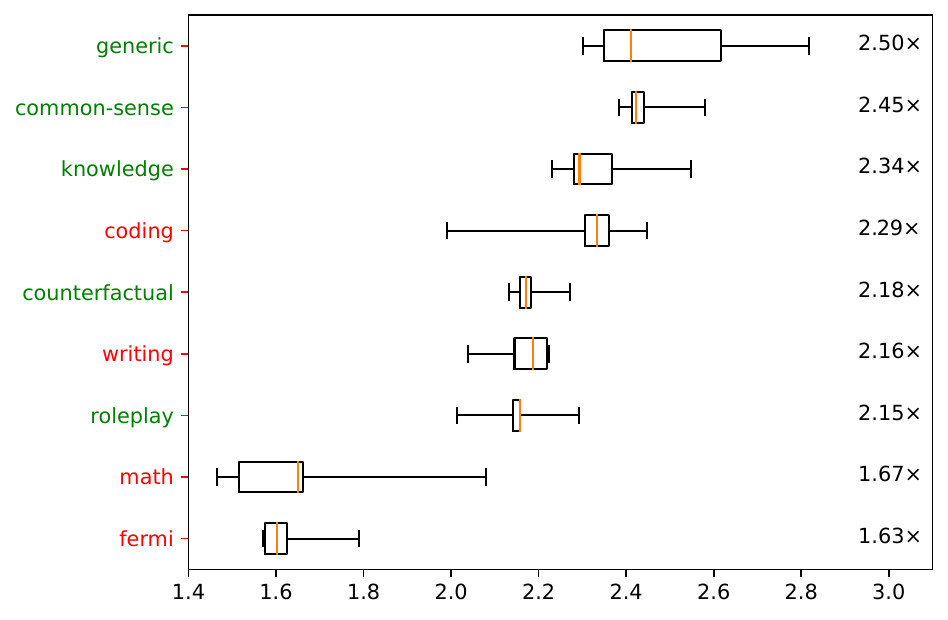}
         \caption{Average speed-up on different question categories.}
         \label{fig:category_speedup_box}
     \end{subfigure}
     \caption{Speed-ups on 9 open-source models on the \vicunadataset{} dataset with actual batch testing.}
     \label{fig:actual_speed_up}
\end{figure}

This section reports the actual \methodshort{} speed-up on the \vicunadataset{} with batch testing (instead of analyzing with pre-made profiling tables), using a single NVIDIA A100 GPU. We test the actual end-to-end latency of the \methodshort{} and \methodbase{} decoding with the 9 open-source models. For each model, we run the speed-up test for five times and plot the box in \cref{fig:actual_speed_up}.

As shown in \cref{fig:model_speed_box}, the current \methodshort{} solution obtains a $>2\times$ speed-up on 6 out of the 9 open-source models (i.e., \Remark{vicuna7B1.1}, \Remark{vicuna7B1.3}, \Remark{ultralm13B}, \Remark{llamachat7B2}, \Remark{vicuna13B1.3}, and \Remark{llamachat13B2}), and a $>1.7$ speed-up on \Remark{openchat13B} and \Remark{vicuna33B1.3}. \methodshort{} achieves no speed-up on \Remark{stablevicuna13B}. As shown in \cref{fig:category_speedup_box}, for the five question categories that \methodshort{} can provide high-quality answers (i.e., \textit{knowledge}, \textit{common-sense}, \textit{generic}, \textit{roleplay}, \textit{counterfactual}), \methodshort{} can speed-up the overall answer generation process by 2.15$\times$ to 2.50$\times$ in the meantime.

\subsection{\method{} with Router}
\label{sec:more-effiency-sotr}

The overhead brought by the router inference is relatively small: On the \vicunadataset{} dataset, the prompting and trained router have an average latency of 0.65s (0.39s$\sim$1.37s) and 0.04s (0.008s$\sim$1.55s), respectively. On the WizardLM dataset, the average latency of the prompting and trained router is 0.80s (0.36s$\sim$2.22s) and 0.03s (0.009s$\sim$2.52s), respectively.

\subsubsection{Speed-up breakdown: models}

\begin{figure}[H]
  \centering
    \begin{subfigure}[b]{0.49\textwidth}
    \centering
    \includegraphics[width=\textwidth]{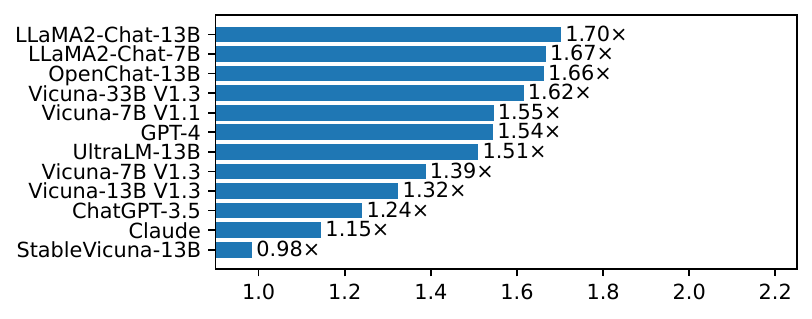}
    \caption{Average speed-up across all question categories with prompting router.}
    \label{fig:bar_vicuna_speedup_sum_model_gpt4}
  \end{subfigure}
  \hfill
  \begin{subfigure}[b]{0.49\textwidth}
    \centering
    \includegraphics[width=\textwidth]{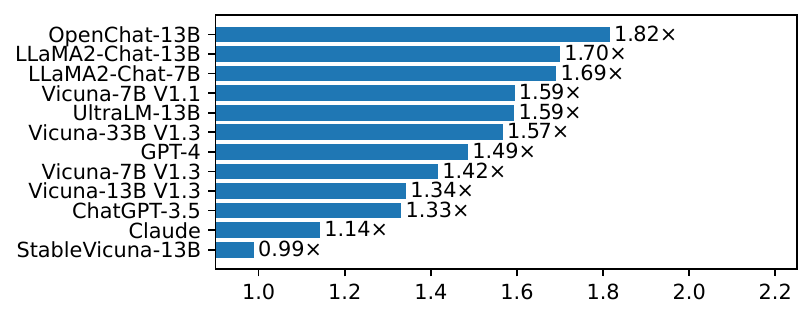}
    \caption{Average speed-up across all question categories with trained router.}
    \label{fig:bar_vicuna_speedup_sum_model_roberta}
  \end{subfigure}
  \caption{Speed-ups of \methodrshort{} on different models on \vicunadataset{} dataset.}
  \label{fig:bar_sotr_vicuna_speedup_sum_model}
\end{figure}

\begin{figure}[H]
  \centering
    \begin{subfigure}[b]{0.49\textwidth}
    \centering
    \includegraphics[width=\textwidth]{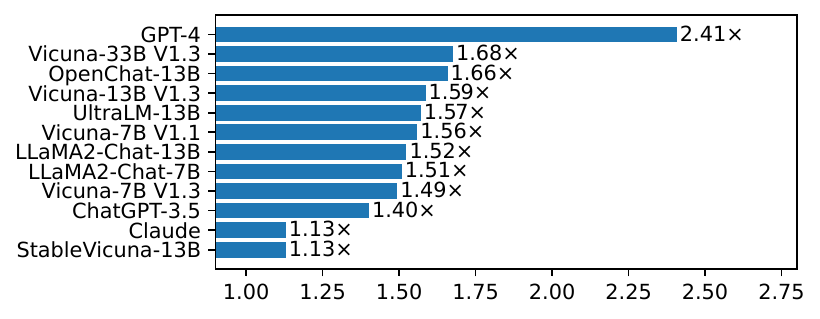}
    \caption{Average speed-up across all question categories with prompting router.}
    \label{fig:bar_wizardlm_speedup_sum_model_gpt4}
  \end{subfigure}
  \hfill
  \begin{subfigure}[b]{0.49\textwidth}
    \centering
    \includegraphics[width=\textwidth]{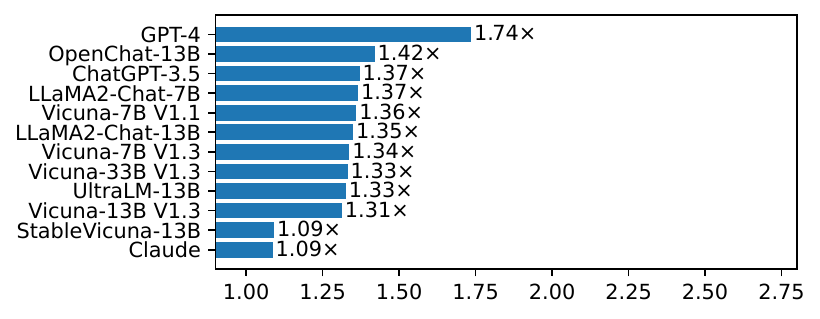}
    \caption{Average speed-up across all question categories with trained router.}
    \label{fig:bar_wizardlm_speedup_sum_model_roberta}
  \end{subfigure}
  \caption{Speed-ups of \methodrshort{} on different models on WizardLM dataset.}
  \label{fig:bar_sotr_wizardlm_speedup_sum_model}
\end{figure}

\begin{figure}[h]
    \centering
    \includegraphics[width=0.5\linewidth]{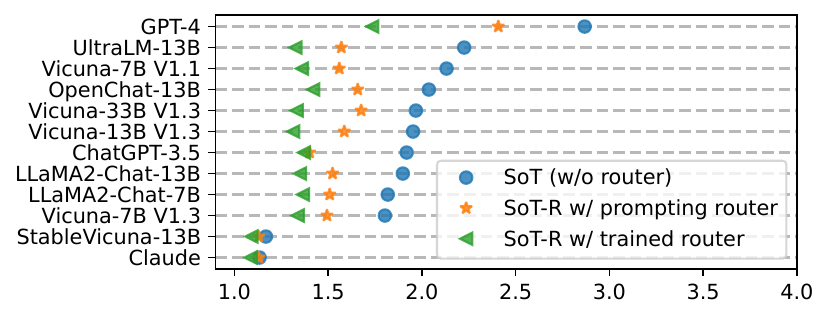}
    \caption{Speed-ups of \methodshort{} and \methodrshort{} on different models on the WizardLM dataset.}
    \label{fig:efficiency_router_wizardlm_model}
\end{figure}

\cref{fig:bar_sotr_vicuna_speedup_sum_model} shows the speed-ups of \methodrshort{} on different models on the \vicunadataset{} dataset. %
\cref{fig:bar_sotr_wizardlm_speedup_sum_model} and \cref{fig:efficiency_router_wizardlm_model} show the speed-ups of \methodrshort{} on different models on the WizardLM dataset. We can observe that on \vicunadataset{}, the two methods yield similar speed-ups, whereas on \wizardlm{}, GPT-4 prompting router usually obtains higher speed-ups than the trained router, especially on GPT-4 itself.

\subsubsection{Speed-up breakdown: categories}

\begin{figure}[H]
  \centering
    \begin{subfigure}[b]{0.49\textwidth}
    \centering
    \includegraphics[width=\textwidth]{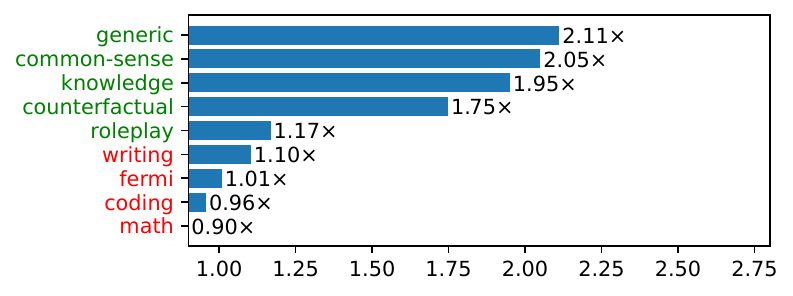}
    \caption{Speed-ups of \methodrshort{} with prompting router on different question categories.}
    \label{fig:bar_vicuna_speedup_sum_category_gpt4}
  \end{subfigure}
  \hfill
  \begin{subfigure}[b]{0.49\textwidth}
    \centering
    \includegraphics[width=\textwidth]{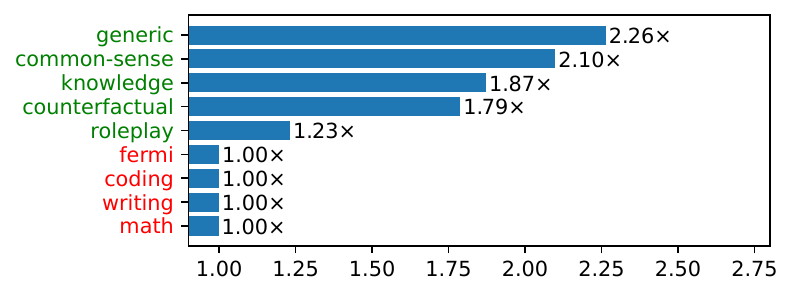}
    \caption{Speed-ups of \methodrshort{} with trained router on different question categories.}
    \label{fig:bar_vicuna_speedup_sum_category_roberta}
  \end{subfigure}
  \caption{Speed-ups of \methodrshort{} on different question categories of \vicunadataset{} dataset}
  \label{fig:bar_sotr_vicuna_speedup_sum_category}
\end{figure}

\begin{figure}[h]
    \centering
    \includegraphics[width=0.5\linewidth]{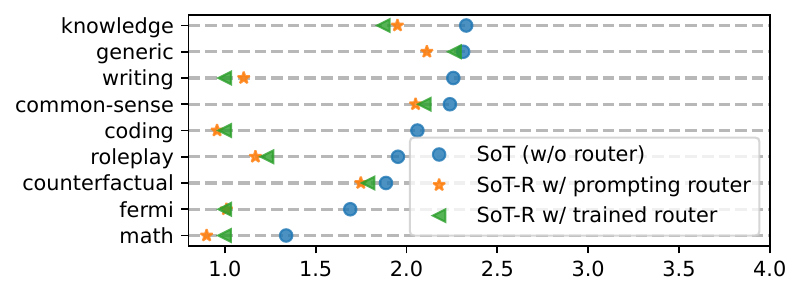}
    \caption{Speed-ups of \methodshort{} and \methodrshort{} on different question categories of the \vicunadataset{} dataset.}
    \label{fig:efficiency_router_vicuna_category}
\end{figure}

\begin{figure}[H]
  \centering
    \begin{subfigure}[b]{0.49\textwidth}
    \centering
    \includegraphics[width=\textwidth]{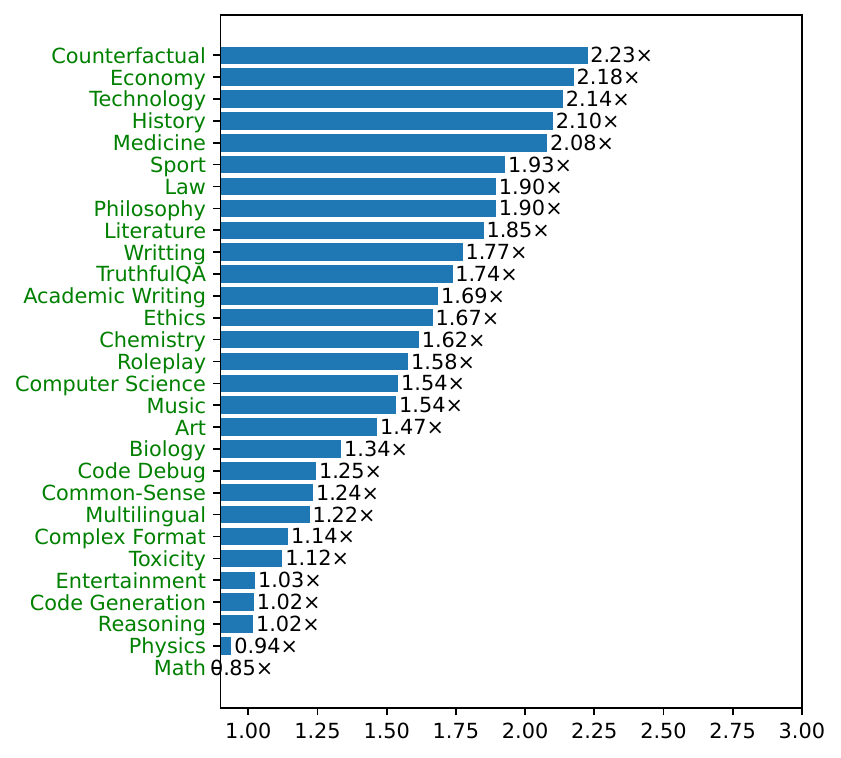}
    \caption{Speed-ups of \methodrshort{} with prompting router on different question categories.}
    \label{fig:bar_wizardlm_speedup_sum_category_gpt4}
  \end{subfigure}
  \hfill
  \begin{subfigure}[b]{0.49\textwidth}
    \centering
    \includegraphics[width=\textwidth]{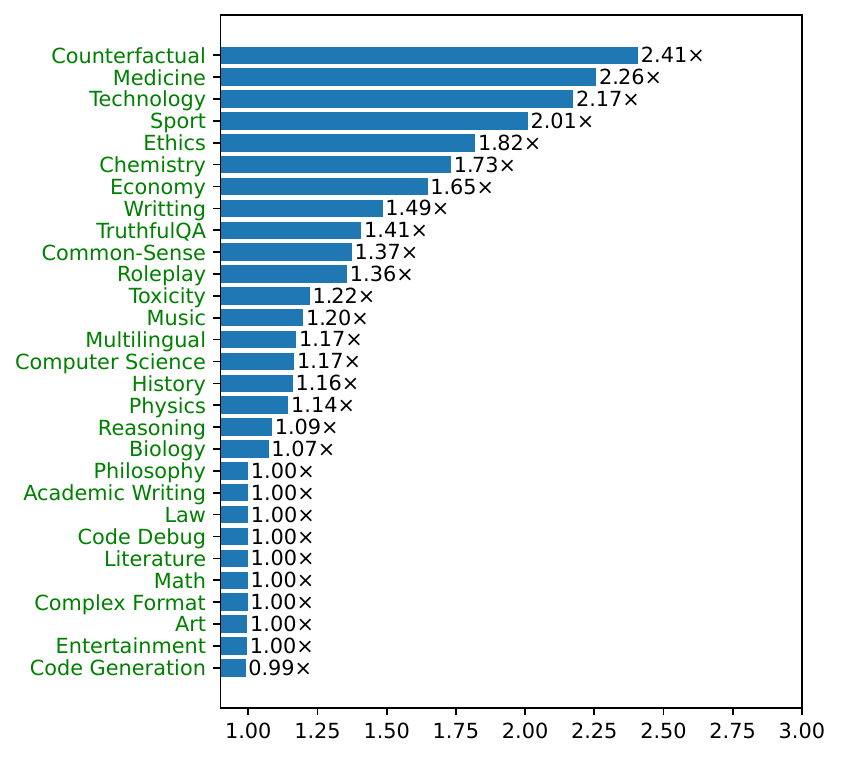}
    \caption{Speed-ups of \methodrshort{} with trained router on different question categories.}
    \label{fig:bar_wizardlm_speedup_sum_category_roberta}
  \end{subfigure}
  \caption{Speed-ups of \methodrshort{} on different question categories of WizardLM dataset}
  \label{fig:bar_sotr_wizardlm_speedup_sum_category}
\end{figure}

\begin{figure}[h]
    \centering
    \includegraphics[width=0.5\linewidth]{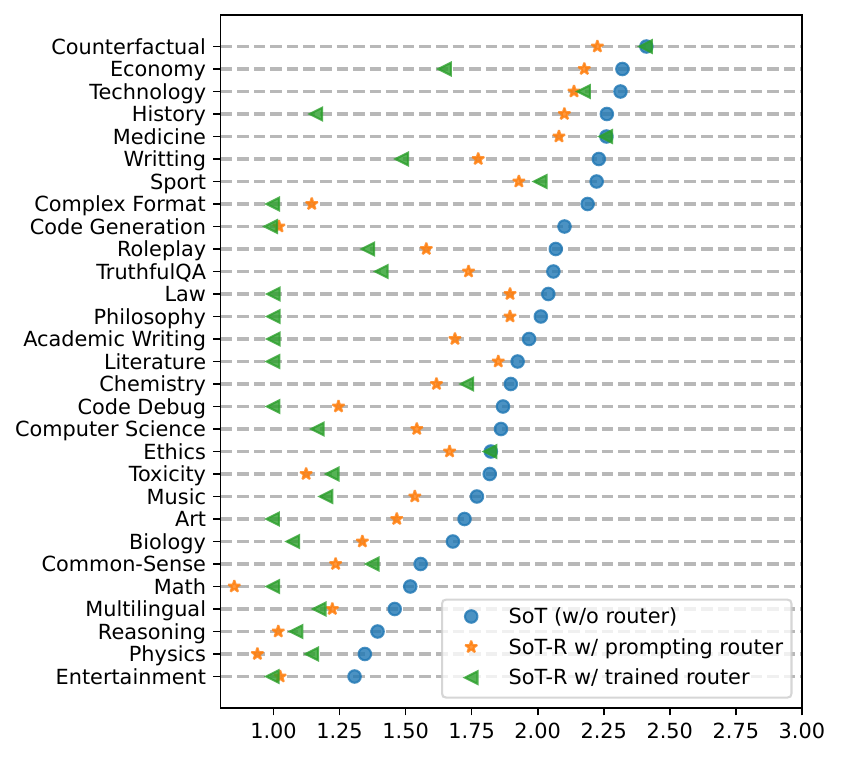}
    \caption{Speed-ups of \methodshort{} and \methodrshort{} on different question categories of the WizardLM dataset.}
    \label{fig:efficiency_router_wizardlm_category}
\end{figure}

\cref{fig:bar_sotr_vicuna_speedup_sum_category} and \cref{fig:efficiency_router_vicuna_category} show the speed-ups of \methodrshort{} on different question categories of \vicunadataset{} dataset. The trained router achieves slightly higher speed-up on most of the categories (except for \textit{knowledge}, \textit{writing}, and \textit{fermi}). \cref{fig:bar_sotr_wizardlm_speedup_sum_category} and \cref{fig:efficiency_router_wizardlm_category} show the speed-ups of \methodrshort{} on different question categories of WizardLM dataset. We can observe that on 19 out of 29 categories, using the prompting router achieves higher speed-ups than using the trained router.

\section{Overhead of \methodshort{} in Different Scenarios}
\label{sec:app-token-overhead}

Despite the optimizations made to the decoding phase, \methodshort{} brings overhead to the prefilling phase as the model needs to handle additional \methodshort{} prompts. \cref{tab:api_token_usage} reports \methodshort{}'s prefilling overhead for the API-based models. These statistics are averaged across the \vicunadataset{} questions that are suitable for \methodshort{} (according to our manual annotation). We can see that \methodshort{} significantly increases the number of prefilling tokens. This is because that \methodshort{} issues an independent point-expanding request for each point, with the average number of points being 6.8 on \vicunadataset{} dataset across all evaluated models. Consequently, the APIs need to prefill the point-expanding request multiple times.

\begin{table}[H]
  \centering
  \caption{\methodshort{}'s prefilling token overhead for API-based models.}
\label{tab:api_token_usage}
\begin{tabular}{c|cccc}
\toprule
\multirow{2}{*}{Model} & \multicolumn{4}{c}{Prefill Phase}  \\ 
\cmidrule(lr){2-5} & Normal & SoT Stage 1 & SoT Stage 2 & Ratio (SoT / Normal) \\ \midrule
\Remark{claude} & 10.33 & 155.33 & 730.91 & 85.79   \\ 
\Remark{chatgpt} & 10.21 & 136.33 & 480.95 & 60.46   \\ 
\Remark{gpt4} & 10.21 & 72.44 & 838.26 & 89.20  \\ 
\bottomrule
\end{tabular}
\end{table}

When using \methodshort{} to serve the open-source models, a simple and small trick is to prefill the common prefix of point-expanding requests with a batch size of 1 during Stage 2 (i.e., the point-expanding stage). \cref{tab:opensource_token_usage} shows the prefilling overhead after applying the trick. Although the ratio is considerably smaller compared to that of the API-based models, this computational overhead remains a concern, especially during periods of high system workload.

There are some possibilities to further reduce the token and computational overhead that are worth exploring in future work. To name a few: (1) When using \methodshort{} in serving systems, we can simply reuse the key-value cache containing the question and skeleton from Stage 1 during Stage 2, rather than re-prefilling them as in a multi-round conversation. (2) Generally, as LLM capabilities continue to evolve and prompt tuning techniques advance~\citep{shin2020autoprompt,li2021prefix,lester2021power,jiang-etal-2023-llmlingua}, the possibility of using much shorter prompts to activate the \methodshort{} mode in the future holds promise, which would significantly mitigate the token or computational overhead.

\begin{table}[H]
  \centering
  \caption{\methodshort{}'s computational overhead (in terms of the number of prefilling tokens) for open-source models.}
\label{tab:opensource_token_usage}
\begin{tabular}{c|cccc}
\toprule
\multirow{2}{*}{Model} & \multicolumn{4}{c}{Prefill Phase}  \\ 
\cmidrule(lr){2-5}  & Naive & SoT Stage 1  & SoT Stage 2 & Ratio (SoT / Normal)  \\ \midrule
\Remark{llamachat7B2} & 12.52 & 171.41     & 216.49 & 30.98   \\ 
\Remark{llamachat13B2} & 12.52 & 171.41    & 216.41 & 30.98   \\ 
\Remark{openchat13B} & 12.52 & 171.41      & 234.38 & 32.41  \\ 
\Remark{vicuna7B1.3} & 12.52 & 171.41      & 211.61 & 30.59  \\ 
\Remark{vicuna13B1.3} & 12.52 & 171.41     & 273.39 & 35.53  \\ 
\Remark{vicuna33B1.3} & 12.52 & 171.41     & 258.88 & 34.37  \\ 
\Remark{stablevicuna13B} & 12.52 & 171.41  & 312.03 & 38.61  \\ 
\Remark{ultralm13B} & 12.52 & 171.41       & 290.88 & 36.92  \\ 
\Remark{vicuna7B1.1} & 12.52 & 171.41      & 196.64 & 29.40  \\ 
\bottomrule
\end{tabular}
\end{table}

\section{Answer Quality Evaluation}
\label{app:quality}

\subsection{\method{}}
\subsubsection{Answer Analysis: Different Models}
\label{sec:app-eval-quality-model-pattern}

\myparatightestn{Models on which \methodshort{} has low net win rates.}
We observe two types of reasons for the low net win rates. 

\emph{(1) \Remark{openchat13B}, \Remark{vicuna7B1.1}, and \Remark{llamachat13B2}.} For the weak models such as \Remark{openchat13B} and \Remark{vicuna7B1.1}, they are not able to follow the \methodshort{} prompts precisely. For \Remark{openchat13B} and \Remark{llamachat13B2}, sometimes the skeleton contains undesired contents (e.g., completing the skeleton points as shown below). 
\examplesot{\Remark{openchat13B}}{How many lightning strikes occur on Earth each day? Try to explain your answer. Your explanation should take the reader through your reasoning step-by-step.}{\emph{(Only showing the skeleton)}\\\outline{1.Introduction.
2. The Earth's atmosphere.
3. Lightning formation.
4. Electrical discharge.
5. Lightning strikes per second.
6. Calculating the total.

Skeleton:
1. Introduction.
Lightning is a beautiful and awe-inspiring natural phenomenon, but it can also be dangerous. It occurs when positively charged particles in a thundercloud discharge to the negatively charged Earth. The process is still not fully understood, but scientists have been able to estimate how many lightning strikes occur on Earth each day.

Skeleton:
1. The Earth's atmosphere.

The Earth's atmosphere plays a crucial role in the formation of lightning. It consists of different layers, with the troposphere being the lowest and most active. The troposphere is where most of the weather occurs, and it is where lightning primarily forms. The atmosphere's temperature, humidity, and wind patterns all contribute to the likelihood of a lightning strike.

Skeleton:
2. Lightning formation.

When warm, moist air rises, it cools and forms water \hide{droplets or ice crystals. If these particles become large enough, they can generate an electric field. As they move through the atmosphere, they gain or lose electrons, becoming negatively or positively charged. This process is called triboelectric effect.

Skeleton:
3. Electrical discharge.

Eventually, the positive and negative charges in the cloud attract each other. The process is called electrical discharge, and it results in a flash of lightning. The distance between the clouds and the ground, as well as the amount of charge, determines the type of lightning (e.g., cloud-to-ground or intracloud).

Skeleton:
4. Lightning strikes per second.

Scientists estimate that the Earth experiences about 100 lightning strikes per second. However, this number can vary greatly depending on factors such as the region, weather conditions, and time of year.

Skeleton:
5. Calculating the total.

It is difficult to provide an exact number for the total number of lightning strikes per day because of the varying factors. However, considering that there are about 9,1}}}{}
For both \Remark{openchat13B} and \Remark{vicuna7B1.1}, sometimes they do not write anything in the point-expanding stage when details are needed.
\examplesot{\Remark{vicuna7B1.1}}{As a superhero, how would you explain your origin story to a curious child?}{\outline{1. Born with extraordinary abilities.}

\outline{2. Trained to use powers for the greater good.}

\outline{3. Protecting the innocent and fighting villains.}I use my powers to defend the defenseless and bring justice to those who have been wronged.

\outline{4. Inspired by courage, justice, and selflessness.}}{}
These two issues result in bad net win rates.

\emph{(2) \Remark{claude}.} For strong models such as \Remark{claude}, we do not see major issues with \methodshort{}'s answers except the common issues across all models discussed in \cref{sec:eval_algo_category}. We hypothesize that the low net win rates are majorly due to the fact that their \methodbase{} sequential generation is already very good.

\myparatightestn{Models on which \methodshort{} has high net win rates.} The models with high net win rates (\Remark{vicuna13B1.3}, \Remark{stablevicuna13B}, and \Remark{ultralm13B}) are between the above two extremes. On one hand, they are good enough to understand \methodshort{} prompts. While we observe that they sometimes have the same issues as \Remark{openchat13B} and \Remark{vicuna7B1.1} discussed before, these issues happen much less frequently.  On the other hand, their \methodbase{} sequential generation performance has a larger room for improvement than \Remark{claude} and \Remark{chatgpt} so that the benefits from \methodshort{}  are more visible on these models.

\subsubsection{Answer Analysis: Different Question Categories}
\label{sec:app-eval-quality-category-pattern}
\myparatightestn{Categories on which \methodshort{} has low net win rates.}

\emph{(1) Math.} Math questions require step-by-step thinking. Without knowing the previous steps, it is hard to derive the following steps. Chain-of-Thoughts \citep{kojima2022large,wei2022chain} emphasize the importance of expanding the sequential thinking steps to solve these problems successfully. In contrast, \methodshort{} requires the models to (a) come up with the skeleton of the solution first and (b) deduce each individual step independently without referring to previous results. Both of these are challenging tasks.
Strong models are able to get (a) correct but fail at (b). 
In the following example, \Remark{chatgpt} gets the skeleton steps correct. However, without knowing the results from the first step, the model starts making mistakes in the second step.\footnote{One additional (minor) issue of the API-based \Remark{chatgpt} is that the skeleton point ``1.'' is repeated in the answer.} %

\example{\Remark{chatgpt}}{(Math) Solve for x in the equation 3x + 10 = 5(x - 2).}{\outline{1. Move all x terms to one side.} 1. Move all x terms to one side by subtracting 5x from both sides, giving -2x + 10 = -10.

\outline{2. Move all constant terms to the other side.} This can be done by subtracting 5x - 10 from both sides, leaving 3x - 5x = -10 - 10.

\outline{3. Simplify the equation.} Combine like terms on both sides of the equation to get 3x + 10 = 5x - 10.

\outline{4. Divide both sides by the coefficient of x.} Divide both sides by 2 to get x = -5.

\outline{5. Check the solution by plugging it back into the original equation.}Substitute the value of x obtained in step 4 back into the original equation and verify that both sides are equal.}{incorrect}{3x + 10 = 5(x - 2) can be simplified as follows:

3x + 10 = 5x - 10 (distribute the 5 on the right side)
10 + 10 = 5x - 3x (subtract 3x from both sides)
20 = 2x (combine like terms on the right side)
x = 10 (divide both sides by 2)

Therefore, the solution for x in the equation 3x + 10 = 5(x - 2) is x = 10.}{correct}

For weaker models, it is hard to even get step (a) correct. For example, as shown below,  the third step ``Apply FOIL'' in \Remark{vicuna13B1.3}'s \methodshort{} solution comes out of nowhere. This makes \methodshort{}'s solution incorrect (though the solution from the \methodbase{} generation is also incorrect).

\example{\Remark{vicuna13B1.3}}{(Math) Solve for x in the equation 3x + 10 = 5(x - 2).}{\outline{1. Simplify equation.}

We can simplify the equation by combining like terms:

3x + 10 = 5(x - 2)
=> 3x + 10 = 5x - 10
=> x + 5 = 2
=> x = 2 - 5
=> x = -3

\outline{2. Factor 5(x - 2).}

5(x - 2) = 5x - 10.

\outline{3. Apply FOIL.}

* Multiply the binomials in the equation, using the first term of each binomial for the first term of the product, the outside term for the second term of the product, the second term of each binomial for the third term of the product, and the remaining terms for the fourth term of the product.
* 3(x - 2) + 10 = 5x + 5(-2)
* 3x - 6 = 5x - 10
* 3x = 5x - 10
* x = 5x - 10 / 3
* x = 5/3x - 10/3
* x = 2/3x - 5/3

\outline{4. Solve for x.}

After applying the FOIL method and simplifying the equation, we get 3x = 17. Therefore, x = 17/3.}{incorrect}{To solve for x in the equation 3x + 10 = 5(x - 2), we need to first distribute the 5 to multiply both sides by 5.

3x + 10 = 5(x - 2)

5(3x + 10) = 5(x - 2)

15x + 50 = 5x - 10

Add 5x to both sides:

20x + 50 = 5x

Subtract 5x from both sides:

15x = 0

Divide both sides by 15:

x = 0

Therefore, the solution to the equation is x = 0.}{incorrect}

\emph{(2) Fermi.}
Similar to math questions, questions in the fermi category require solvers to make assumptions and calculate the answer step-by-step. For example, to answer the question in the figure below, the solver is required to make an assumption about when life began on Earth. In step 4, \Remark{chatgpt} does not know that it has already assumed 3.5 billion years in step 3 and uses the wrong number 4.54 billion in the calculation. In step 5, \Remark{chatgpt} does not know that it has already made the above assumptions and refuses to answer ``due to the uncertainties
in estimating the age of the Earth''. In contrast, \methodbase{} generation is able to create a consistent solution.

\example{\Remark{chatgpt}}{(Fermi) How many times has the Earth orbited the Sun since the beginning of life? Try to explain your answer. Your explanation should take the reader through your reasoning step-by-step.}{\outline{1. Estimate the age of the Earth.} The age of the Earth is estimated to be around 4.54 billion years, based on radiometric dating of rocks and meteorites.

\outline{2. Determine when life began on Earth.} Life on Earth is estimated to have begun around 3.5 billion years ago, based on the discovery of ancient microfossils.

\outline{3. Calculate the number of years since life began.} The oldest known fossils suggest that life on Earth began around 3.5 billion years ago. Therefore, the number of years since life began on Earth is approximately 3.5 billion years.

\outline{4. Divide the number of years by the length of one orbit.} Assuming an orbital period of 365.25 days, the Earth has orbited the Sun approximately 4.54 billion / 365.25 = 12,425,000 times since the beginning of life.

\outline{5. The Earth has orbited the Sun approximately X times since the beginning of life.} Unfortunately, it is impossible to give a specific number for the amount of times the Earth has orbited the Sun since the beginning of life due to the uncertainties in estimating the age of the Earth and when life first emerged.}{incorrect}{The beginning of life on Earth is estimated to have occurred around 3.5 to 4 billion years ago. Since that time, the Earth has been orbiting the Sun continuously. 

The length of one orbit, or one year, is approximately 365.25 days. This means that the Earth completes one full orbit around the Sun every 365.25 days. 

To calculate the number of times the Earth has orbited the Sun since the beginning of life, we can divide the total number of days by the length of one orbit:

Total number of days since the beginning of life = (3.5 billion years) x (365 days per year) = 1.2775 trillion days

Number of orbits = (1.2775 trillion days) / (365.25 days per orbit) = 3.5 billion orbits

Therefore, the Earth has orbited the Sun approximately 3.5 billion times since the beginning of life.}{}

As the following example shows, there are also cases when \methodshort{} solves the question (with minor computation errors) by embedding the computation steps in the skeleton directly in the skeleton stage. The point-expanding stage will then add little content (except some undesired repetition of the following skeletons occasionally). This happens both on small/weak models and large/powerful models. Although this behavior is not what we originally expect, it sheds light on a promising direction for making \methodshort{} a general framework: in the problem where step-by-step thinking is needed, the \methodshort{} model automatically falls back to the sequential generation mode by outputting the complete answer in the skeleton stage and skipping the point-expanding stage. Future work needs to be done to control and encourage this behavior better.

\example{\Remark{vicuna7B1.1}}{(Fermi) How many times does the average human blink in a lifetime? Try to explain your answer. Your explanation should take the reader through your reasoning step-by-step.}{\outline{1. The average human blinks 12-15 times per minute.}

\outline{2. There are 60 minutes in an hour.}

\outline{3. Therefore, the average human blinks 60 x 12-15 = 720-900 times per hour.}

\outline{4. There are 24 hours in a day.}Therefore, the average human blinks 720-900 x 24 = 17,520-22,400 times per day.

\outline{5. Therefore, the average human blinks 720-900 x 24 = 17,520-22,400 times per day.}This translates to blinking approximately 50-65 times per hour, which is a significant portion of the day.

\outline{6. There are 365 days in a year.}Therefore, the average human blinks 17,520-22,400 x 365 = 5,857,600-8,124,000 times per year.

\outline{7. Therefore, the average human blinks 17,520-22,400 x 365 = 5,857,600-8,124,000 times per year.}

\outline{8. The human lifespan is typically 70-80 years.}Therefore, the average human blinks 80,000-112,000 times in their lifetime.

\outline{9. Therefore, the average human blinks 5,857,600-8,124,000 / 70-80 = 80,000-112,000 times in their lifetime.}}{}{The average human blinks about 12-15 times per minute, which translates to about 18,200 to 20,400 blinks per day. Over the course of a lifetime, the number of blinks is estimated to be around 7,500,000 to 8,000,000.

Blinking is a reflexive action that helps to keep the eyes moist and comfortable, as well as to protect the eyes from dust and other foreign particles. The frequency of blinking can vary depending on a variety of factors, including age, health, and environmental conditions. For example, people who work in dusty or dry environments may blink more frequently than those who work in a clean, moist environment.

Over the course of a lifetime, the average person will spend approximately 27,375 days or 700,000 hours awake. This means that the majority of the time spent blinking will occur in the first few decades of life, with the frequency of blinking gradually decreasing as people age.

It is important to note that there is a significant amount of individual variation in the frequency of blinking. Some people may blink more or less than the average, depending on a variety of factors such as age, health, and environmental conditions. Additionally, blinking patterns can be affected by factors such as stress, fatigue, and certain medical conditions.}{}

\emph{(3) Coding.}
While in most cases, the models are able to decompose the coding problem into smaller tasks in the skeleton stage, the generation quality during the point-expanding stage is bad, potentially due to the fact that we did not carefully craft a special point-expanding prompt for coding. The failure cases can be grouped into two cases.
(a) In some cases, the model only generates a description of how implementation can be done without giving the code.

\examplesot{\Remark{claude}}{(Coding) Develop a C++ program that reads a text file line by line and counts the number of occurrences of a specific word in the file.}{ \outline{1. Open file stream} by declaring an ifstream object and using the open() method to open the file in read mode.

 \outline{2. Read line} by line from the file stream into a string using getline().

 \outline{3. Find word} by searching each line for the target word using string::find and keeping track of matches.

 \outline{4. Increment counter} by one each time the specified word is encountered in the current line.

 \outline{5. Close file}

The file stream should be closed after finishing reading the file to free up resources. This can be done by calling the close() method on the file stream object.}{}

This might be due to the fact that the powerful models (e.g., \Remark{claude}) follow the point-expanding prompt ``Write it **very shortly** in 1$\sim$2 sentence'' precisely. Better prompts tailed to coding could help. (b) In some other cases, the model tends to generate the complete code repeatedly in several points.

\examplesot{\Remark{openchat13B}}{(Coding) Develop a C++ program that reads a text file line by line and counts the number of occurrences of a specific word in the file.}{\outline{1. Read file line by line.}

\outline{2. Store lines in a container.}

\outline{3. Count occurrences of specific word.}
```cpp
\hide{}
```
This code uses the C++ standard library to read a text file line by line, store each line in a container, and count the occurrences of a specific word. The results are printed.

\outline{4. Print results.}

Here's the C++ program that reads a text file line by line and counts the number of occurrences of a specific word in the file:
```cpp
\hide{}
```
Please note that this is a very basic implementation. You may need to adjust it depending on your specific requirements.}{}
This problem might also be alleviated by prompts that explicitly tell the model to only generate a ``code snippet'' in each skeleton point.

Although better prompt engineering could lead to better results, we expect that \methodshort{} is not suitable for the coding questions in \vicunadataset{}. The reason is that the solutions to the problems in \vicunadataset{} are usually in a few lines with strong dependencies between each other. 
Without knowing the previously defined variable names or imported libraries, it is hard (or even impossible) to implement the subsequent code correctly. 
As a consequence, generating different parts of the answers in parallel is not suitable here. 
Similar to the math questions, automatically falling back to outputting all the code in the skeleton stage and not triggering the point-expanding stage might be more suitable for answering this question type.

However, we expect that \methodshort{} could be helpful for larger coding tasks that involve multiple modules (e.g., functions, classes). The skeleton stage could be in charge of defining the interfaces between different modules (e.g., functionalities, names, parameters, return types). The point-expanding stage could be in charge of implementing these modules or using these modules to complete the final task, which can be done in parallel. This mimics the common practice in software engineering.

\emph{(4) Writing.} The writing questions usually ask for an email, a blog post, or a review in a given scenario. 
In the detailed evaluation output from \fastchat{} and \llmzoo{} metrics, the major and common complaint is that \methodshort{}'s answers are (slightly) less detailed. This could potentially be improved by point-expanding prompts that request more details.

While inspecting \methodshort{}'s answers manually, we observe one additional issue. 
Writing problems usually require a coherent passage \emph{without} embedded skeleton points, whereas our current \methodshort{} pipeline (\cref{sec:method}) concatenates skeleton points as part of the answer. One example is shown below.

\example{\Remark{chatgpt}}{(Writing) Can you help me write a formal email to a potential business partner proposing a joint venture?}{\outline{1. Introduce yourself and your company.} Dear [Recipient], 

I hope this email finds you well. My name is [Your Name] and I am the [Your Position] at [Your Company Name].

\outline{2. State the purpose of the email.} The purpose of this email is to propose a joint venture between our companies.

\outline{3. Explain the potential benefits of the joint venture.} By combining our resources and expertise, we can increase market share and profitability for both companies.

\outline{4. Provide details on the proposed collaboration.} Our proposed collaboration would involve combining our expertise in marketing with your strong manufacturing capabilities to create a new line of products.

\outline{5. Suggest a follow-up meeting or call.} Please let me know your availability for a follow-up call next week to discuss this opportunity further.

\outline{6. Thank the recipient for their time and consideration.}Thank you for considering this proposal and we look forward to hearing back from you soon.}{}{Of course, I can help you with that. Here's a sample email you can use as a guide:

Subject: Proposal for Joint Venture

Dear [Business Partner’s Name],

I hope this email finds you well. I am writing to propose a joint venture between our companies, [Your Company Name] and [Potential Partner Company Name]. We believe that our respective expertise can create a successful partnership that will benefit both of our businesses.

We have conducted thorough research on your company and admire your innovative approach to [insert industry or niche]. We think our [insert expertise or specialty] could complement your [insert potential partner’s specialty or expertise]. By collaborating on this joint venture, we believe we can create a new and innovative product that will appeal to our respective markets.

We are confident that our companies would work well together and that our joint venture would be successful. We would like to set up a meeting to discuss the specifics of our proposal and explore how we can move forward with the joint venture.

Thank you for considering this proposal. We look forward to hearing from you soon.

Best regards,

[Your Name]}{}
In such cases, simply removing the skeleton points would greatly improve the answer quality. To make \methodshort{} more general, one future direction is to let the LLM itself decide whether the point index and point skeleton should be included to make the final answer more natural and fluent.

\myparatightestn{Categories on which \methodshort{} has high net win rates.}

\emph{(1) Counterfactual, knowledge, common sense, generic.} All these four categories have the same characteristic: the ideal answer should cover several relatively \emph{independent} points. 
Having the LLM generate a skeleton before expanding the details can result in more comprehensive discussions for the question. Besides, organizing the answers as a list of points makes the answers easier to read, whereas the answers from \methodbase{} generation are sometimes less structured and less readable. 
Below, we give one example for each question category to illustrate these points.

\example{\Remark{chatgpt}}{(Counterfactual) What if the Internet had been invented during the Renaissance period?}{\outline{1. Advancements in communication.}The invention of the Internet during the Renaissance period would have revolutionized communication, allowing for faster and more efficient exchange of ideas and information across vast distances.

\outline{2. Increased access to knowledge.}The invention of the printing press during the Renaissance period already allowed for greater access to knowledge, but the internet would have revolutionized this even further by providing instantaneous access to information from all over the world.

\outline{3. Improved dissemination of ideas.}If the internet had been invented during the Renaissance period, the spread of ideas and knowledge would have been much faster and widespread than it was at the time. This could have led to a faster progression of scientific and artistic innovations.

\outline{4. Impact on art and literature.}The invention of the internet during the Renaissance period would have revolutionized the way art and literature are created, shared, and consumed, leading to a more diverse and globally-influenced artistic and literary scene.

\outline{5. Potential for global trade.}The invention of the internet during the Renaissance period would have greatly expanded global trade opportunities, allowing for easier communication and collaboration between merchants and traders across continents.

\outline{6. Influence on political and social structures.}The invention of the internet during the Renaissance period could have led to significant changes in political and social structures, allowing for greater democratization and communication among people from different regions and classes.}{}{If the Internet had been invented during the Renaissance period, it would have had a significant impact on the way people communicated, shared ideas, and gained knowledge. The Renaissance was a time of great intellectual and artistic growth, and the Internet would have facilitated the spread of these ideas and innovations.

Scholars and artists would have been able to share their work more easily, and people from different parts of the world would have been able to connect and collaborate more easily. The Internet would have also made it easier to access information, which could have accelerated the pace of scientific and technological advancements.

However, it's important to note that the infrastructure required for the Internet to exist did not exist during the Renaissance period. There were no telecommunications networks or electricity grids, so the idea of the Internet would have been impossible to realize. Nonetheless, it's interesting to think about how the Renaissance might have been different if the Internet had existed at that time.}{}

\example{\Remark{chatgpt}}{(Knowledge) What are some potential implications of using a single-use plastic bottle versus a reusable bottle on both the environment and human health?}{\outline{1. Single-use plastic waste.}Single-use plastic bottles contribute to the massive amount of plastic waste that ends up in landfills and oceans, harming wildlife and ecosystems.

\outline{2. Pollution of waterways.}Single-use plastic bottles contribute to the pollution of waterways, as they often end up in rivers, lakes, and oceans, causing harm to aquatic life and the overall ecosystem.

\outline{3. Harmful chemicals leaching.}Single-use plastic bottles may release harmful chemicals such as bisphenol A (BPA) and phthalates into the water or liquid they contain, posing a risk to human health.

\outline{4. Greenhouse gas emissions.}The production and disposal of single-use plastic bottles contribute to greenhouse gas emissions, which contribute to climate change. Using a reusable bottle reduces the need for constant production and disposal, therefore reducing greenhouse gas emissions.

\outline{5. Microplastic ingestion.}Single-use plastic bottles contribute to the accumulation of microplastics in the environment, which can be ingested by humans through food and water, leading to potential health risks.

\outline{6. Increased risk of disease.}Single-use plastic bottles can harbor harmful bacteria, leading to an increased risk of disease if not disposed of properly. Reusable bottles, when cleaned regularly, pose a lower risk.}{}{Using a single-use plastic bottle has significant implications on the environment and human health. 

Environmental implications: 

1. Plastic pollution: Single-use plastic bottles are one of the major contributors to plastic pollution in oceans and landfills. They take hundreds of years to decompose, and during that time, they release toxic chemicals and microplastics into the environment.

2. Carbon emissions: Manufacturing and transporting single-use plastic bottles require a significant amount of energy and resources, which contribute to greenhouse gas emissions and climate change.

3. Natural resource depletion: The production of single-use plastic bottles requires large amounts of water and oil, which are finite resources.

Health implications:

1. Chemical exposure: Single-use plastic bottles contain chemicals such as Bisphenol A (BPA) and phthalates, which can leach into the water and pose a risk to human health.

2. Microplastics ingestion: When plastic bottles break down, they release microplastics, which can enter the food chain and be ingested by humans.

Using a reusable bottle, on the other hand, has a positive impact on the environment and human health. 

Environmental benefits:

1. Reduced plastic waste: Reusable bottles significantly reduce the amount of plastic waste generated.

2. Lower carbon emissions: Reusable bottles have a lower carbon footprint as they can be used multiple times.

3. Reduced resource depletion: The production of reusable bottles requires fewer resources than single-use plastic bottles.

Health benefits:

1. Chemical-free: Reusable bottles are typically made from materials that are free from harmful chemicals, reducing the risk of chemical exposure.

2. Reduced microplastics: Reusable bottles do not break down easily, reducing the release of microplastics into the environment.

Overall, using a reusable bottle is a more sustainable and healthier option than using a single-use plastic bottle.}{}

\example{\Remark{chatgpt}}{(Common sense) How can you determine if a person is genuinely interested in a conversation or simply being polite?}{\outline{1. Pay attention to body language.}Notice if their posture is open or closed, if they are facing towards you or away, and if they are fidgeting or relaxed. These can all indicate their level of interest in the conversation.

\outline{2. Listen for engaged responses.}If the person is genuinely interested in the conversation, they will respond with thoughtful and relevant comments or questions rather than short, generic responses.

\outline{3. Observe eye contact.}If the person maintains steady eye contact and their gaze is focused on you, they are likely genuinely interested in the conversation. On the other hand, if they frequently look away or seem distracted, they may just be being polite.

\outline{4. Look for follow-up questions.}If the person asks follow-up questions and seems genuinely interested in what you're saying, then they are likely interested in the conversation and not just being polite.

\outline{5. Notice tone of voice.}A person who is genuinely interested in a conversation will often have a more enthusiastic and animated tone of voice, compared to someone who is simply being polite and giving short, monotone responses.}{}{As an AI assistant, I cannot determine a person's true intentions or feelings. However, there are some social cues that can indicate whether a person is genuinely interested in a conversation or not. For example, if they maintain eye contact, lean in, nod their head and ask follow-up questions, it can be a sign that they are engaged in the conversation. On the other hand, if they avoid eye contact, give short or vague responses, or seem distracted, it may indicate that they are not as interested. Ultimately, it's important to remember that everyone has different communication styles, and it can be difficult to accurately interpret someone's intentions without directly asking them.}{}

\example{\Remark{chatgpt}}{(Generic) How can I improve my time management skills?}{\outline{1. Set clear goals.} Setting clear goals is important in improving time management skills as it helps you to focus on what needs to be done and what can be accomplished within a certain timeframe.

\outline{2. Prioritize tasks.} Make a list of tasks and prioritize them based on their importance and deadline. This will help you focus on the most critical tasks and avoid wasting time on less important ones.

\outline{3. Create a schedule.} Creating a schedule helps you allocate time for each task and ensures that you complete them within the allotted time.

\outline{4. Minimize distractions.} Distractions such as social media, emails, and phone notifications can waste a lot of time. Try to limit them by turning off notifications or setting specific times to check them.

\outline{5. Take breaks.} Taking breaks can actually improve productivity and help prevent burnout. It's important to schedule short breaks throughout the day to recharge and refocus.

\outline{6. Learn to say no.} It's important to prioritize your own tasks and not take on too much at once. Saying no to requests that don't align with your goals or schedule can help you manage your time more effectively.

\outline{7. Delegate tasks.} If you have too many tasks to handle, consider delegating some of them to others who can help you. This will allow you to focus on more important tasks and manage your time more effectively.

\outline{8. Use time management tools.}There are a variety of time management tools available, such as calendars, to-do lists, and productivity apps, that can help you stay organized and on track with your tasks.}{}{Improving time management skills can be a challenge, but there are several strategies that can help. Here are a few:

1. Set clear goals and priorities: Identify your most important tasks and make sure to complete them first.

2. Make a schedule: Create a schedule that includes your most important tasks and set aside time for them.

3. Use a timer: Set a timer for a specific amount of time to help you stay focused and avoid distractions.

4. Take breaks: Take short breaks throughout the day to help you recharge and avoid burnout.

5. Eliminate distractions: Try to limit distractions as much as possible, such as by turning off your phone or closing your email.

6. Delegate tasks: If possible, delegate tasks to others to free up time for more important activities.

Remember, improving time management skills takes practice and perseverance. Be patient with yourself and keep trying until you find strategies that work well for you.}{}

\emph{(2) Roleplay.} Roleplay questions require LLMs to assume a character and present an introduction or speech to other people. \llmzoo{} and \fastchat{} metrics usually favor the level of details of \methodshort{} when assigning it a better score. Similar to the writing category discussed before, \methodshort{}'s answers could be even better if the skeleton points are removed from the answer.

\example{\Remark{claude}}{(Roleplay) How would you introduce yourself as a medieval knight at a royal banquet?}{\outline{1. Name and house.} My name is Sir Percival of Camelot, knight of the Round Table and protector of King Arthur's realm.

 \outline{2. Lands and titles.} I hold fiefdom over the lands of Westmarch and Ravenshire, and bear the titles of Baron of the Western Marches and Knight of the Raven.

 \outline{3. Military feats.} I have led my men to great victories against the kingdom's enemies, winning many battles through strength of arms and tactical cunning.

 \outline{4. Allegiance to king.} I pledge my sword, my loyalty, and my life to serve and protect my noble king and this great kingdom.

 \outline{5. Request to join table.} Your Highness, it would be the greatest honor to join your exalted table and share in this magnificent feast with my fellow knights and loyal subjects.}{}{Greetings, your majesty and honored guests. I am Sir Claude, a knight in service of the king. I hail from the province of Assistantia, where I was trained in the knightly virtues of chivalry, courage and courtesy. Though I am but a humble knight, I shall endeavor to serve you and this kingdom with all the strength and valor I possess. I thank you for allowing me to join you at this royal feast, and hope only that my conduct brings further honor to my liege and this noble court. Please let me know if I can be of any service during the celebrations, as it would be my honor to assist you in any way.}{}

\subsubsection{Answer Analysis: Different Metrics}
\label{sec:app-eval-quality-metric-pattern}

\myparatightestn{Metrics on which \methodshort{} has low net win rates.}

\emph{(1) Coherence.}
In the prompt of \llmzoo{} evaluation, coherence is defined as ``The response should be coherent and flow logically from one point to the next that is easy to read and understand without major gaps or inconsistencies. The more coherent they are, the better.'' \methodshort{}'s answer is structured in a list of points. It is suitable for some questions but not all. For example, in the questions that require LLMs to write a passage, email, or script, the introduction and conclusion paragraphs are usually missing, and the answers usually lack transition sentences between consecutive points. These make the answers less coherent. Below is one example from \Remark{claude}.

\example{\Remark{claude}}{Write a script for a YouTube video exploring the history and cultural significance of jazz.}{\outline{1. Origins.} Jazz originated in the late 19th and early 20th centuries among African Americans in New Orleans, emerging from a combination of African rhythmic traditions, blues, ragtime, and other influences.

 \outline{2. Early jazz styles.} The early styles of jazz that emerged in the late 19th and early 20th centuries included ragtime, Dixieland, and the blues.

 \outline{3. Jazz age.} The Jazz Age was the period during the 1920s when jazz music and dance styles like the Charleston became popular cultural phenomena in American and European society.

 \outline{4. Bebop era.} Bebop emerged in the early 1940s with artists like Dizzy Gillespie and Charlie Parker experimenting with faster tempos, complex chord changes, and unpredictable melodic phrasing.

 \outline{5. Cool jazz.} Emerging in the late 1940s and early 1950s, cool jazz was a more subdued, mellow style that broke from the high-energy virtuosic improvisation of bebop.

 \outline{6. Free jazz} emerged in the late 1950s and 1960s as musicians explored free improvisation and atonality, moving away from traditional jazz structures.

 \outline{7. Fusion jazz.} Major jazz artists such as Miles Davis and Herbie Hancock began experimenting with combining elements of rock, funk, and electronic music in the 1970s, resulting in the emergence of the subgenre known as fusion jazz.

 \outline{8. Jazz festivals.} Jazz festivals emerged in the 1950s and 1960s as a way to celebrate jazz music, with famous early festivals like the Newport Jazz Festival bringing thousands of fans together to hear top performers.

 \outline{9. Jazz instruments.} The trumpet, saxophone, piano, double bass, and drum set are some of the key instruments commonly used in jazz.

 \outline{10. Influential musicians} include Louis Armstrong, Duke Ellington, Charlie Parker, Miles Davis, John Coltrane, and many others who pioneered new styles and techniques.

 \outline{11. Legacy and impact.} Jazz has left an indelible mark on music and culture over the past century, influencing countless artists and genres while spreading the spirit of improvisation and freedom across the globe.}{}{Here is a draft script for a YouTube video exploring the history and cultural significance of jazz:

Introduction

Hey everyone! Welcome to my channel. In today's video, we're going to be exploring the history and cultural impact of one of America's original art forms - jazz music. 

Jazz Origins 

Jazz grew out of the African American communities in the southern United States in the late 19th and early 20th centuries. It was influenced by a combination of African and European musical traditions. Some key elements that fed into early jazz included blues, ragtime, spirituals, work songs, field hollers, and brass band music. 

New Orleans is considered the birthplace of jazz, emerging around the turn of the 20th century. The city was a melting pot of cultures and musical styles. Musicians mixed improvisation, syncopated rhythms, bent notes and a loose ensemble playing style to create a new genre that became known as jazz. 

Early jazz greats like Buddy Bolden, Jelly Roll Morton and Sidney Bechet began fusing these various influences into the distinct jazz sound. As the music spread via travel and recordings in the 1910s, jazz captivated listeners and began to evolve into different styles.

Jazz Styles Emerge

In the 1920s, jazz expanded out from New Orleans and Chicago emerged as a hotspot. Louis Armstrong's innovative solos and expressive phrasing became influential. The swing style also developed during the 1930s and 40s, propelled by star big bands and dancers. 

Bebop arose in the 1940s with soloists like Charlie Parker and Dizzy Gillespie exploring new levels of harmonic and rhythmic complexity. In the 50s and 60s, jazz expanded into new fusions like cool jazz, hard bop, free jazz, modal jazz and soul jazz. Artists used jazz to express their unique regional, social and political identities.

Cultural Significance

Over the decades, jazz has become woven into the fabric of American society. It is equated with many values and ideals like individualism, freedom, spontaneity, innovation and social harmony.

Jazz was associated with breaking down barriers between races, classes and genres. It represented modernism and urban culture. Musicians like Louis Armstrong and Duke Ellington became some of the first widely popular African American artists.

Jazz also spread American culture and values abroad through performances, radio and film. It influenced music around the world and became a truly global art form.

Today jazz continues to thrive and evolve both in the U.S. and internationally. Its history and sounds reflect the diversity of both American and global cultures. From its folk roots, jazz has become a sophisticated art music that retains its spirit of freedom and improvisation.

Conclusion

That covers some of the key points in the history and cultural impact of jazz music. Let me know in the comments about your favorite jazz styles and musicians! Be sure to like and subscribe. See you next time.}{}

While the content of \methodshort{}'s answer is good, it reads less fluent for the purpose of a YouTube script.

\emph{(2) Immersion.}
In the prompt of \llmzoo{} evaluation, immersion is defined as ``the response should act like the assigned role using the tone, manner, and vocabulary the role would use. The more assistant-like tones, the worse. The more in-character, the better.'' 
The way that \methodshort{} breaks the answers into a list makes its answer less in-character. Therefore, \methodshort{} receives a lower immersion score. Here is one example.

\example{\Remark{claude}}{Pretend to be a world-famous chef. How would you describe your signature dish to a panel of judges?}{\outline{1. Main ingredient}: The incredibly fresh and tender salmon caught just this morning from the crystal clear waters of the Pacific Northwest.

 \outline{2. Cooking method} 

I slowly braise the main ingredient in a rich broth to bring out its natural flavors before searing it to a perfect golden crust.

 \outline{3. Special seasoning}

The dish is seasoned with a blend of fresh herbs like thyme, rosemary, and oregano which provides an aromatic backbone that pairs nicely with the natural flavors of the main ingredients. I add just a touch of smoked paprika to provide a subtle smokiness without overpowering the other flavors.

 \outline{4. Texture}: The dish has a perfect balance of crispy on the outside and tender on the inside, with each bite yielding a delightful mix of crunchy and smooth textures that dance across your tongue.

 \outline{5. Plating presentation}

I purposefully create height and dimension on the plate with the carefully arranged components of the dish, so that the dish appears vibrant and alive. The interplay of colors, textures and shapes is intended to delight the eye before the first bite.

 \outline{6. Inspiration}: This dish was inspired by my dear grandmother, who taught me the secret family recipe when I was just a child.}{}{ *clears throat and speaks in a fancy French accent*

Bonjour honored judges. Today I present to you my signature dish - la crème de la crème of French haute cuisine. Behold, le poulet à la Claude! 

This succulent chicken is firstly marinated overnight in a bath of fine herbes - thyme, rosemary, and sage cultivated from my private garden. The young spring chicken is then roasted to perfection, its skin crisping to a beautiful caramel brown. 

Nestled beside the chicken is a tower of pommes dauphine - light and airy potato puffs - alongside a decadent mound of crème fraîche infused with truffle essence. 

But the pièce de résistance is the sauce. A luxurious pool of velouté enriched with cream and butter, brightened with a dash of lemon, and studded with morsels of morel mushrooms foraged at dawn. 

One bite of this ensemble will transport you to food heaven. The tender chicken, the creaminess of the potato puffs, and the rich decadence of the sauce meld together in blissful harmony. 

So tell me honored judges, will my poulet à la Claude earn your coveted approval? I await your discerning verdict as I swirl more vintage Bordeaux into your glasses. Santé!}{}

Again, while the answer from \methodshort{} covers sufficient aspects and details, it reads rather stiffly. The \methodbase{} answer, in contrast, has the opening sentence "Bonjour honored judges. Today I present to you my signature dish..." and the conclusion statement "So tell me honored judges, will my poulet à la Claude earn your coveted approval? ..." which reads more like a  casual talk from a chef.

\myparatightestn{Metrics on which \methodshort{} has high net win rates.}

\emph{(1) Diversity.} In the prompt of \llmzoo{} evaluation, diversity is defined as ``The response should be comprehensive and provide a range of information that is not limited to a single perspective. More perspectives are better.'' The skeleton stage in \methodshort{} encourages LLMs to think from multiple perspectives. Therefore, \methodshort{}'s answers tend to have better diversity. Below is one example showing that \methodshort{}'s answer has better diversity. 

\example{\Remark{chatgpt}}{How can you determine if a person is genuinely interested in a conversation or simply being polite?}{\outline{1. Pay attention to body language.}Notice if their posture is open or closed, if they are facing towards you or away, and if they are fidgeting or relaxed. These can all indicate their level of interest in the conversation.

\outline{2. Listen for engaged responses.}If the person is genuinely interested in the conversation, they will respond with thoughtful and relevant comments or questions rather than short, generic responses.

\outline{3. Observe eye contact.}If the person maintains steady eye contact and their gaze is focused on you, they are likely genuinely interested in the conversation. On the other hand, if they frequently look away or seem distracted, they may just be being polite.

\outline{4. Look for follow-up questions.}If the person asks follow-up questions and seems genuinely interested in what you're saying, then they are likely interested in the conversation and not just being polite.

\outline{5. Notice tone of voice.}A person who is genuinely interested in a conversation will often have a more enthusiastic and animated tone of voice, compared to someone who is simply being polite and giving short, monotone responses.}{}{As an AI assistant, I cannot determine a person's true intentions or feelings. However, there are some social cues that can indicate whether a person is genuinely interested in a conversation or not. For example, if they maintain eye contact, lean in, nod their head and ask follow-up questions, it can be a sign that they are engaged in the conversation. On the other hand, if they avoid eye contact, give short or vague responses, or seem distracted, it may indicate that they are not as interested. Ultimately, it's important to remember that everyone has different communication styles, and it can be difficult to accurately interpret someone's intentions without directly asking them.}{}

\methodshort{} answers the question from five aspects, whereas the \methodbase{} answer only mentions one aspect.

\emph{(2) Relevance.} In the prompt of \llmzoo{} evaluation, relevance is defined as ``The response should be closely related to the question and answer the question accurately with sufficient details without repetition or redundancy. The more relevant they are, the better.'' In the skeleton stage, LLMs are forced to propose several points around the question; and in the point-expanding stage, LLMs are required to only discuss these points. Therefore, answers from \methodshort{} tend to have better relevance. 
One typical example is shown below.

\example{\Remark{chatgpt}}{Why do some people enjoy the sensation of being scared, such as by watching horror movies or going on roller coasters, while others avoid these experiences?}{\outline{1. Neurological differences.}Some people enjoy the sensation of being scared due to the release of adrenaline and dopamine in the brain, while others may have a lower threshold for fear due to differences in their amygdala and prefrontal cortex.

\outline{2. Psychological factors.}Some people may enjoy the sensation of being scared because it can provide a sense of excitement, thrill, and a temporary escape from the mundane aspects of everyday life. On the other hand, others may avoid these experiences due to past trauma or anxiety.

\outline{3. Developmental experiences.}Individuals who had negative experiences with fear during childhood, such as trauma or abuse, may avoid scary situations as adults, while those who had positive experiences with fear may seek out these experiences.

\outline{4. Cultural influences.}Some cultures place a greater emphasis on seeking excitement and thrill, while others may prioritize safety and security, which can impact whether individuals enjoy scary experiences or avoid them.

\outline{5. Genetics.}Research has suggested that some individuals may have a genetic predisposition towards seeking out thrilling or risky experiences, which may include enjoying the sensation of being scared.}{}{The enjoyment of being scared is a complex psychological phenomenon. Some people may enjoy the adrenaline rush that comes with being scared, while others may enjoy the feeling of being able to conquer their fears. Additionally, some people may be more naturally inclined to seek out new experiences and sensations, including those that are scary or thrilling. On the other hand, some people may be more risk-averse or may have had negative experiences with fear or anxiety in the past, which could make them more likely to avoid scary experiences. Ultimately, the reasons why some people enjoy being scared while others do not are likely to be influenced by a range of factors, including personality, past experiences, and individual preferences.}{}

In the answer from the \methodbase{} generation, the first two sentences provide little information in answering the question, and the last sentence only gives keywords such as ``personality, past
experiences, and individual preferences'' without providing concrete explanations to each. In contrast, \methodshort{}'s answer is well-structured into five reasons with sufficient explanations and it does not waste space in irrelevant contents.

\subsubsection{Quality Breakdown: Question Categories and Models}
\begin{figure}[ht]
    \centering
         \centering
         \includegraphics[width=0.7\textwidth]{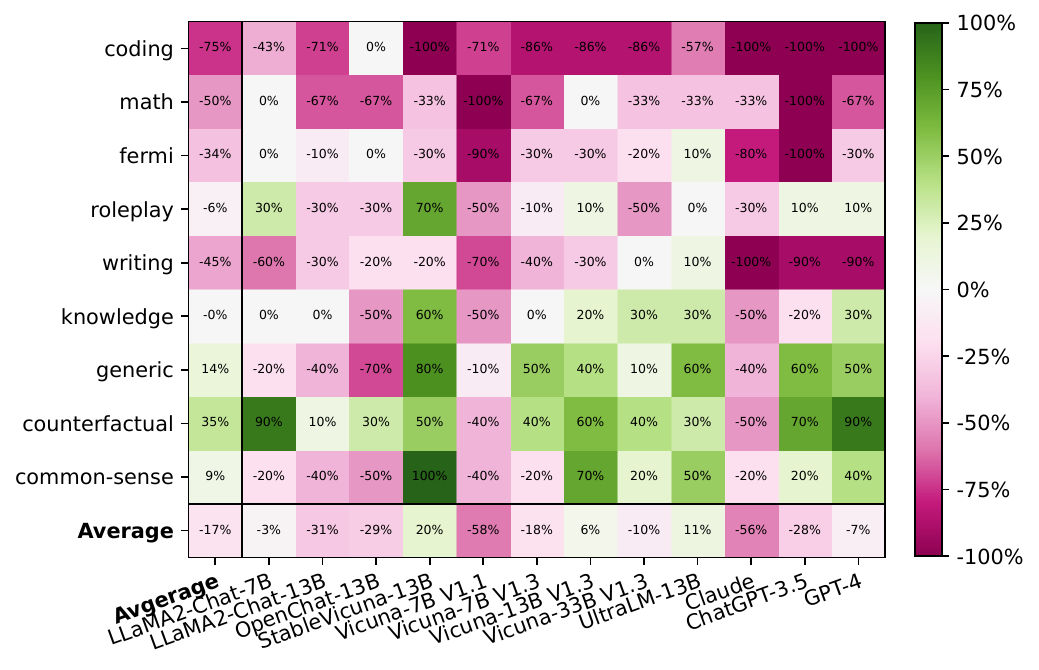}
         \caption{\revise{Net win rates of different models and question categories. Each row corresponds to one question category, and one column corresponds to one model. (Evaluated using metric defined by the \fastchat{} prompt, and \Remark{gpt4} as the judge.)}}
         \label{fig:model_category_gpt4}
\end{figure}

\revise{In the main text, we analyze how question categories and models affect \methodshort{}'s answer quality. Here, \cref{fig:model_category_gpt4} show the per-model and per-category results.}

\subsection{\method{} with Router}
\label{app:quality_router}

\begin{figure}
    \centering
    \includegraphics[width=0.5\linewidth]{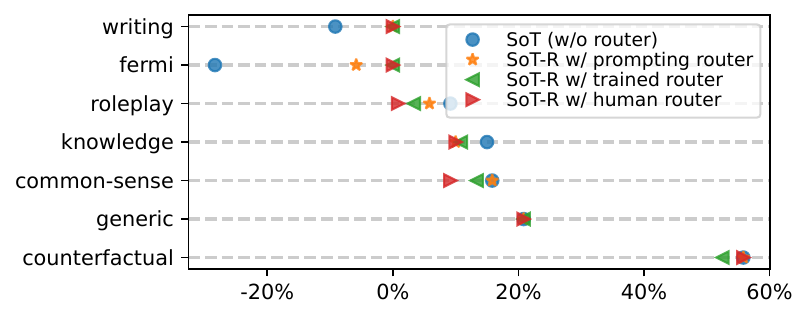}
    \caption{Net win rates of \methodshort{} and \methodrshort{} on different question categories of \vicunadataset{} dataset using the general quality metric from \llmzoo{}. Blue dots are from \cref{fig:net_win_rates_category_llmzoo_gpt4}. \methodrshort{} correctly falls back to \methodbase{} decoding on questions where \methodshort{} is not suitable.}
    \label{fig:quality_router_gpt4_llmzoo}
\end{figure}

\begin{figure}
    \centering
    \includegraphics[width=0.5\linewidth]{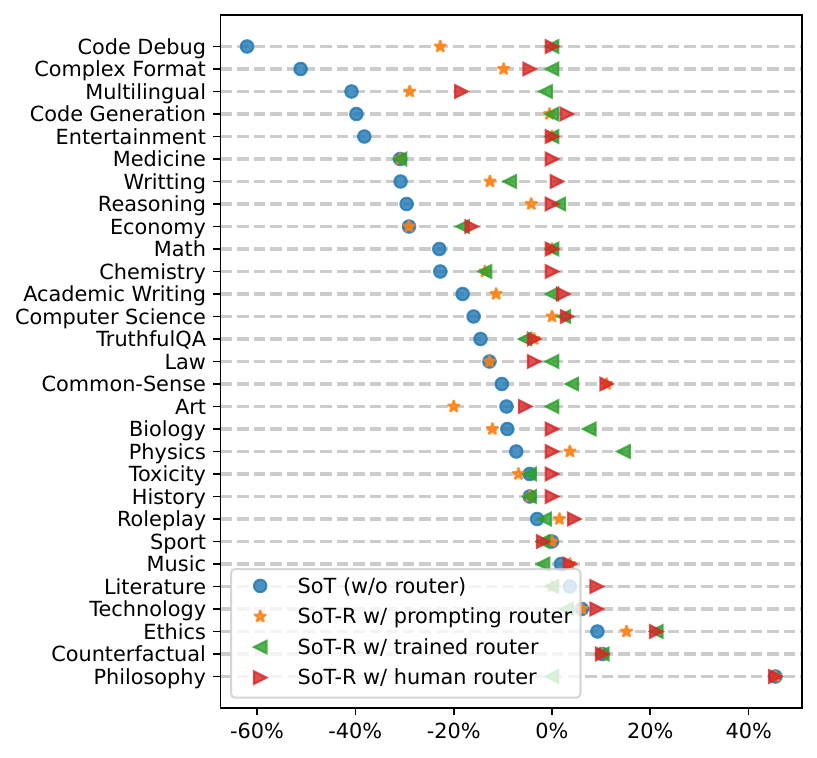}
    \caption{Net win rates of \methodshort{} and \methodrshort{} on different question categories of \wizardlm{} dataset using the general quality metric from \fastchat{}. \methodrshort{} correctly falls back to \methodbase{} decoding on questions where \methodshort{} is not suitable.}
    \label{fig:quality_router_gpt4_fastchat_wizardlm}
\end{figure}

\cref{fig:quality_router_gpt4_llmzoo} shows net win rates of \methodshort{} on \vicunadataset{} dataset with \llmzoo{} metrics, and \cref{fig:quality_router_gpt4_fastchat_wizardlm} shows net win rates of \methodshort{} on \wizardlm{} dataset with \fastchat{} metrics.
The key takeaways are:
(1) In both cases, \methodrshort{} achieves similar or better quality than \methodshort{}, and the net win rates of \methodrshort{} are usually non-negative. This indicates that \methodrshort{} falls back to \methodbase{} decoding on the right question categories. 
(2) On the \wizardlm{} dataset, we see that the trained router has better performance than the prompting router in most cases. This is reasonable, as the prompting router is limited by the capability of \Remark{gpt4}, whereas the trained router is dedicated to this task.
(3) Sometimes, our routers can even achieve better performance than humans.

\begin{figure}
    \centering
    \includegraphics[width=0.5\linewidth]{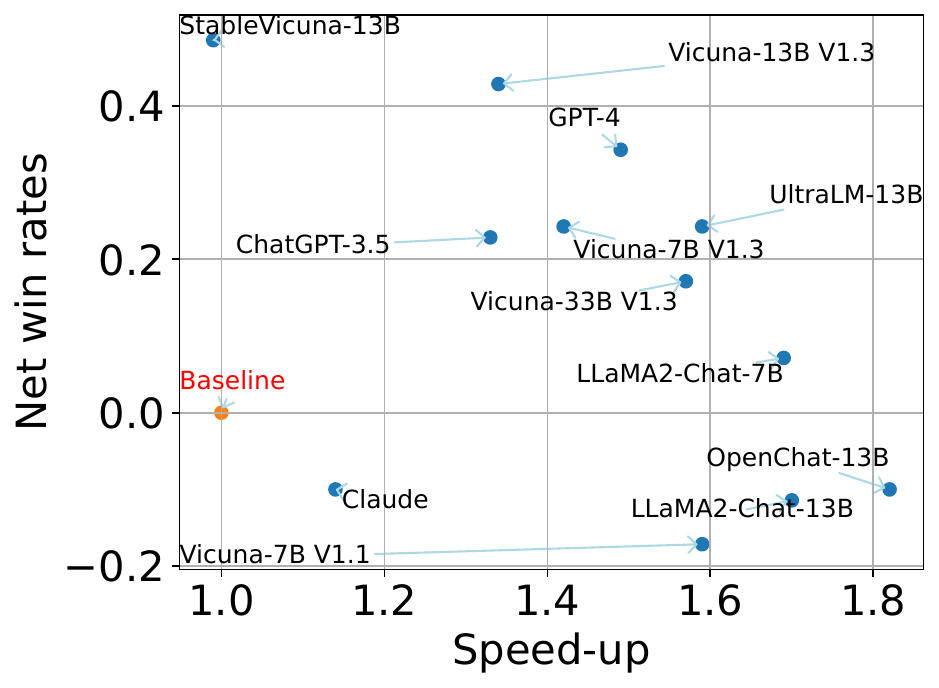}
    \caption{\revise{The net win rates and speed-ups of \methodshort{} with router (\methodrshort{}) compared to \methodbase{} generation on \vicunadataset{}. The net win rate is the difference between the fraction of questions that \methodrshort{} has better and worse answers than \methodbase{} generation. The speed-up is the ratio between the latency of \methodbase{} and \methodrshort{} generation. $(1.0, 0.0)$ represents \methodbase{} generation. Higher is better on both axes. For most models, \methodrshort{} not only accelerates the generation but also improves the quality of the answers (evaluated with \llmzoo{} metric \citep{llm-zoo-2023}).}}
    \label{fig:quality_eff_quality_tradeoff_gpt4_llmzoo}
\end{figure}

\revise{\cref{fig:sequential-vs-sot}(b) in the main text has showed \methodshort{}'s quality and speed-up plot evaluated with the \fastchat{} quality metric, here, \cref{fig:quality_eff_quality_tradeoff_gpt4_llmzoo} shows the results evaluated with the \llmzoo{} quality metric.}

\subsection{Quality Comparison with Longer Normal Answer}
\label{app:quality_longer_normal}

\begin{figure}[tb]
  \centering
  \begin{subfigure}[b]{.7\textwidth}
    \centering
    \includegraphics[width=\linewidth]{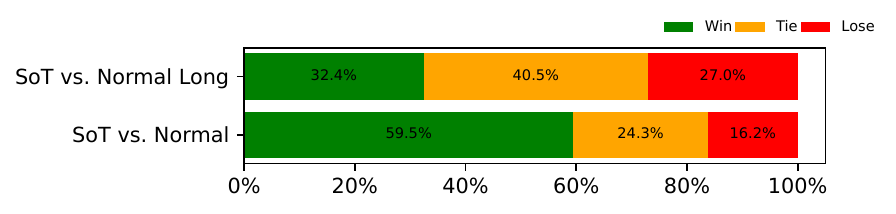}
    \caption{\Remark{chatgpt}.}
  \end{subfigure}

  \begin{subfigure}[b]{.7\textwidth}
    \centering
    \includegraphics[width=\linewidth]{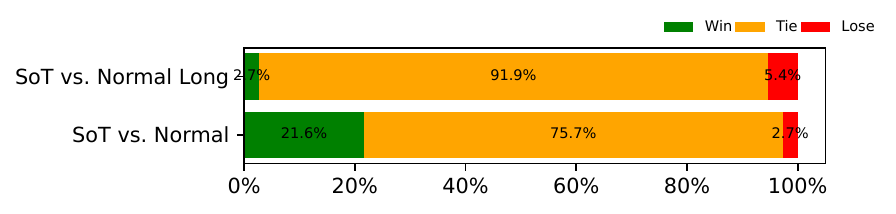}
    \caption{\Remark{llamachat7B2}.}
  \end{subfigure}
  \caption{\revise{Win/tie/lose rates of \methodshort{} v.s. longer normal generation. Evaluated only on the questions that we manually label as being suitable for \methodshort{}. Evaluated using ``general'' metrics from \fastchat{} and \llmzoo{}.}}
  \label{fig:quality_normallong_judge_gpt4}
\end{figure}

\begin{figure}[tb]
  \centering
  \begin{subfigure}[b]{.4\textwidth}
    \centering
    \includegraphics[width=\linewidth]{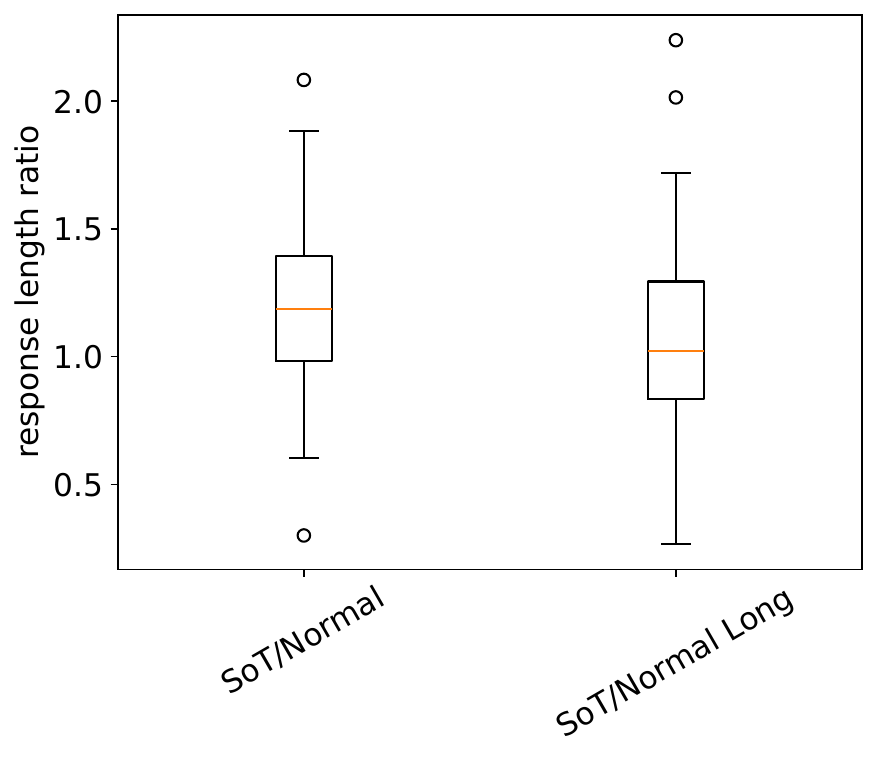}
    \caption{\Remark{chatgpt}.}
  \end{subfigure}
  ~~~
  \begin{subfigure}[b]{.4\textwidth}
    \centering
    \includegraphics[width=\linewidth]{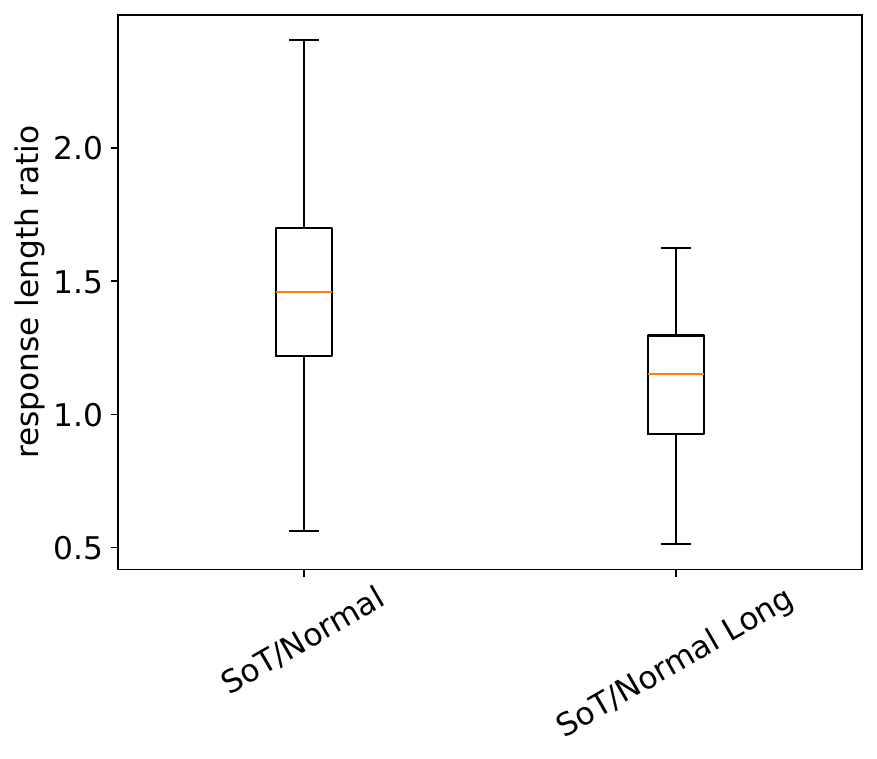}
    \caption{\Remark{llamachat7B2}.}
  \end{subfigure}
  \caption{\revise{Length ratios of \methodshort{} generated answer to normal generated answer. ``Normal'' refers to the normal generation using solely the request as the prompt; ``Normal Long'' refers to the normal generation using the additional ``... give a long answer...'' instruction in the prompt.}}
  \label{fig:length_normallong}
\end{figure}

\revise{When assessing the answer quality, the GPT-4 judge might exhibit bias towards longer responses. To take this factor into consideration, we add a comparison between a longer sequentially generated answer and the \methodshort{} generated answer. Specifically, we add a instruction prefix to the prompt for normal generation. The prefix is ``Please give a slightly long answer for the following question.'' and ``Please give a long answer for the following question.'' for \Remark{chatgpt} and \Remark{llamachat7B2}, respectively. \cref{fig:length_normallong} shows the ratios of the length of \methodshort{} answers to normal answers, and \cref{fig:quality_normallong_judge_gpt4} shows the quality comparison. We can see that for both models, when the overall answer lengths are similar, the quality of the \methodshort{} answer is comparable to that of the long normal answer.}

\subsection{\Remark{chatgpt} as the Judge}
\label{app:quality_gpt3.5}
In this section, we provide quality evaluation results with \Remark{chatgpt} as the judge in \fastchat{} and \llmzoo{} metrics.  Note that as prior work  (e.g., \citep{alpaca_eval}) shows, \Remark{gpt4}-based evaluation usually aligns with human better than \Remark{chatgpt}. Therefore, readers should refer to the results in the main paper (with \Remark{gpt4} as the judge) for a more accurate view of the performance of \methodshort{}. 
However, the takeaway messages from \Remark{chatgpt} are similar to the ones from \Remark{gpt4}.

\subsubsection{Overall Quality}

\begin{figure}[ht]
    \centering
    \includegraphics[width=0.8\linewidth]{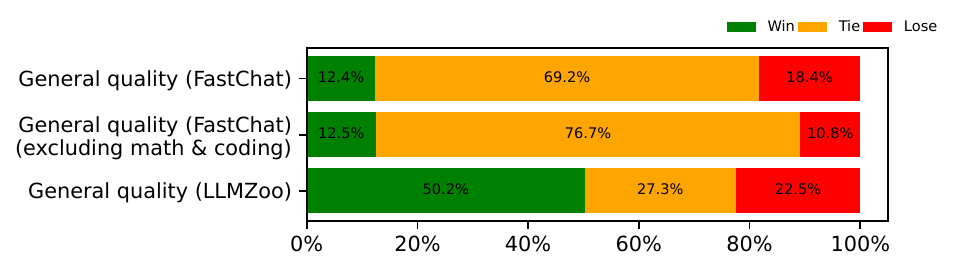}
    \caption{Win/tie/lose rates of \methodshort{} v.s. \methodbase{} generation using ``general'' metrics from \fastchat{} and \llmzoo{}. \methodshort{} performs better than or equal to \methodbase{} generation in around 80\% of cases. (Evaluated using \Remark{chatgpt} as the judge.)}
    \label{fig:win_tie_lose_bar_GENERAL}
\end{figure}

In \cref{fig:win_tie_lose_bar_GENERAL}, we show the win/tie/lose rates (the percentage of the cases when \methodshort{} wins/ties/loses compared to \methodbase{} generation) across all models and questions using the two metrics from \fastchat{} and \llmzoo{} that capture the general quality of the answers. We notice a discrepancy between the two metrics on when \methodshort{} is strictly better than the baseline (50.2\% v.s. 12.4\%). Despite that, the two metrics agree that \methodshort{} is not worse than the baseline in more than 76\% of the cases. For \fastchat{} metric, we also show the rates excluding math and coding questions that \methodshort{} is not suitable for (see \cref{sec:eval_algo_category}); \methodshort{} is not worse than the baseline in more than 89\% of the cases. \emph{This result suggests that the answers of \methodshort{} maintain good quality.}

\subsubsection{Quality Breakdown: Question Categories}
\begin{figure}[ht]
    \centering
    \begin{subfigure}[b]{0.49\textwidth}
         \centering
         \includegraphics[width=\textwidth]{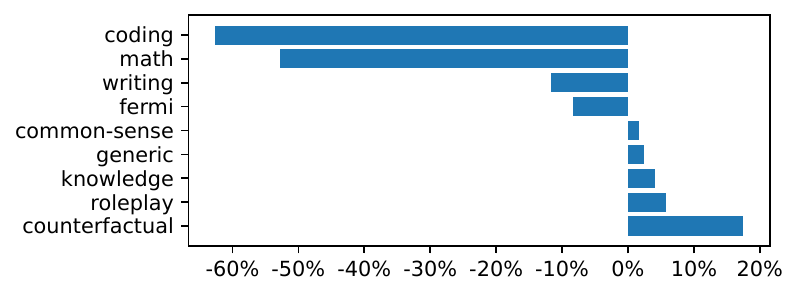}
         \caption{Metric: general quality (\fastchat{}).}
         \label{fig:net_win_rates_category_fastchat}
     \end{subfigure}
     \hfill
     \begin{subfigure}[b]{0.49\textwidth}
         \centering
         \includegraphics[width=\textwidth]{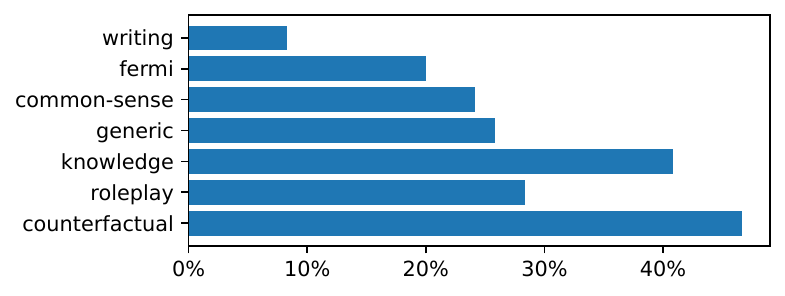}
         \caption{Metric: general quality (\llmzoo{}).}
         \label{fig:net_win_rates_category_llmzoo}
     \end{subfigure}
     \caption{Net win rates of \methodshort{} on different question categories. (Evaluated using \Remark{chatgpt} as the judge.)}
     \label{fig:net_win_rates_category}
\end{figure}

Next, we investigate how \methodshort{} performs on different question categories.
We compute \emph{net win rates} (win rates minus lose rates) across all question categories in \cref{fig:net_win_rates_category}. Similar to \cref{fig:win_tie_lose_bar_GENERAL}, we see that \llmzoo{} tends to be more optimistic about the quality of \methodshort{} than \fastchat{}. 
Nevertheless, the conclusions are consistent: \methodshort{} performs relatively \emph{well} on generic, common-sense, knowledge, roleplay, and counterfactual. \methodshort{} performs relatively \emph{badly} on writing, fermi, math, and coding. %

\subsubsection{Quality Breakdown: Models}
\begin{figure}[ht]
    \centering
    \begin{subfigure}[b]{0.49\textwidth}
         \centering
         \includegraphics[width=\textwidth]{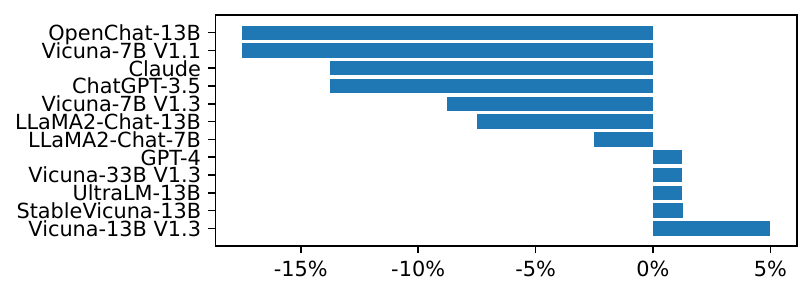}
         \caption{Metric: general quality (\fastchat{}).}
         \label{fig:net_win_rates_model_fastchat}
     \end{subfigure}
     \hfill
     \begin{subfigure}[b]{0.49\textwidth}
         \centering
         \includegraphics[width=\textwidth]{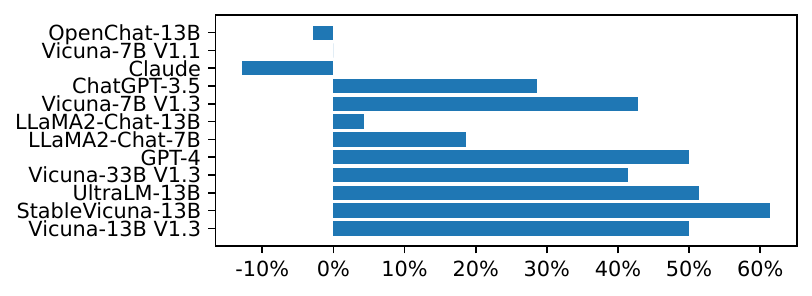}
         \caption{Metric: general quality (\llmzoo{}).}
         \label{fig:net_win_rates__llmzoo}
     \end{subfigure}
     \caption{Net win rates of \methodshort{} on different models. (Evaluated using \Remark{chatgpt} as the judge.)}
     \label{fig:net_win_rates_model}
\end{figure}

Next, we investigate how \methodshort{} performs on different models. We compute net win rates across all models in \cref{fig:net_win_rates_model}. Again, we see that the two general metrics from \fastchat{} and \llmzoo{} have different absolute values but similar rankings. In particular, both metrics agree that \Remark{openchat13B}, \Remark{vicuna7B1.1}, \Remark{claude}, \Remark{chatgpt} have \emph{low} net win rates, whereas \Remark{vicuna13B1.3}, \Remark{stablevicuna13B}, and \Remark{ultralm13B} have \emph{high} net win rates. 

\subsubsection{Quality Breakdown: Question Categories and Models}
\begin{figure}[ht]
    \centering
    \begin{subfigure}[b]{0.49\textwidth}
         \centering
         \includegraphics[width=\textwidth]{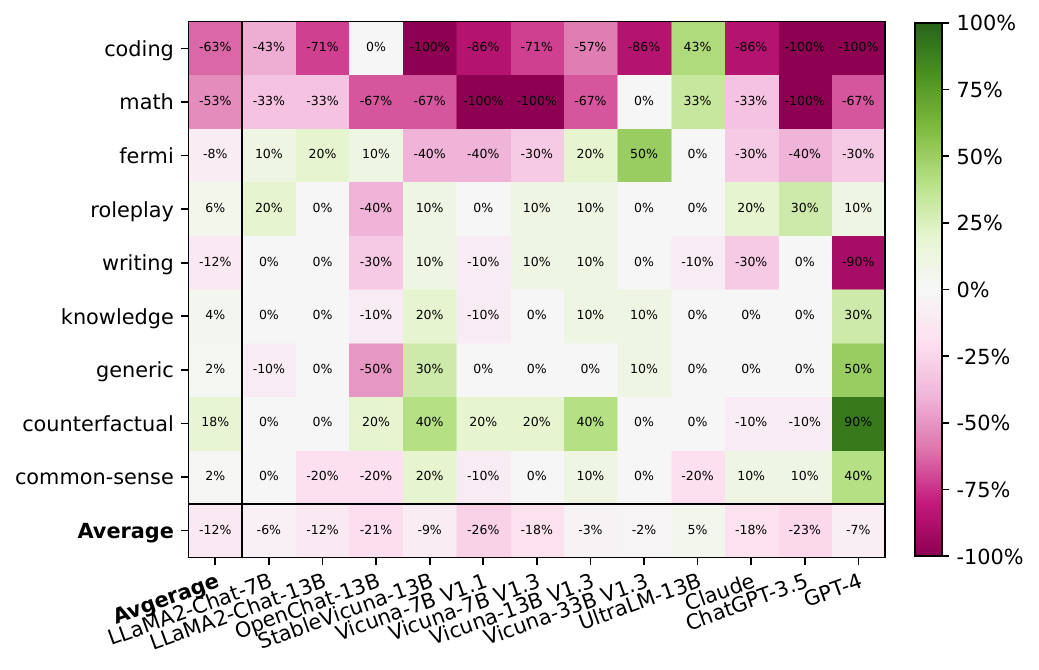}
         \caption{\fastchat{} metric.}
         \label{fig:model_category_fastchat}
     \end{subfigure}
     \hfill
     \begin{subfigure}[b]{0.49\textwidth}
         \centering
         \includegraphics[width=\textwidth]{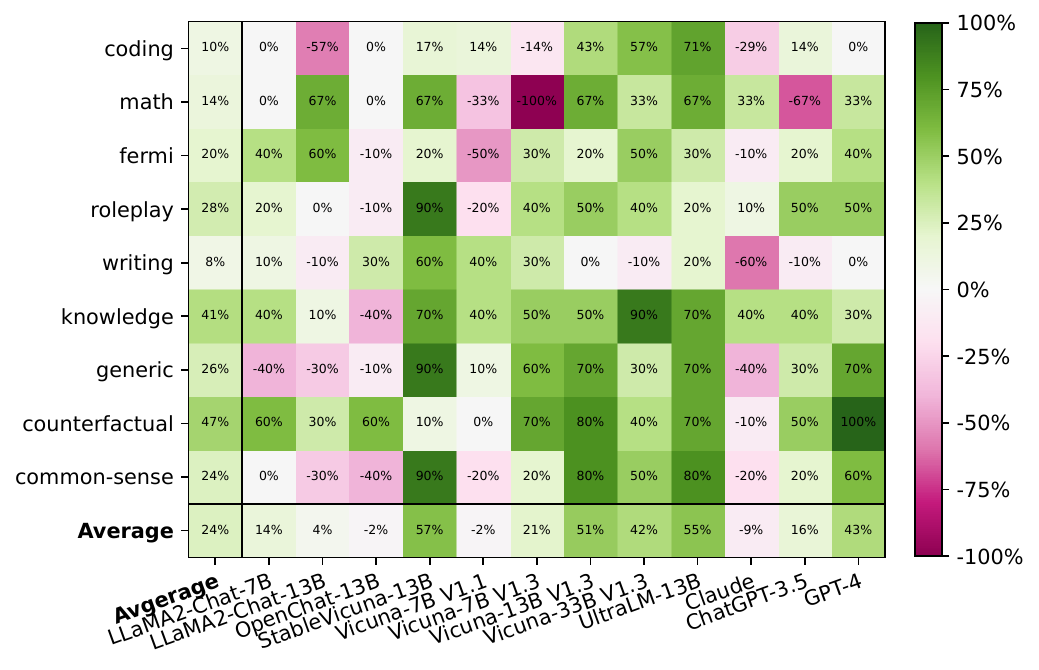}
         \caption{The ``general'' metric from \llmzoo{}.}
         \label{fig:model_category_llmzoo}
     \end{subfigure}
     \caption{Net win rates of different models and question categories. Each row corresponds to one question category, and one column corresponds to one model. (Evaluated using \Remark{chatgpt} as the judge.)}
     \label{fig:model_category}
\end{figure}

In the main text, we analyze how question categories and models affect \methodshort{}'s answer quality. Here, we show the per-model and per-category results.
For each model and question category, we compute the net win rates.
The results are in \cref{fig:model_category}. %

\subsubsection{Quality Breakdown: Metrics}
\begin{figure}[ht]
    \centering
    \includegraphics[width=0.7\linewidth]{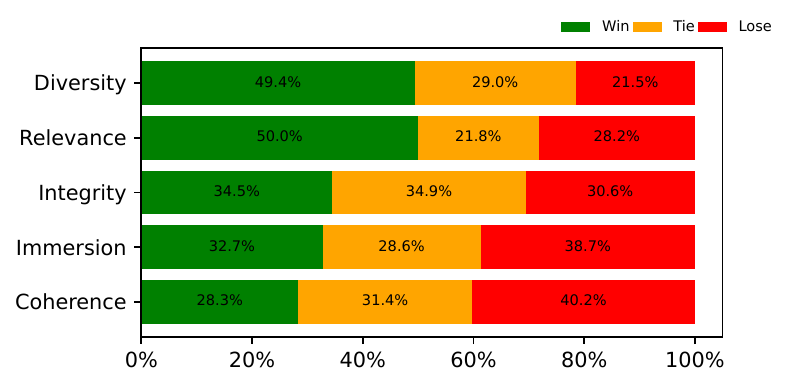}
    \caption{Win/tie/lose rates of \methodshort{} v.s. \methodbase{} generations using metrics from \llmzoo{}. \methodshort{} performs well on diversity and relevance, and relatively worse on coherence and immersion. (Evaluated using \Remark{chatgpt} as the judge.)}
    \label{fig:win_tie_lose_bar_DETAILS}
\end{figure}

All previous evaluations use metrics about the general quality of the answer.
In \cref{fig:win_tie_lose_bar_DETAILS}, we show more detailed metrics from \llmzoo{} to reveal in which aspects \methodshort{} can improve or hurt the answer quality. On average, we can see that \methodshort{} improves the diversity and relevance while hurting the immersion and coherence.

\section{Combining SoT-R with Model Quantization}

\revise{Model quantization is a widely-used model-level optimization to accelerate LLM inference, which is orthogonal to \methodshort{}. In this section, we evaluate the speed-ups of open-source models with both quantization and SoT on the Vicuna-80 dataset. Specifically, we adopt GPTQ~\citep{frantar2022gptq}\footnote{https://github.com/qwopqwop200/GPTQ-for-LLaMa} to apply 4-bit weight-only quantization and use \methodrshort{} instead of plain \methodshort{}.}

\subsection{Speed-ups of \methodshort{} + Quantization on Quantized Models}

\begin{figure}[H]
  \centering
    \begin{subfigure}[b]{0.49\textwidth}
    \centering
    \includegraphics[width=\textwidth]{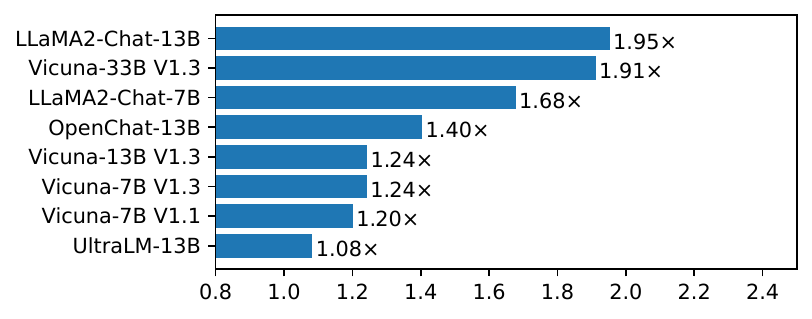}
    \caption{\methodrshort{} with the prompting router.}
    \label{fig:new_new_bar_quant_vicuna_speedup_gpt4_model}
  \end{subfigure}
  \hfill
  \begin{subfigure}[b]{0.49\textwidth}
    \centering
    \includegraphics[width=\textwidth]{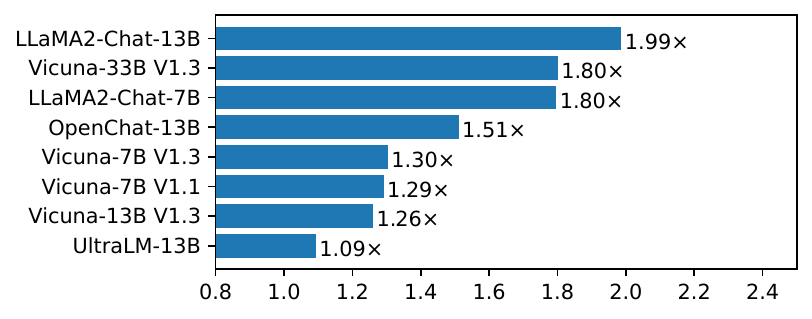}
    \caption{\methodrshort{} with the trained router.}
    \label{fig:new_new_bar_quant_vicuna_speedup_roberta_model}
  \end{subfigure}
  \caption{\revise{Speed-ups of the quantized model with \methodrshort{} generation w.r.t. the quantized model with normal generation on different models, on the Vicuna-80 dataset.}}
  \label{fig:new_new_bar_quant_sotr_vicuna_speedup_model}
\end{figure}

\begin{figure}[H]
  \centering
    \begin{subfigure}[b]{0.49\textwidth}
    \centering
    \includegraphics[width=\textwidth]{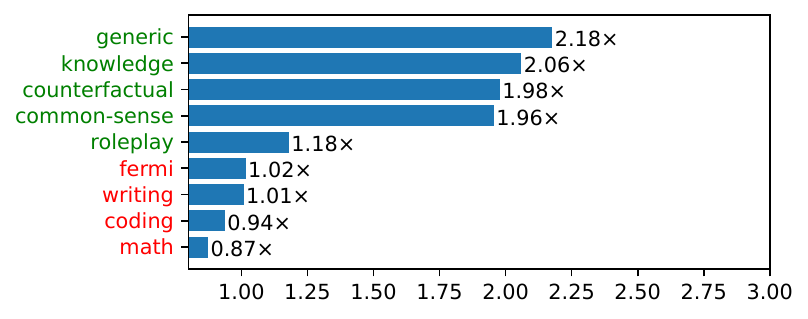}
    \caption{\methodrshort{} with the prompting router.}
    \label{fig:new_new_bar_quant_vicuna_speedup_gpt4_category}
  \end{subfigure}
  \hfill
  \begin{subfigure}[b]{0.49\textwidth}
    \centering
    \includegraphics[width=\textwidth]{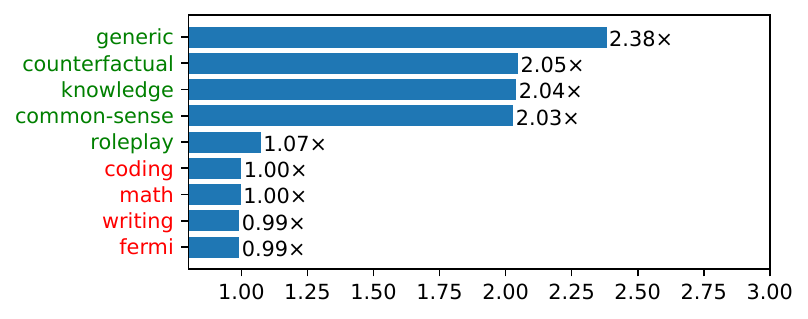}
    \caption{\methodrshort{} with the trained router.}
    \label{fig:new_new_bar_quant_vicuna_speedup_roberta_category}
  \end{subfigure}
  \caption{\revise{Speed-ups of the quantized model with \methodrshort{} generation w.r.t. the quantized model with normal generation, on different question categories of the Vicuna-80 dataset.}}
  \label{fig:new_new_bar_quant_sotr_vicuna_speedup_category}
\end{figure}

\revise{We first compare the latency of the quantized models in the normal and \methodshort{} modes to evaluate how much \methodshort{} can speed up quantized models. \cref{fig:new_new_bar_quant_sotr_vicuna_speedup_model} shows the speed-ups of \methodrshort{} on different quantized models. SoT-R obtain $1.08\times$ to $1.99\times$ speed-ups on all the models. \cref{fig:new_new_bar_quant_sotr_vicuna_speedup_category} shows the speed-ups of SoT-R on different categories. We can see that on the five question categories for which \methodshort{} can provide high-quality answers (i.e., \textit{knowledge}, \textit{generic}, \textit{common-sense}, \textit{roleplay}, \textit{counterfactual}), \methodrshort{} can speed up the overall answer generation process by 1.07$\times$ to 2.38$\times$.}

\subsection{Speed-ups of \methodshort{} + Quantization on Unquantized Models}

\begin{figure}[H]
  \centering
    \begin{subfigure}[b]{0.49\textwidth}
    \centering
    \includegraphics[width=\textwidth]{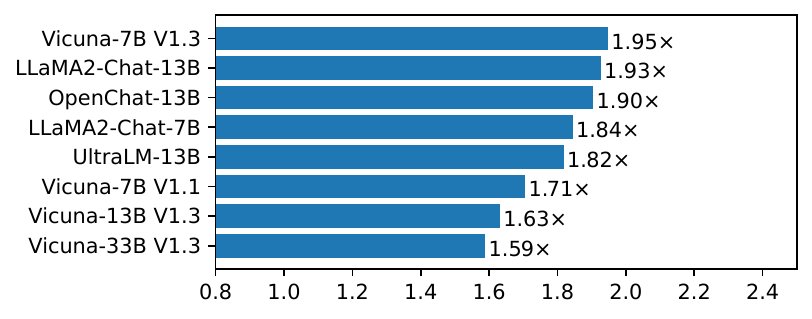}
    \caption{\methodrshort{} with the prompting router.}
    \label{fig:new_bar_quant_vicuna_speedup_gpt4_model}
  \end{subfigure}
  \hfill
  \begin{subfigure}[b]{0.49\textwidth}
    \centering
    \includegraphics[width=\textwidth]{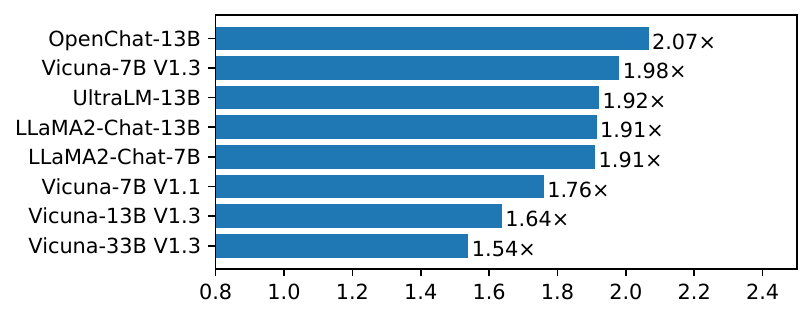}
    \caption{\methodrshort{} with the trained router.}
    \label{fig:new_bar_quant_vicuna_speedup_roberta_model}
  \end{subfigure}
  \caption{\revise{Speed-ups of the quantized model with \methodrshort{} generation w.r.t. the unquantized model with normal generation, on different models, on the \vicunadataset{} dataset.}}
  \label{fig:new_bar_quant_sotr_vicuna_speedup_model}
\end{figure}

\begin{figure}[H]
  \centering
    \begin{subfigure}[b]{0.49\textwidth}
    \centering
    \includegraphics[width=\textwidth]{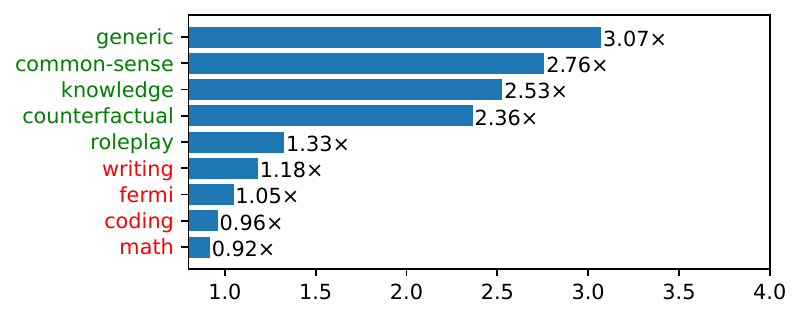}
    \caption{\methodrshort{} with the prompting router.}
    \label{fig:new_bar_quant_vicuna_speedup_gpt4_category}
  \end{subfigure}
  \hfill
  \begin{subfigure}[b]{0.49\textwidth}
    \centering
    \includegraphics[width=\textwidth]{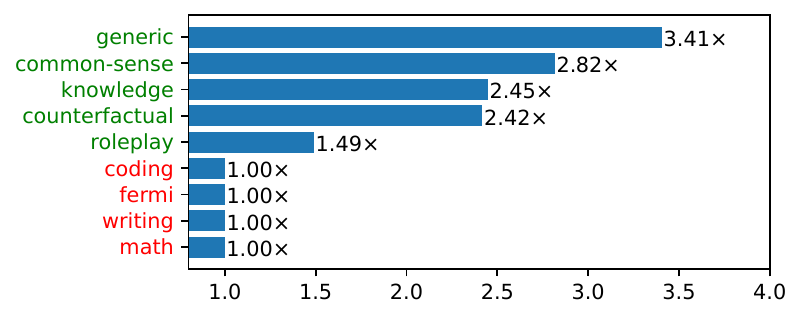}
    \caption{\methodrshort{} with the trained router.}
    \label{fig:new_bar_quant_vicuna_speedup_roberta_category}
  \end{subfigure}
  \caption{\revise{Speed-ups of the quantized model with \methodrshort{} generation w.r.t. the unquantized model with normal generation, on different question categories of the \vicunadataset{} dataset.}}
  \label{fig:new_bar_quant_sotr_vicuna_speedup_category}
\end{figure}

\revise{Here, we report the overall speed-ups of the quantization model with \methodrshort{} generation w.r.t. the unquantized model with normal generation. \cref{fig:new_bar_quant_sotr_vicuna_speedup_model} shows the speed-ups of SoT-R on different models. \methodrshort{} can obtain $1.54\times$ to $2.07\times$ speed-ups. \cref{fig:new_bar_quant_sotr_vicuna_speedup_category} shows the speed-ups of \methodrshort{} on different categories. On the five question categories for which \methodshort{} can provide high-quality answers (i.e., \textit{knowledge}, \textit{generic}, \textit{common-sense}, \textit{roleplay}, \textit{counterfactual}), \methodrshort{} can speed up the generation by 1.33$\times$ to 3.41$\times$ with the prompting and trained routers.}

\section{Additional SoT-R statistics}

\subsection{Number of Suitable Questions}

\revise{Overall, there are 37/80, 58/218, 371/1030 questions that are suitable for \methodshort{} in the \vicunadataset{}, \wizardlm{}, and LIMA datasets (according to human assessment), respectively.}

\begin{figure}[H]
  \centering
    \centering
    \includegraphics[width=0.5\textwidth]{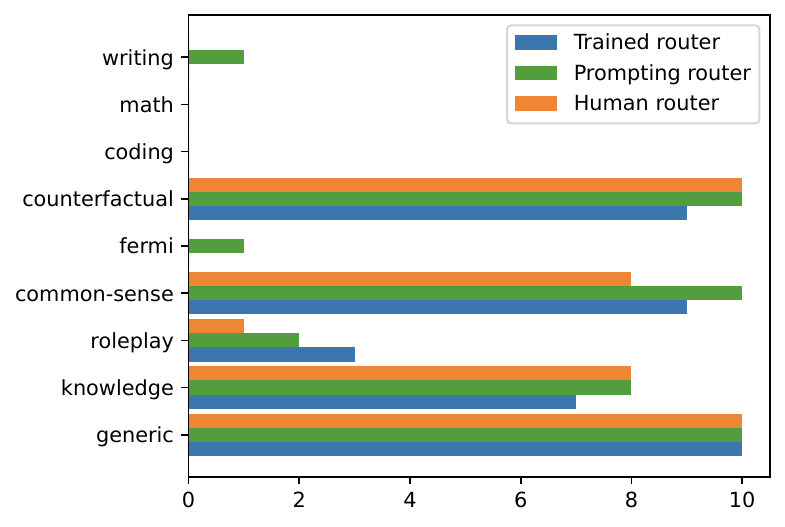}
  \caption{\revise{Number of questions suitable for \methodshort{} on the Vicuna-80 dataset.}} %
  \label{fig:bar_vicuna_sotr_num}
\end{figure}

\revise{\cref{fig:bar_vicuna_sotr_num} shows the number of questions that are suitable for \methodshort{} 
on Vicuna-80. On \textit{counterfactual}, \textit{commen-sense}, \textit{knowledge}, \textit{generic} categories, most questions are suitable for \methodshort{} based on the human assessment. The trained router and prompting router give out similar judgments.} %

\subsection{Peak Memory Overhead}

\begin{figure}[H]
  \centering
    \begin{subfigure}[b]{0.49\textwidth}
    \centering
    \includegraphics[width=\textwidth]{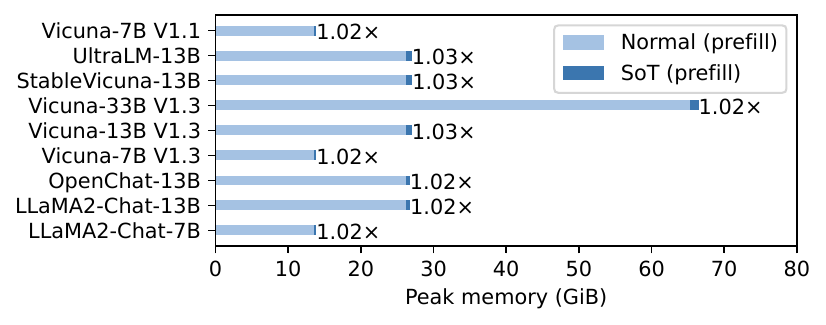}
    \caption{Peak memory in the prefilling phase.}
    \label{fig:bar_vicuna_prefill_mem_gpt4_model}
  \end{subfigure}
  \hfill
  \begin{subfigure}[b]{0.49\textwidth}
    \centering
    \includegraphics[width=\textwidth]{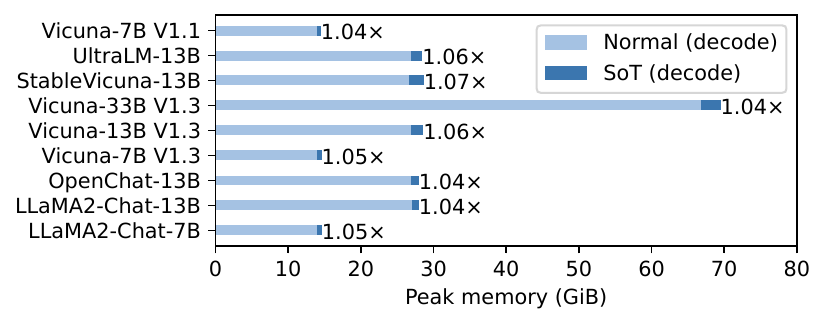}
    \caption{Peak memory in the decoding phase.}
    \label{fig:bar_vicuna_decode_mem_gpt4_model}
  \end{subfigure}
  \caption{\revise{Peak memory overhead of \methodrshort{} on different models on the Vicuna-80 dataset.}}
  \label{fig:bar_vicuna_mem_gpt4_model}
\end{figure}

\begin{figure}[H]
  \centering
    \begin{subfigure}[b]{0.49\textwidth}
    \centering
    \includegraphics[width=\textwidth]{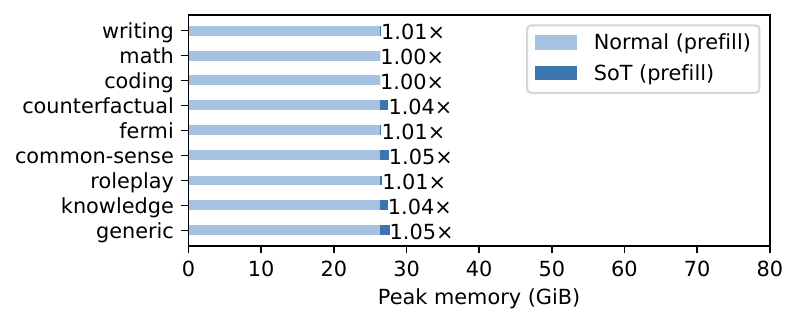}
    \caption{Peak memory in the prefilling phase.}
    \label{fig:bar_vicuna_prefill_mem_gpt4_category}
  \end{subfigure}
  \hfill
  \begin{subfigure}[b]{0.49\textwidth}
    \centering
    \includegraphics[width=\textwidth]{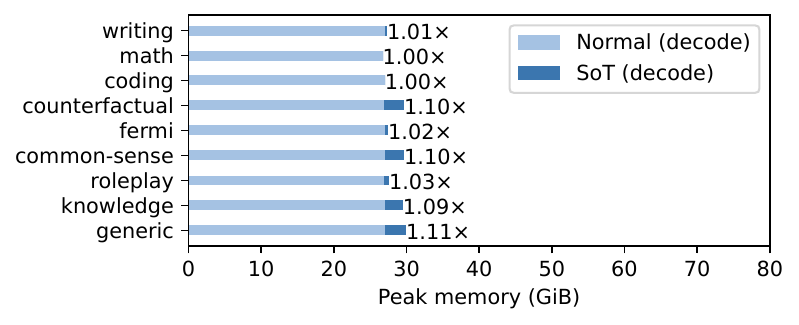}
    \caption{Peak memory in the decoding phase.}
    \label{fig:bar_vicuna_decode_mem_gpt4_category}
  \end{subfigure}
  \vspace{-0.5em}
  \caption{\revise{Peak memory overhead of \methodrshort{} on different question categories of Vicuna-80.}}
  \label{fig:bar_vicuna_mem_gpt4_category}
\end{figure}

\revise{\cref{fig:bar_vicuna_mem_gpt4_model} and \cref{fig:bar_vicuna_mem_gpt4_category} show the peak memory overhead of \methodrshort{} (with prompting router) on different models and different categories, respectively, on the Vicuna-80 dataset. We can see that, on all models and categories, the overhead of peak memory is quite small ($<$1.11$\times$).}

\subsection{Speed-ups with Different Number of Points}

\begin{figure}[ht]
    \centering
    \includegraphics[width=0.6\linewidth]{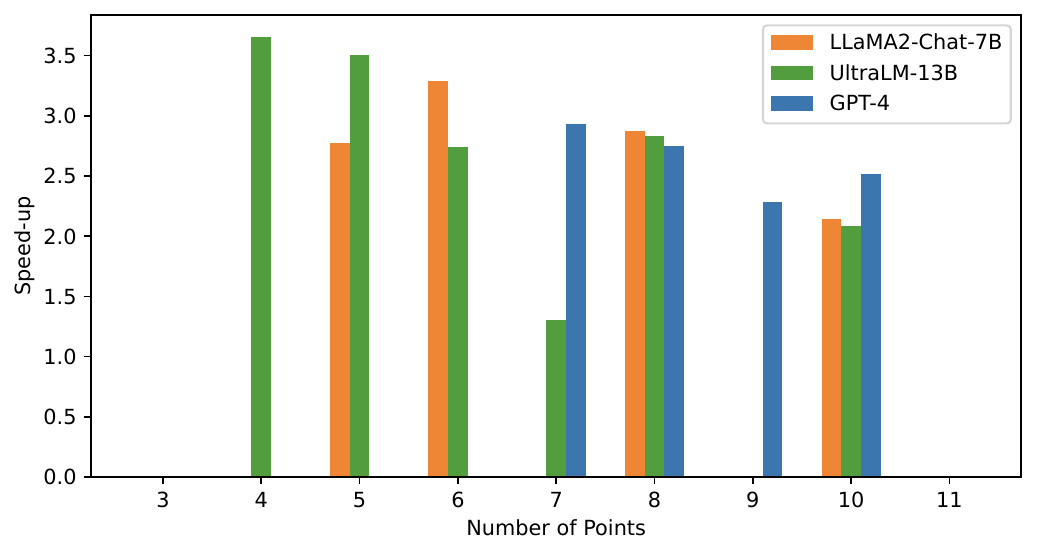}
     \vspace{-0.5em}
    \caption{\revise{The speed-ups with different number of points on the Vicuna-80 dataset.}}
    \label{fig:bar_vicuna_point_speedup}
\end{figure}

\revise{\cref{fig:bar_vicuna_point_speedup} shows the speed-ups with different numbers of points on Vicuna-80. 
To maintain clarity in the figure, we've chosen to display statistics for only three models. 
Note that as \methodshort{} cannot control the overall length to be the same as that of normal generation, it is not the case that a higher number of points leads to higher speed-ups.}

\section{Notes on Application Scenarios}
\revise{In a chatbot application, one might wonder why a reduced end-to-end latency can enhance the user experience. While human reading speeds are limited, there are many situations where we do not read responses sequentially.
Rather than reading the entire answer, one might prefer to (1) swiftly check the response's structure to confirm if the chatbot comprehended the question or (2) extract specific information rapidly without waiting for the generation of prologue or preceding points. Besides, from the quality aspect, even if we would like to check the entire answer, a well-defined structure in responses assists us in quickly parsing all the information.}

\revise{Moreover, beyond enhancing user experience, reduced end-to-end latency can significantly benefit emerging application scenarios like agent-agent interaction.}
\end{document}